%% file: main.tex
\documentclass[10pt,twocolumn,letterpaper]{article}

\usepackage[pagenumbers]{cvpr}
\usepackage{cuted}
\usepackage{graphicx}
\usepackage{array}
\usepackage{booktabs}

\input{preamble}

\definecolor{cvprblue}{rgb}{0.21,0.49,0.74}
\usepackage[pagebackref,breaklinks,colorlinks,allcolors=cvprblue]{hyperref}

\def\confName{CVPR}
\def\confYear{2027}

\title{TidyFlow: Geometry-Aware Tabletop Arrangement Generation with Human-Aligned Tidiness Evaluation}

\author{
Jiahui Qian \quad
Tiandong Wang \quad
Shubao Fan \\
Chenyu Niu \quad
Erqi Wu \\[0.5em]
Beijing Roborock Technology Co., Ltd. \\
{\tt\small
qianjiahui@roborock.com,
wangtiandong@roborock.com,
fanshubao@roborock.com} \\
{\tt\small
niuchenyu@roborock.com,
rafe@roborock.com}
}

\begin{document}
\maketitle

% Main paper.
\input{sec/0_abstract}
\input{fig/teaser_tidyflow}

\input{sec/introduction}
\input{sec/relatedwork}
\input{sec/ProposedMthod}

\input{sec/experiments}
\input{sec/conclusion}

% Statement carried over from the user-provided manuscript.
\section*{LLM Usage Statement}
Qwen 3.7 Plus was used for object semantic recognition
and arrangement planning during dataset construction.
OpenAI Codex assisted with code development and inspection.
ChatGPT was used for manuscript drafting, language editing,
and the preparation of illustrative figures.
All AI-assisted outputs were reviewed and verified
by the authors, who take full responsibility for
the research and manuscript.

{\small
\bibliographystyle{ieeenat_fullname}
\bibliography{main}
}

% Supplement appears in the SAME PDF, after main-paper references.
% The supplementary file should contain CONTENT ONLY (no
% \documentclass, \begin{document}, \end{document}, or separate
% \bibliography). Remove any duplicate supplementary heading inside it.
\clearpage
\appendix
\section*{Supplementary Material}
\input{sec/X_suppl}

\end{document}

%% file: preamble.tex
\usepackage{multirow}
\providecommand{\pospart}[1]{\left[#1\right]_{+}}

%% file: sec/0_abstract.tex
\begin{abstract}
Generating tidy tabletop arrangements that align with human organizational preferences requires considering both functional relationships among objects and geometric constraints within limited tabletop space. We present TidyFlow, a tabletop arrangement generation framework that integrates conditional Flow Matching with topology-aware geometric refinement. We construct two tabletop arrangement datasets based on functional grouping and geometric constraints, namely General Tabletop Layouts (GTL) and Workstation Tabletop Layouts (WTL), and introduce a six-dimensional tidiness scoring function. Conditional Flow Matching is employed to learn a multimodal distribution of target arrangements conditioned on object attributes and tabletop geometry. To address geometric errors in generated layouts, we construct topology-aware refinement data based on collision and boundary-violation patterns, and employ a pair-aware attention network to iteratively refine object poses. Experiments demonstrate that, after three refinement iterations, TidyFlow improves geometric validity from 23.21\% to 88.91\% on GTL and from 10.58\% to 57.91\% on WTL. Furthermore, a human preference study demonstrates a positive correlation between the proposed tidiness scores and human rankings, achieving a Spearman correlation coefficient of 0.5543.
\end{abstract}

%% file: fig/teaser_tidyflow.tex
\begin{figure}[t]
    \centering
    \includegraphics[
        width=\columnwidth
    ]{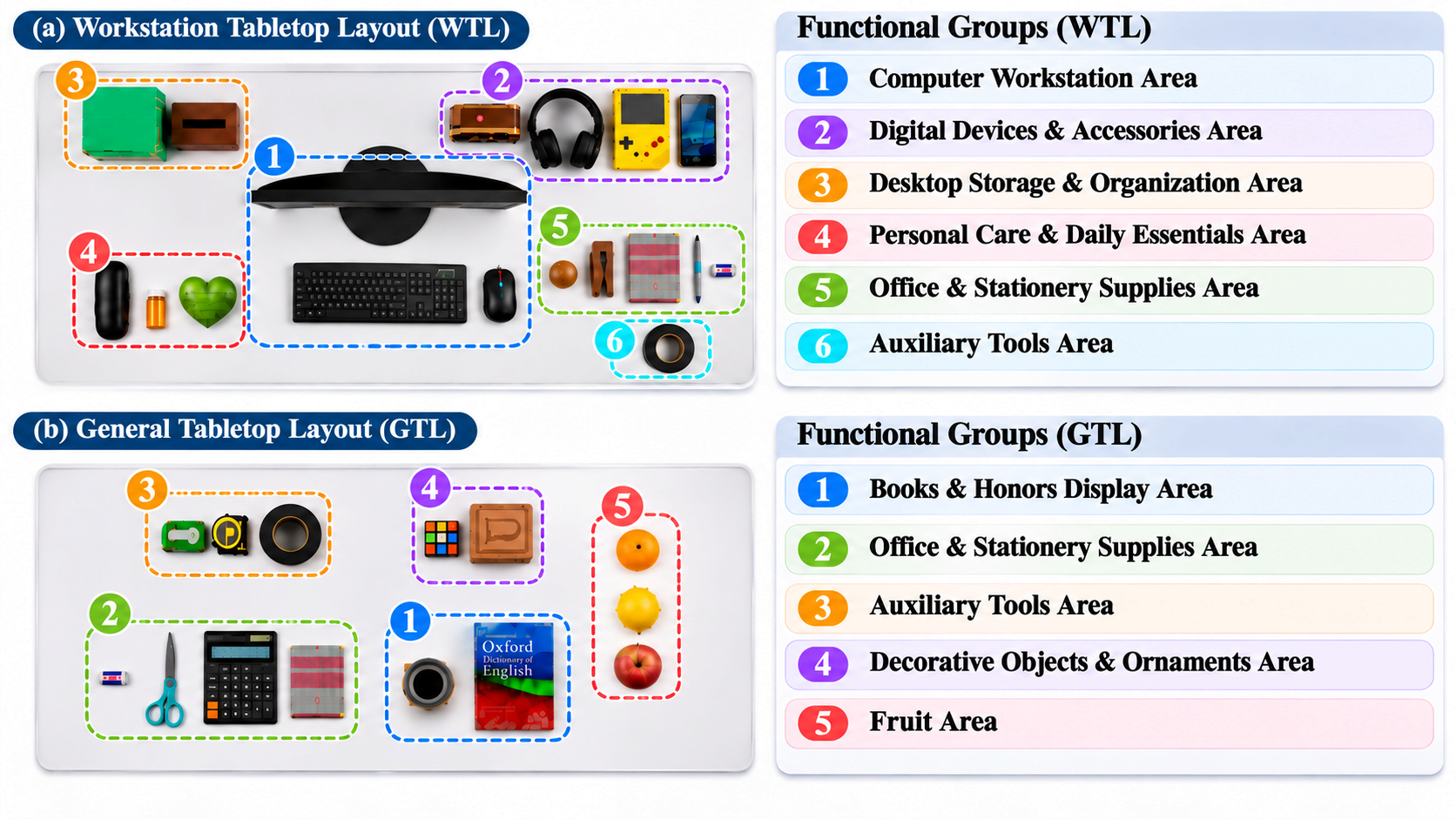}

    \caption{
    \textbf{GTL and WTL tabletop arrangement settings.}
    Examples of
    (a) Workstation Tabletop Layouts (WTL) and
    (b) General Tabletop Layouts (GTL).
    Colored dashed outlines indicate functional groups.
    }
    \label{fig:datasets}
\end{figure}

%% file: sec/introduction.tex
\section{Introduction}
\label{sec:introduction}

In robotic manipulation, tabletop target-layout generation
serves as an important intermediate step between scene
perception and robotic-arm planning. Rather than simply moving
objects into available space, tidy arrangements that reflect
human preferences must account for the relative positions,
orientations, and functional relationships of objects, as well
as the use of tabletop space. For example, a keyboard and mouse
typically form a workspace around a monitor, stationery items
are often placed together, and groups serving different
purposes should be maintain appropriate separation. Moreover,
even with the same objects and tabletop conditions, multiple
reasonable arrangements may exist, which are difficult to
represent with a single deterministic pose prediction. Figure~\ref{fig:datasets} illustrates the two
tabletop layout settings considered in this work.

Previous research has explored object arrangement through
functional semantics~\cite{sarch2022tidee},
layout generation~\cite{kapelyukh2023dallebot},
and geometric constraints~\cite{huang2023voxposer,ding2024open6dor}.
Tidy Up the Table~\cite{xu2023tidyup} and
Set It Up!~\cite{xu2024setitup} use functional relationships
among objects to guide arrangement generation, while
LEGO-Net~\cite{wei2023legonet} and
SceneFlow~\cite{wu2025sceneflow} learn spatial relationships
among objects to generate different configurations.
To improve geometric feasibility,
SceneFlow~\cite{wu2025sceneflow} and
STABLE~\cite{luo2026stable} incorporate geometric constraints
and layout refinement. In addition,
SceneScore~\cite{kapelyukh2023scenescore} and
Tidiness Score-Guided MCTS~\cite{kee2025tidiness}
evaluate arrangement quality using learned arrangement costs
or tidiness scores. However, generating diverse arrangements
does not guarantee that all of them satisfy geometric
constraints, and a collision-free, in-bounds arrangement does
not necessarily match human organizational preferences.

Generating tidy layouts from known object attributes and tabletop geometry presents three challenges. First, Reference Layouts used for training must reflect functional relationships among objects while accounting for object dimensions and tabletop boundary constraints. Second, generated Raw Layouts can exhibit heterogeneous geometric errors, including multiple objects involved in shared collisions, several disconnected collision regions, and simultaneous collisions and boundary violations. Generic geometric correction may reduce individual overlaps without resolving coupled conflicts, or may disturb spatial relationships that were already valid. Refinement must therefore address different error structures while distinguishing objects requiring repair from those whose arrangements should be preserved. Third, even a geometrically valid layout must be assessed for its functional grouping, alignment, and spacing, and the resulting scores must be evaluated for consistency with human tidiness preferences.

To address these challenges, we propose \emph{TidyFlow},
a tabletop target-layout generation framework combining
conditional layout generation, geometric refinement,
and tidiness evaluation.
First, we construct geometrically valid reference layouts
with meaningful functional organization through semantic
planning and geometry-constrained search, and train a
conditional Flow Matching model~\cite{lipman2023flowmatching}
to learn a distribution over target layouts and generate
multiple candidate configurations.
We then analyze collisions and boundary violations produced
by the generator on the training set and use these error
patterns to construct refinement data with diverse error
structures. Refiner models geometric relationships between objects
to predict local pose adjustments, while safety constraints
prevent the introduction of new collisions and boundary
violations during refinement.
Finally, we select a target layout from geometrically
valid candidates using an interpretable six-dimensional
tidiness score and assess the score's agreement with
human organizational preferences through subjective evaluations.

On General Tabletop Layouts (GTL) and Workstation Tabletop
Layouts (WTL), geometric refinement increases the
candidate-level Geometric Validity Rate (GVR) from 23.21\% to
88.91\% and from 10.58\% to 57.91\%, respectively. To sum up, we contribute:

\begin{itemize}

    \item Two tabletop layout datasets, GTL and WTL,
    that incorporate functional organization and
    geometric validity across varying scene complexity.

    \item A conditional Flow Matching framework for
    diverse layout generation, coupled with a
    topology-aware Refiner that uses generator-induced
    errors, pairwise geometry, and repair-preservation
    supervision to reduce violations and unnecessary
    pose changes.

    \item A six-dimensional tidiness score for evaluating
    functional organization and spatial arrangement,
    together with human-rating and preference-ranking
    experiments to assess its agreement with human
    preferences.

\end{itemize}

%% file: sec/relatedwork.tex
\section{Related Work}
\label{sec:related_work}

We review three lines of research closely related to our work: (1) functional object arrangement and tidiness evaluation, (2) generative modeling for object layouts, and (3) geometric constraints and layout refinement.

\begin{figure*}[t]
    \centering

    \includegraphics[
        width=\textwidth
    ]{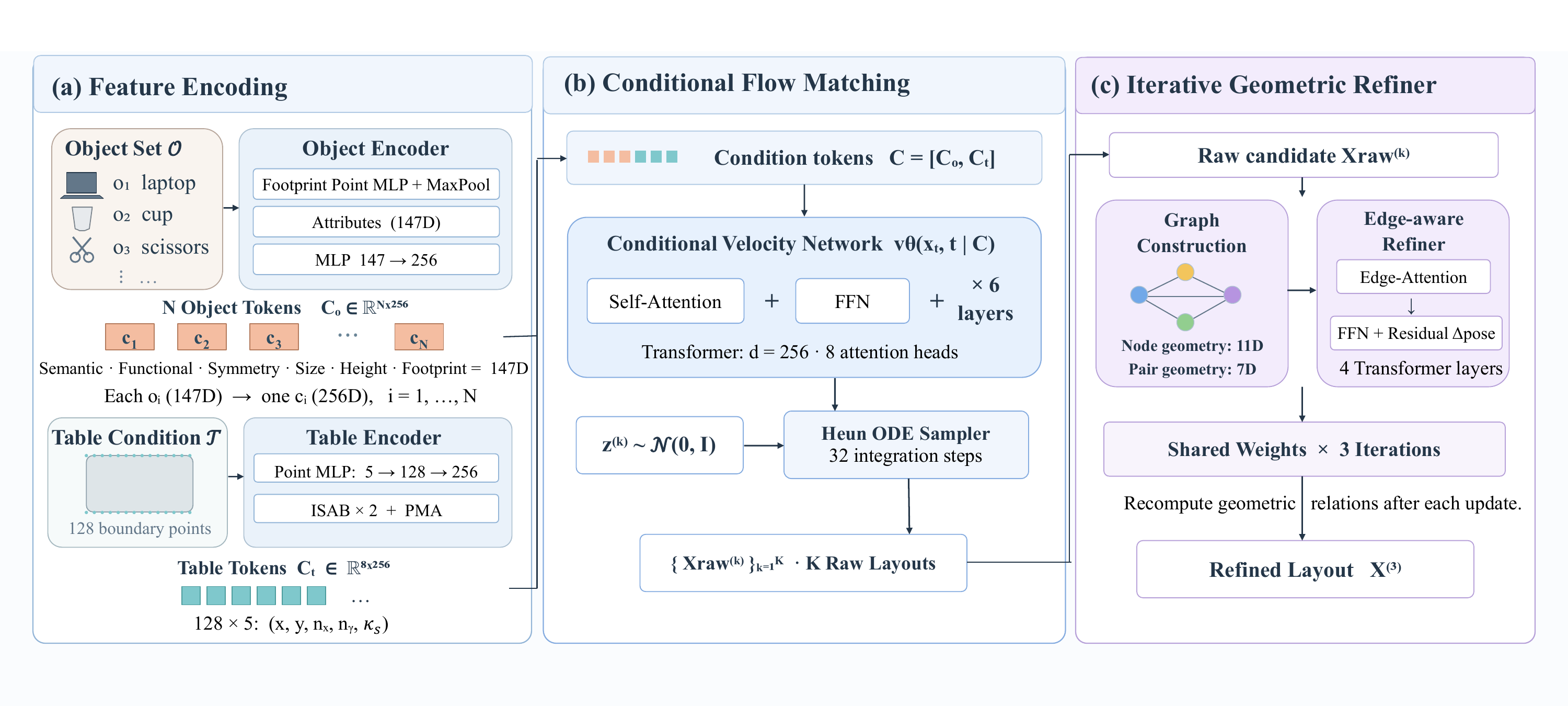}

    \caption{
    \textbf{Overview of the TidyFlow framework.}
    (a) Object attributes and footprint features
    are encoded into 256-dimensional object tokens,
    while 128 Tabletop Support Polygon boundary samples are summarized
    into eight table tokens.
    (b) A six-layer Transformer predicts the
    conditional flow velocity, and a 32 integration steps generate multiple Raw Layouts
    from independent Gaussian samples.
    (c) A four-layer edge-aware geometric Refiner
    performs three weight-shared updates, with
    geometric relations recomputed after each iteration.
    }

    \label{fig:tidyflow_architecture}
\end{figure*}

\noindent\textbf{Functional Object Arrangement and Tidiness Evaluation.}
Meaningful object arrangements require functional organization
and spatial regularity beyond geometric feasibility.
\emph{Tidy Up the Table}~\cite{xu2023tidyup}
combines commonsense reasoning with visual-spatial tidiness
assessment for tabletop rearrangement.
TidyBot~\cite{wu2023tidybot} learns personalized placement
preferences from user examples, while
\emph{Set It Up!}~\cite{xu2024setitup}
grounds functional instructions into spatial relations.
Scene Retargeting~\cite{kim2026sceneretargeting}
transfers semantically coherent arrangements across
different scenes.

Arrangement evaluation is closely related to this problem.
Tidiness Score-Guided MCTS~\cite{kee2025tidiness}
learns a visual tidiness discriminator to guide
rearrangement search, whereas
SceneScore~\cite{kapelyukh2023scenescore}
estimates arrangement costs through object-relation graphs.
Fashae \etal~\cite{fashae2026preferences}
investigate interpretable factors underlying human
organizational preferences.
These studies highlight the importance of evaluating
functional tidiness beyond collision-free placement.

% Keep the original framework figure unchanged.

\noindent\textbf{Generative Modeling for Object Layouts.}
Generative models capture diverse spatial arrangements
rather than predicting a single deterministic layout.
ATISS~\cite{paschalidou2021atiss}
models indoor scenes as unordered object sets,
while LEGO-Net~\cite{wei2023legonet}
learns regular object rearrangements through iterative denoising.
DiffuScene~\cite{tang2024diffuscene}
uses diffusion models to jointly generate object attributes
and spatial configurations.
For object rearrangement,
StructFormer~\cite{liu2022structformer}
predicts spatial structures from language instructions,
and StructDiffusion~\cite{liu2023structdiffusion}
generates object poses conditioned on language and geometry.
TarGF~\cite{wu2022targf}
learns target gradient fields to guide rearrangement
without explicitly specified goal configurations.

Flow Matching~\cite{lipman2023flowmatching}
provides an alternative generative framework by learning
continuous velocity fields along probability paths.
LayoutFlow~\cite{guerreiro2024layoutflow}
applies flow-based generation to graphic layouts,
while SceneFlow~\cite{wu2025sceneflow}
extends this approach to indoor scene synthesis
with geometric constraints.
Despite their ability to model diverse configurations,
generated layouts may still exhibit collisions
and boundary violations, particularly in crowded spaces.

\noindent\textbf{Geometric Constraints and Layout Refinement.}
Geometric feasibility is commonly addressed through
collision assessment, constrained search,
or post-generation correction.
StructDiffusion~\cite{liu2023structdiffusion}
employs a learned collision discriminator to improve
the physical validity of generated arrangements,
while LGMCTS~\cite{chang2024lgmcts}
integrates spatial reasoning with geometrically informed search.
PhyScene~\cite{yang2024physcene}
introduces physics-based guidance for indoor scene generation.

More closely related to our approach,
SceneFlow~\cite{wu2025sceneflow}
combines flow-based scene generation with non-overlap
constraints and a geometry refinement network.
STABLE~\cite{luo2026stable}
integrates language-based semantic reasoning
with a flow-based physics corrector
for tabletop scene generation.
These methods demonstrate the value of combining
generative models with geometric correction.
Building on this direction, TidyFlow focuses on
functionally organized tabletop layouts and employs
a relation-aware Refiner trained on generator-induced
geometric violations.
Its local, safety-constrained updates aim to improve
geometric validity while preserving the functional
organization established during generation.

%% file: sec/ProposedMthod.tex
\section{Proposed Method}
\label{sec:proposed_method}

We propose \emph{TidyFlow}, a framework for tabletop target-layout generation that combines semantic--geometric Reference Layout construction, conditional Flow Matching, and topology-aware refinement. Given a Tabletop Support Polygon and an object set, the model generates multiple candidate layouts, corrects local geometric errors using Refiner, and selects a tidy target layout among the geometrically valid candidates. The overall framework is illustrated in Fig.~\ref{fig:tidyflow_architecture}.

\subsection{Problem Formulation}
\label{sec:problem_formulation}

Given a set of $N$ objects $\mathcal O=\{o_i\}_{i=1}^{N}$ and a Tabletop Support Polygon $T\subset\mathbb R^2$, our goal is to generate tidy tabletop arrangements that satisfy geometric constraints while preserving functional organization.

Each object is represented as $o_i=(c_i,g_i,\rho_i,d_i,F_i^0)$, where $c_i$, $g_i$, $\rho_i$, $d_i$, and $F_i^0$ denote its semantic category, functional group, rotational symmetry, geometric dimensions, and local planar footprint, respectively. We assume single-layer arrangements with fixed support orientations; the model predicts only planar positions and yaw angles in the tabletop coordinate frame. The pose of object $i$ is represented as
\begin{equation}
\label{eq:tidyflow_pose}
X_i=
\left[
\frac{x_i}{L},\frac{y_i}{L},\cos\theta_i,\sin\theta_i
\right]^\top,
\qquad
L=\frac12\max(W_T,D_T),
\end{equation}
where $W_T$ and $D_T$ are the width and depth of the tabletop bounding box. The common scale $L$ preserves geometric proportions across tabletops of different sizes.

A dataset condition is defined as
$C=(\mathcal O,T)$, consisting of an object set
and a Tabletop Support Polygon.
Within each dataset, object sets are unique
across conditions. Given $C$, the model learns the conditional distribution
$p_\phi(X\mid C)$ of target layouts and samples
multiple candidate arrangements.

\subsection{Semantic--Geometric Tabletop Layout Dataset Construction}
\label{sec:dataset_construction}

To learn a distribution of layouts that exhibit functional organization and satisfy geometric constraints, we construct the dataset through two stages: semantic planning and geometric planning. These stages determine the functional organization of objects and their geometrically feasible poses, respectively.

\paragraph{Semantic planning.}
Given captured top-down and oblique tabletop views, we use Qwen 3.7 Plus as a vision--language model for both semantic understanding and arrangement planning. The model identifies object categories, functional attributes, and potential usage relationships from the images. We also extract the Tabletop Support Polygon. Conditioned on the identified object set and tabletop geometry, the same model generates multiple semantic layout plans specifying the ordering of functional groups, their relative positions, and object proximity relationships, rather than directly predicting precise object coordinates~\cite{huang2022zeroshotplanners,liang2023codeaspolicies}. For computer-workstation scenes, we further consider usage-specific constraints, including the relative placement of the keyboard and monitor, the preferred side for mouse placement, and the reservation of tabletop workspace. Semantic planning allows the same object set to admit multiple reasonable arrangements.

\paragraph{Geometric planning.}
Using the actual object dimensions, fixed support orientations, and the Tabletop Support Polygon, we convert semantic layout plans into explicit planar poses. Dynamic free-space search and global backtracking assign positions and orientations to objects under geometric constraints. Collision avoidance and complete footprint containment within the tabletop are treated as hard constraints, whereas the relative positions of functional groups and within-group arrangement conventions are treated as layout preferences.

Let $\mathcal C(X)$ denote the set of colliding object pairs and $\mathcal B(X)$ the set of out-of-bounds objects. We define
\begin{equation}
\label{eq:tidyflow_violation_sets}
\begin{aligned}
\mathcal C(X)
&=\{(i,j)\mid i<j,\ g_{ij}(X)<0\},\\
\mathcal B(X)
&=\{i\mid F_i(X_i)\not\subseteq T\}.
\end{aligned}
\end{equation}
The geometric validity of a layout is then
\begin{equation}
\label{eq:tidyflow_validity}
\operatorname{Valid}(X)=
\mathbb I\!\left[
\mathcal C(X)=\varnothing
\land
\mathcal B(X)=\varnothing
\right].
\end{equation}
Here, $g_{ij}(X)$ is the signed separation gap between two oriented rectangular footprints, computed using the separating axis theorem (SAT), and $F_i(X_i)$ is the footprint of object $i$ after translation and rotation. For non-convex tabletop polygons, boundary validation considers containment of the complete footprint region.

\paragraph{Tidiness evaluation and layout selection.}

Geometric validity ensures that objects can be placed
on the tabletop without conflicts but does not fully
characterize the tidiness of an arrangement. We evaluate valid layouts using six
interpretable dimensions: Compactness, Orientation,
Alignment, Centering, Clearance, and Group Organization.

Let $s_d(X;C)$ denote the six component scores,
with weights
$w=(0.30,0.15,0.20,0.20,0.15,0.10)$
in the same order. For the set $\mathcal A(C)$ of applicable components,
the predefined overall tidiness score is

\begin{equation}
\label{eq:tidyflow_tidiness}
\Psi(X;C)=
\frac{
\sum_{d\in\mathcal A(C)}w_d s_d(X;C)
}{
\sum_{d\in\mathcal A(C)}w_d
}.
\end{equation}

The orientation component uses the corresponding
Reference Layout's yaw as its reference when available. Component definitions and applicability rules are provided in the supplementary material. During dataset construction, $\Psi$ is used to rank
geometrically valid candidates and select Reference
Layouts. The same predefined score is subsequently
used to evaluate generated layouts.

The scoring function was fixed before the human
preference survey, which serves to assess its
agreement with human judgments.
\subsection{Conditional Flow Matching}
\label{sec:flow_matching}

\paragraph{Object and tabletop encoding.}
To model both inter-object relationships and tabletop geometry, we use object and tabletop tokens as joint generation conditions.

The object encoder combines semantic categories, functional groups, rotational symmetry, dimensions, and local footprint features. A pointwise MLP processes eight points on each fixed local footprint, and max pooling aggregates them into a geometric descriptor, which is fused with semantic attributes to obtain object features. Padding masks accommodate variable object counts.

The tabletop encoder directly uses 128 boundary samples from the Tabletop Support Polygon, each containing position, surface normal, and boundary turning angle. Two Induced Set Attention Blocks (ISABs) model relationships among boundary samples. Pooling by Multihead Attention (PMA), with learnable query vectors, then produces tabletop tokens~\cite{lee2019settransformer}.

During generation, we additionally compute the distance of each object to the Tabletop Support Polygon boundary, degree of boundary violation, and correction direction from its current pose, and fuse these features with object, pose, and time embeddings.

Object and tabletop tokens are jointly processed by a Transformer to predict a four-dimensional pose velocity field for each object:
\[
\hat v=v_\theta(X_t,t,C).
\]

\paragraph{Conditional Flow Matching objective.}

We adopt conditional Flow Matching~\cite{lipman2023flowmatching}
to learn the distribution of Reference Layouts. Given Gaussian noise $X_0$, a Reference Layout $X_c$,
and a sampled time $t$, we construct the probability path
\begin{equation}
\label{eq:tidyflow_flow_path}
\begin{aligned}
X_t &= a(t)X_0+tX_c,\\
a(t) &= 1-(1-\sigma)t,\\
v^\star &= X_c-(1-\sigma)X_0.
\end{aligned}
\end{equation}

The velocity network is trained solely with the standard Flow Matching objective:
\begin{equation}
\label{eq:tidyflow_fm_loss}
\mathcal L_{\mathrm{FM}}=
\mathbb E_{X_0,X_c,t}\!\left[
\left\|v_\theta(X_t,t,C)-v^\star\right\|_2^2
\right].
\end{equation}
The corresponding Reference Layout can be estimated from the predicted velocity as:
\begin{equation}
\label{eq:tidyflow_reconstructed_layout}
\hat X_c=(1-\sigma)X_t+a(t)v_\theta(X_t,t,C).
\end{equation}
After training, the conditional Flow model generates multiple candidate target layouts from different noise initializations, which are passed to the subsequent Refiner to correct residual geometric errors.

\subsection{Topology-Aware Refinement Data and Refiner}
\label{sec:zmt_refiner}

Samples from the generative model may contain collisions among multiple objects and Tabletop Support Polygon boundary violations. We propose Refiner to correct local geometric conflicts while preserving already valid relationships through generator-error-statistics-guided refinement data construction, pairwise relationship modeling, and safety-constrained iterative updates.

\paragraph{Error-statistics-guided refinement data.}
Instead of constructing Refiner training data by simply perturbing object positions at random, we use error statistics from a frozen Flow generator on the training set to identify geometric conflict types that should be emphasized.

Specifically, we first collect colliding object pairs, penetration depths, boundary violations, and collision relationship structures in generated layouts, aggregated by tabletop type and object count. Starting from geometrically valid Reference Layouts $X_c$, we then synthesize local perturbations with different error structures.

Treating objects as graph nodes and collision relationships as edges, the refinement data cover single-pair collisions, multiple disjoint collisions, collisions sharing a central object, and multi-edge conflict structures. We additionally include boundary-only, collision-only, and joint collision--boundary errors.

The selection of perturbed objects and error types is guided by training-set statistics, allowing the Refiner to encounter local conflicts closer to the generator's actual error distribution rather than being limited to simple overlaps between individual object pairs.

We also include a proportion of unperturbed valid layouts during training to encourage the network to preserve poses when no correction is needed.

\paragraph{Pair-aware refinement network.}
Refiner employs a separately parameterized edge-aware attention network~\cite{hussain2022egt} to predict local pose corrections for individual objects from the current layout.

Each object node combines static semantic and geometric attributes, its current pose, collision severity, and boundary status. For each object pair $(i,j)$, relative displacement, signed separation gap, collision separation direction, functional-group relationship, and size ratio are encoded as a seven-dimensional edge feature $e_{ij}$.

Unlike attention mechanisms that model object relationships through node features alone, we incorporate edge features into both attention weights and message values:
\begin{equation}
\label{eq:tidyflow_pair_aware_attention}
\begin{aligned}
\alpha_{ij}^{h}
&=\operatorname{softmax}_{j}
\left(
\frac{(q_i^h)^\top k_j^h}{\sqrt{d_h}}
+\psi_b^h(e_{ij})
\right),\\
m_i^h
&=\sum_j\alpha_{ij}^{h}
\left(v_j^h+\psi_v^h(e_{ij})\right),
\end{aligned}
\end{equation}
where $\psi_b^h$ and $\psi_v^h$ generate edge-dependent attention biases and message corrections, respectively. This enables the prediction of pose adjustments to explicitly account for collisions and spatial relationships between neighboring objects.

\paragraph{Repair-and-preservation objective.}

Local refinement must resolve existing collisions
and boundary violations while avoiding unnecessary
changes to already valid arrangements. We divide supervision into repair and preservation
according to the input geometric state. 

Repair targets collision-involved or out-of-bounds
objects, while positional preservation and a
symmetry-aware yaw loss apply to other valid objects. The eight-term training objective is

\begin{equation}
\label{eq:tidyflow_refiner_loss}
\begin{aligned}
\mathcal L_R={}&
\mathcal L_{\mathrm{target}}
+0.075\mathcal L_{\mathrm{stay}}
+0.075\mathcal L_{\mathrm{yaw}}\\
&+0.08\mathcal L_{\mathrm{CR}}
+0.20\mathcal L_{\mathrm{CP}}\\
&+0.08\mathcal L_{\mathrm{BR}}
+0.08\mathcal L_{\mathrm{BD}}
+0.20\mathcal L_{\mathrm{BN}}.
\end{aligned}
\end{equation}

Here, $\mathcal L_{\mathrm{target}}$
supervises normalized pose increments;
$\mathcal L_{\mathrm{stay}}$ and
$\mathcal L_{\mathrm{yaw}}$ discourage
unnecessary movement of non-target objects;
and the remaining terms address collision
and boundary repair and preservation. The yaw-preservation term uses a doubled angle
for $180^\circ$-symmetric objects and excludes objects with continuous rotational symmetry.

Exact masks, signed-gap margins, and boundary
soft-min definitions are provided in the
supplementary material.
\paragraph{Safety-constrained iterative refinement.}
The Refiner is trained for a single pose correction and applied iteratively with shared weights at inference to progressively address residual errors.

Given the current layout $X$, the Refiner predicts a full correction proposal
\[
Y=\mathcal N\big(X+f_\omega(X,C)\big),
\]
where $\mathcal N$ normalizes the yaw representation. For a candidate step size $\alpha$, the updated layout is
\[
X_\alpha=\mathcal N\big(X+\alpha(Y-X)\big).
\]
To prevent new geometric errors during refinement, only updates satisfying
\begin{equation}
\label{eq:tidyflow_safe_update}
\mathcal C(X_\alpha)\subseteq\mathcal C(X),
\qquad
\mathcal B(X_\alpha)\subseteq\mathcal B(X)
\end{equation}
are accepted.

\begin{figure}[t]
    \centering

    \includegraphics[
        width=\columnwidth
    ]{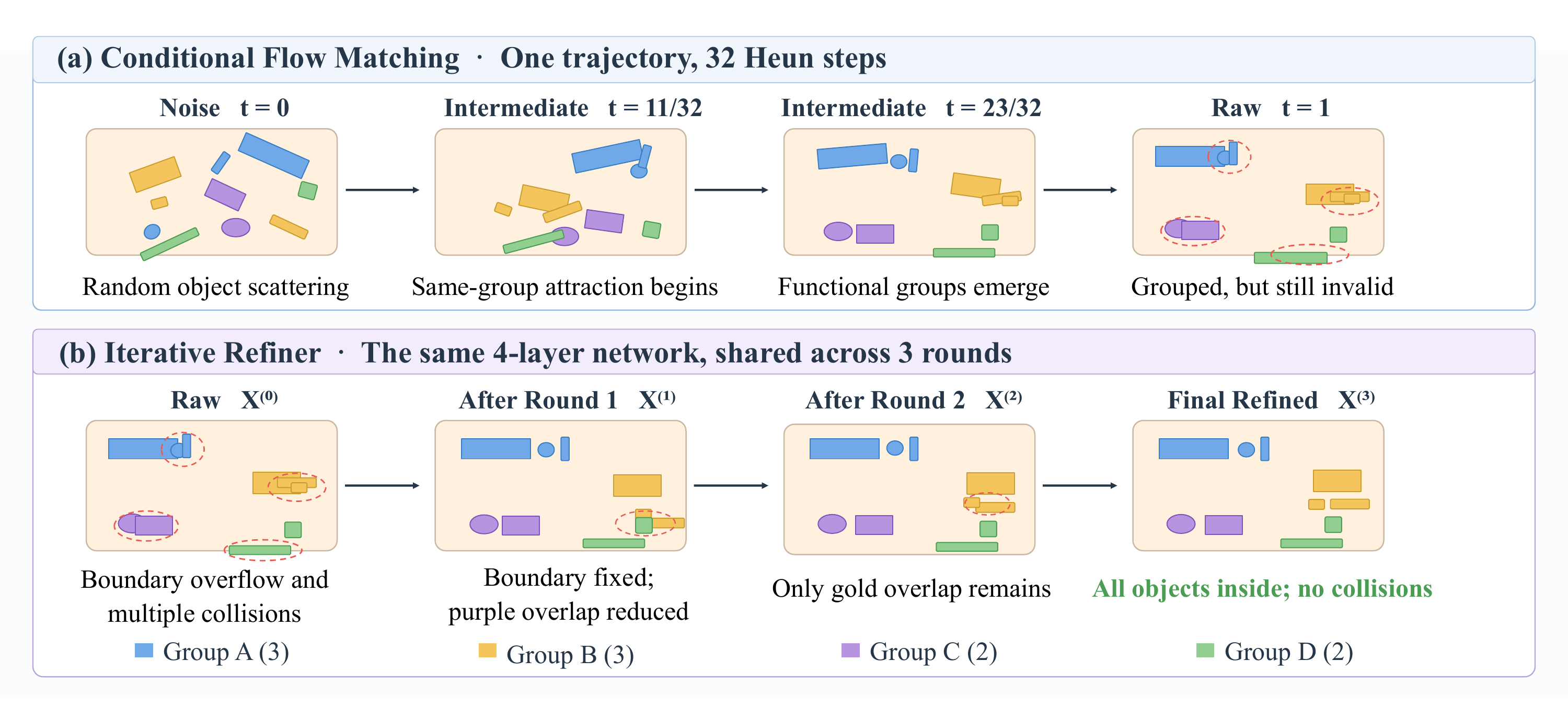}

    \caption{
    \textbf{Layout evolution and iterative geometric refinement.}
    The proposed framework progressively generates structured
    layouts and corrects geometric violations through
    three refinement iterations.
    }

    \label{fig:tidyflow_evolution}
\end{figure}

Among admissible step sizes, we prioritize updates with fewer geometric violations and then lower total penetration depth. 
The $\alpha=0$ option preserves the current
round's input if no improving update is selected. Three refinement iterations are performed at inference, with node and pairwise geometric features recomputed after each iteration. The refined candidates are subsequently filtered by the common hard geometric checker, and the final layout is selected using the score $\Psi$ defined in Sec.~\ref{sec:dataset_construction}. Fig.~\ref{fig:tidyflow_evolution} illustrates the complementary roles of conditional flow matching and iterative geometric refinement. 

%% file: sec/experiments.tex
\section{Experiments}
\label{sec:experiments}

We evaluate TidyFlow on GTL and WTL through four experiments:
comparisons with existing methods, agreement between the tidiness
score $\Psi$ and human preferences, ablations on shared-weight
Refiner iterations, and Refiner component analysis. Additional experimental details and analyses
are provided in the supplementary material.

\subsection{Comparison with Baseline Methods}
\label{sec:baseline_comparison}

Under identical test conditions, 32 candidates
per condition, and a shared geometric evaluator,
we compare TidyFlow with LayoutGPT~\cite{feng2023layoutgpt},
StructDiffusion~\cite{liu2023structdiffusion},
SceneScore~\cite{kapelyukh2023scenescore},
and LEGO-Net~\cite{wei2023legonet}.
As shown in \cref{tab:comparison}, TidyFlow achieves the
highest mean predefined tidiness scores on GTL (0.7192)
and WTL (0.6943). Its GVR reaches 88.91\% and 57.91\%,
outperforming LayoutGPT, StructDiffusion, and LEGO-Net
on both datasets. TidyFlow also obtains the lowest BVR
on WTL (6.23\%) and FCR@32 values of 100.00\% and
97.36\% on GTL and WTL, respectively.

\cref{fig:qualitative_comparison} presents representative
arrangements under identical object sets and tabletop
conditions. Although SceneScore achieves the highest
GVR on both datasets, its displayed layouts exhibit
weak functional organization.
In contrast, TidyFlow produces more coherent functional
grouping and spatial arrangements, achieving the highest
$\Psi$ in all four examples. For the WTL condition with
13 objects, SceneScore generates more valid candidates, whereas TidyFlow obtains a higher
$\Psi$. These results highlight
that geometric validity alone does not guarantee
functional tidiness.

% =========================================================
% Qualitative Comparison: GTL and WTL
% 4 Conditions x 5 Methods
% =========================================================

% ---------------------------------------------------------
% Image cell: larger fonts for Psi and Valid
% ---------------------------------------------------------

\newcommand{\QualCell}[3]{%
    \begin{minipage}[t]{\linewidth}
        \vspace{0pt}
        \centering

        \includegraphics[
            width=\linewidth
        ]{#1}\par

        \vspace{1pt}

        {\fontsize{8}{9}\selectfont
        \makebox[\linewidth][c]{%
            $\Psi=#2$\hspace{3pt}Valid=#3%
        }\par}

    \end{minipage}%
}

% ---------------------------------------------------------
% Dataset label: horizontally and vertically centered
% ---------------------------------------------------------

\newcommand{\QualLabel}[2]{%
    \parbox[t][0.18\textwidth][c]{\linewidth}{%
        \centering
        {\small #1\par}
        {\small $N=#2$\par}
    }%
}

% =========================================================
% Qualitative Comparison Figure
% =========================================================

\begin{figure*}[t]
    \centering

    \begingroup

    \setlength{\tabcolsep}{0.5pt}
    \renewcommand{\arraystretch}{1.0}

    \begin{tabular}{@{}
        >{\centering\arraybackslash}p{0.075\textwidth}
        *{5}{>{\centering\arraybackslash}p{0.18\textwidth}}
        @{}}

    % =====================================================
    % Header
    % =====================================================

    {\scriptsize Dataset/$N$}
    &
    {\small SceneScore}
    &
    {\small StructDiffusion}
    &
    {\small LayoutGPT}
    &
    {\small LEGO-Net}
    &
    {\small\textbf{TidyFlow (Ours)}}

    \\[0pt]

    % =====================================================
    % GTL - N = 8
    % =====================================================

    \QualLabel{GTL}{8}
    &
    \QualCell{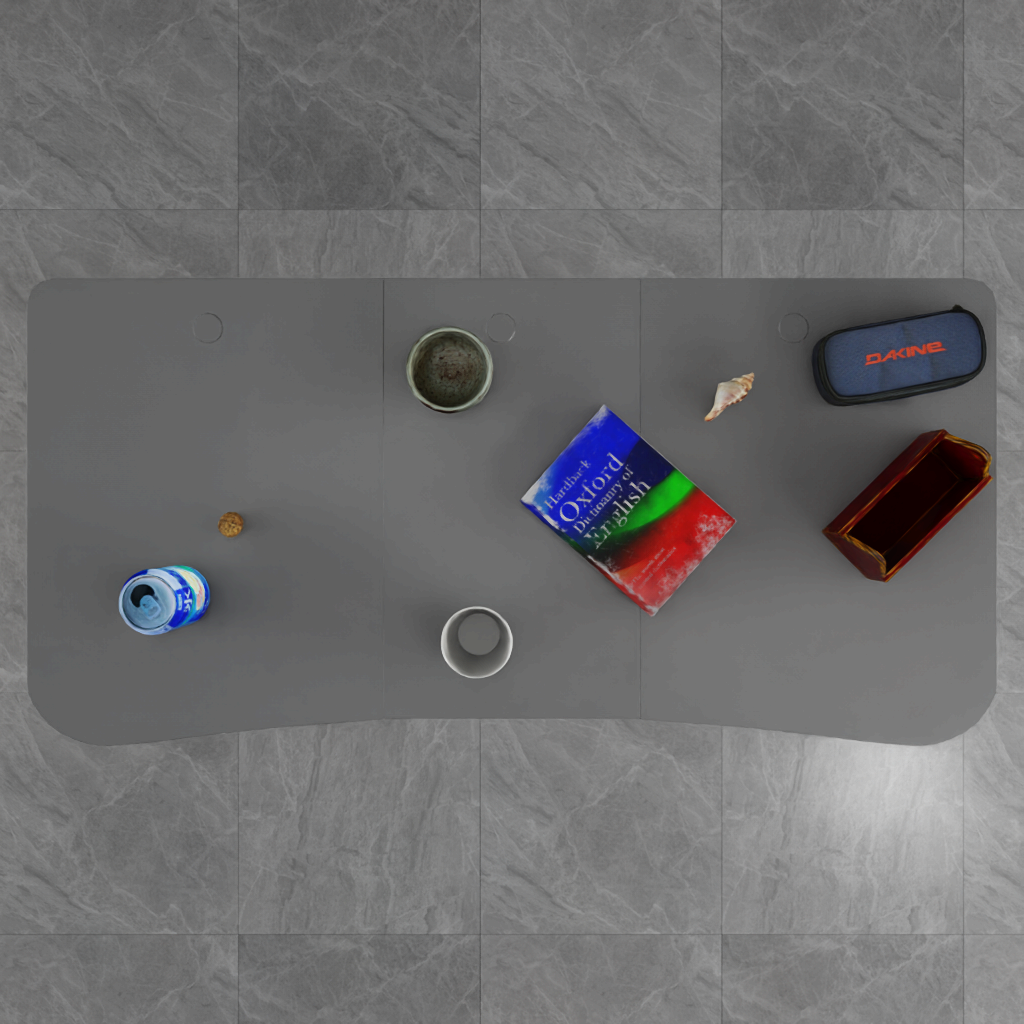}{0.5030}{31/32}
    &
    \QualCell{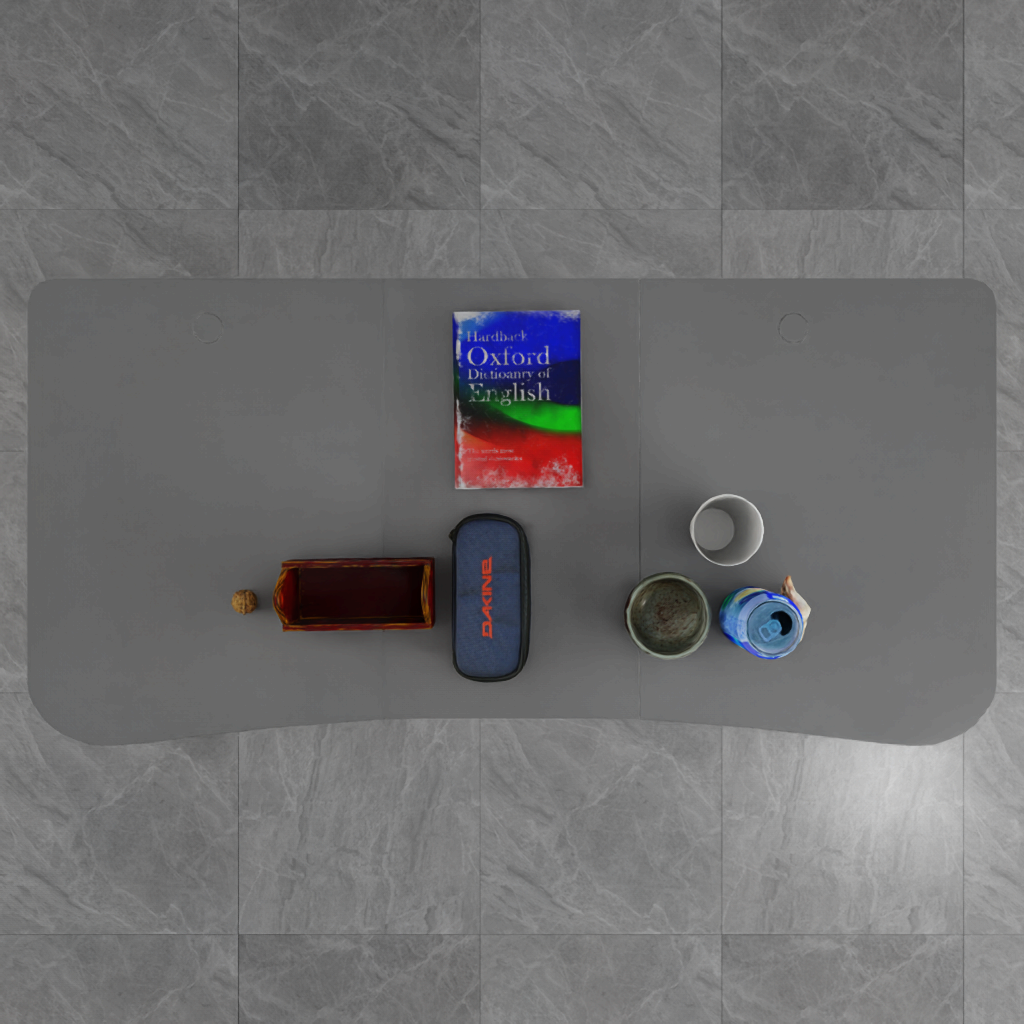}{0.5653}{3/32}
    &
    \QualCell{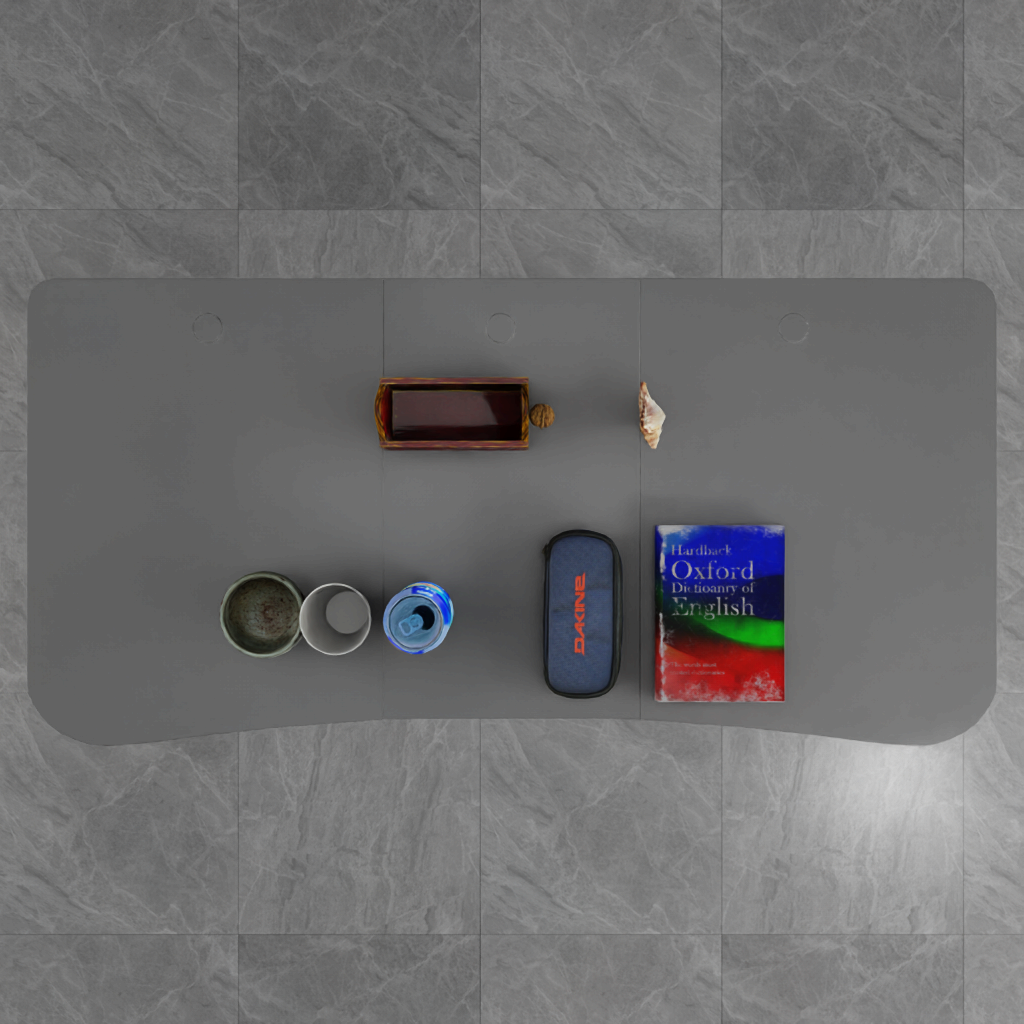}{0.6148}{16/32}
    &
    \QualCell{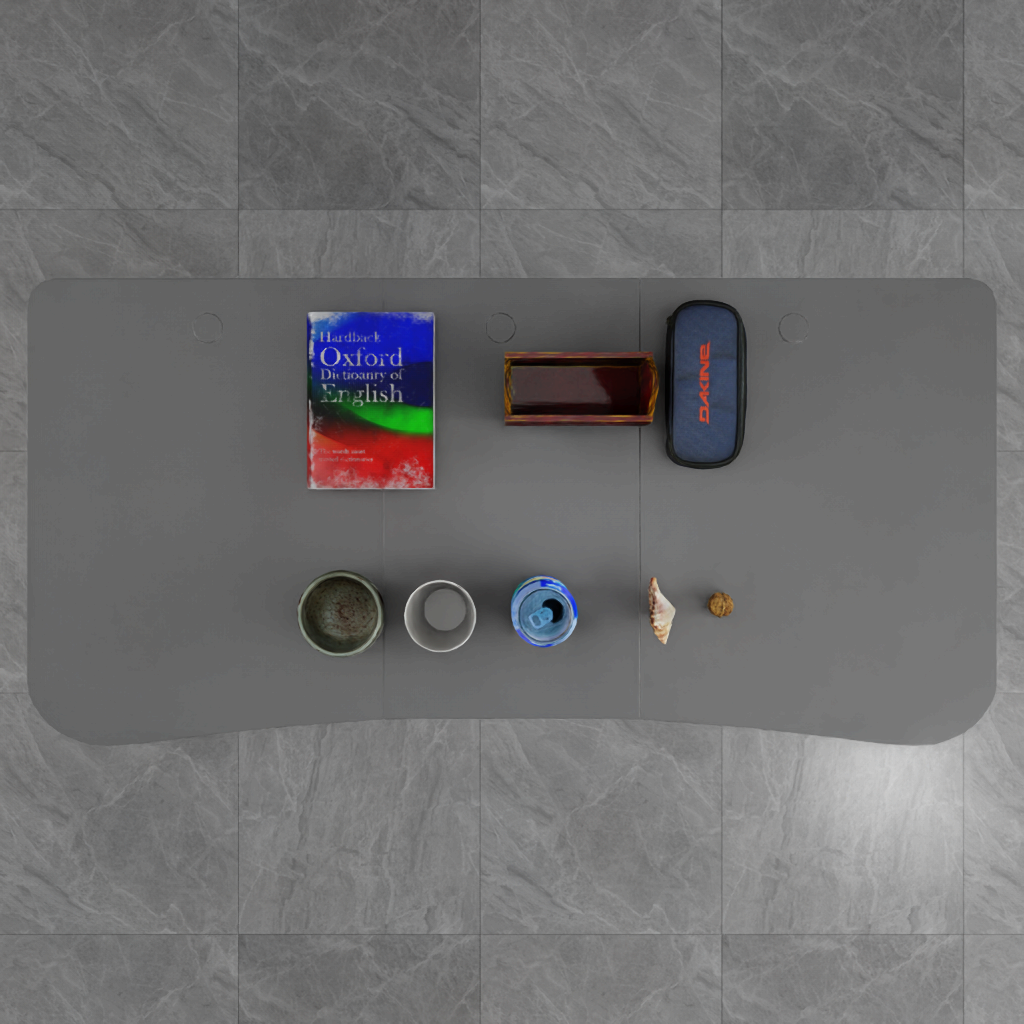}{0.6981}{16/32}
    &
    \QualCell{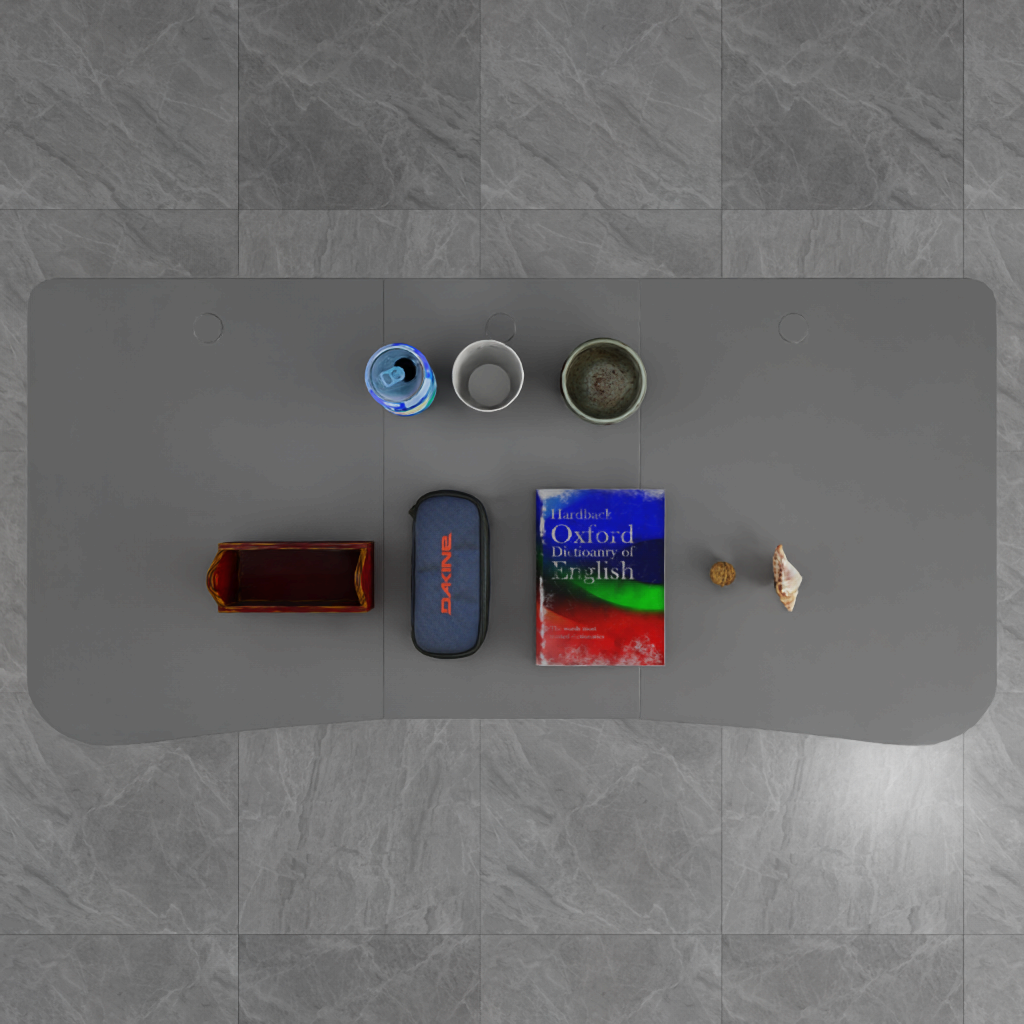}
        {\mathbf{0.7094}}{\textbf{32/32}}

    \\[3pt]

    % =====================================================
    % GTL - N = 10
    % =====================================================

    \QualLabel{GTL}{10}
    &
    \QualCell{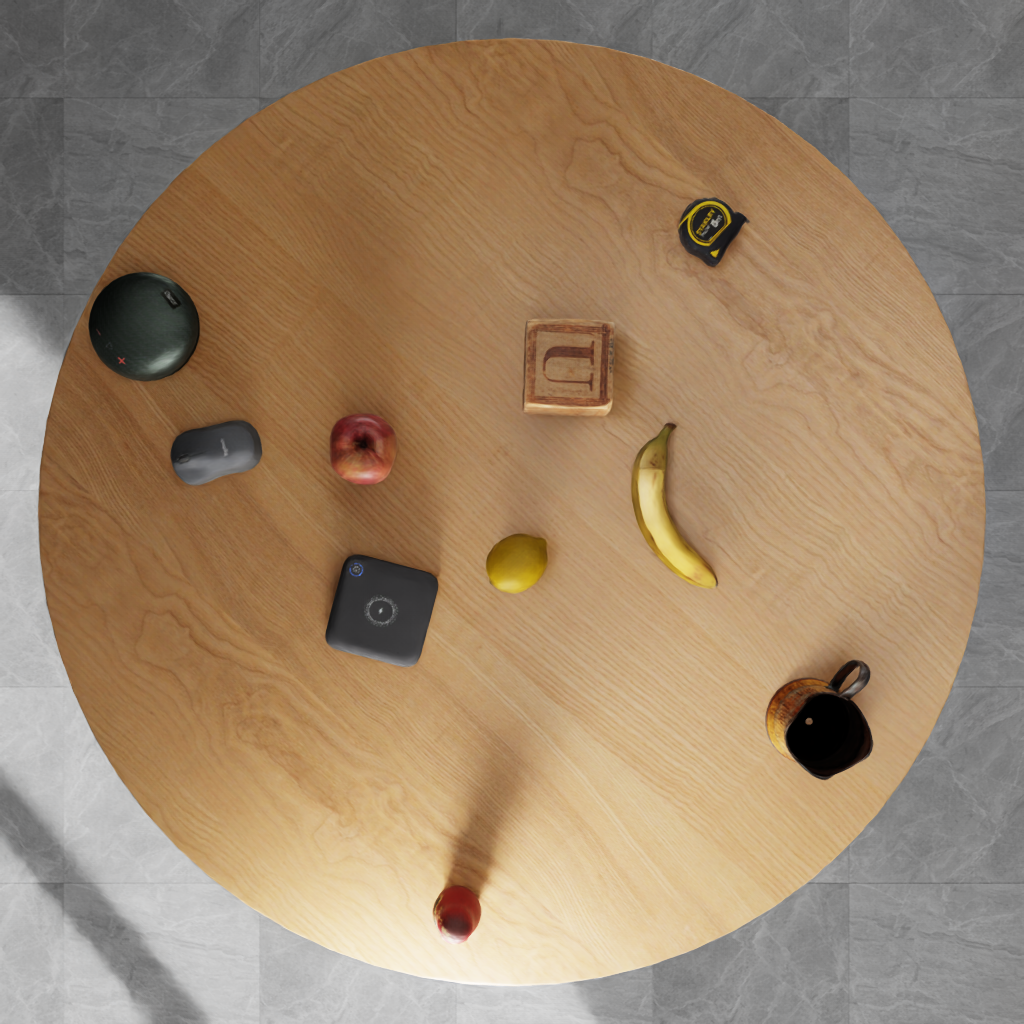}{0.4725}{30/32}
    &
    \QualCell{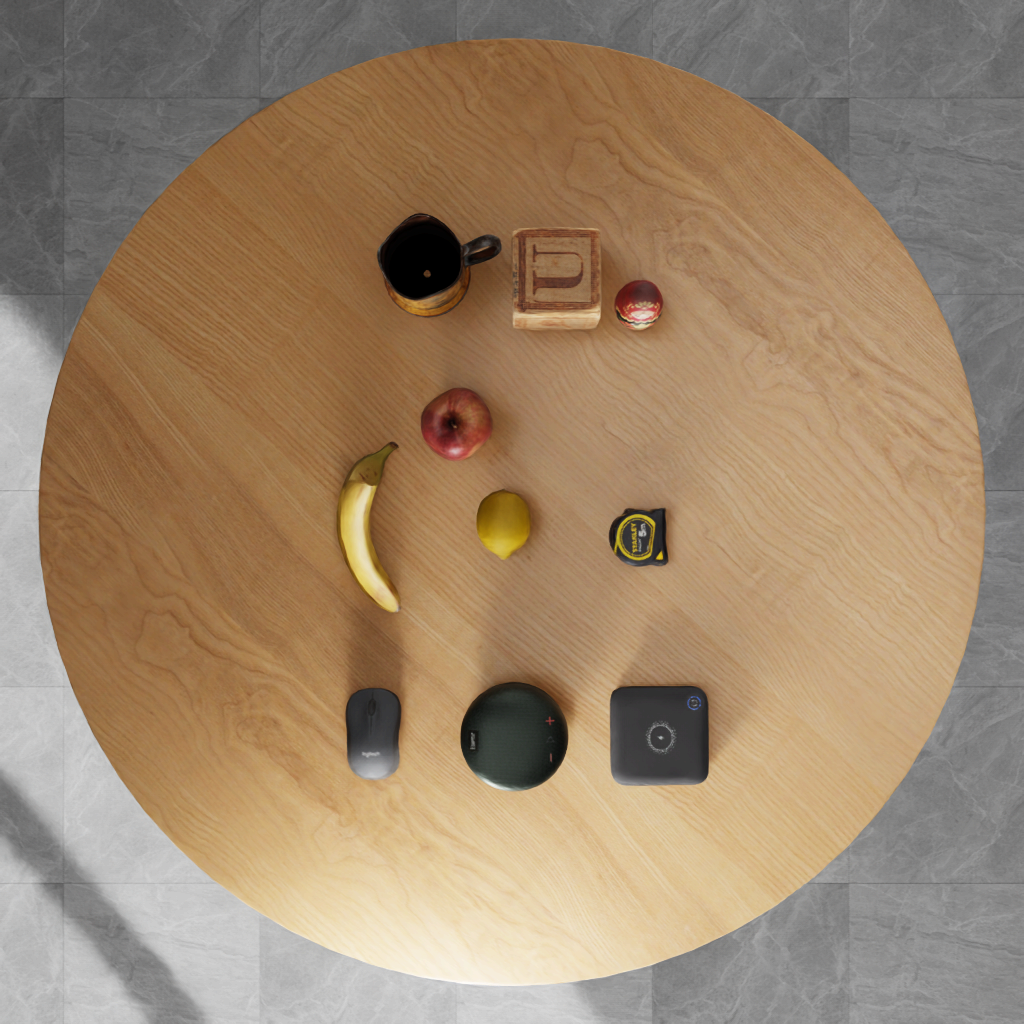}{0.5843}{2/32}
    &
    \QualCell{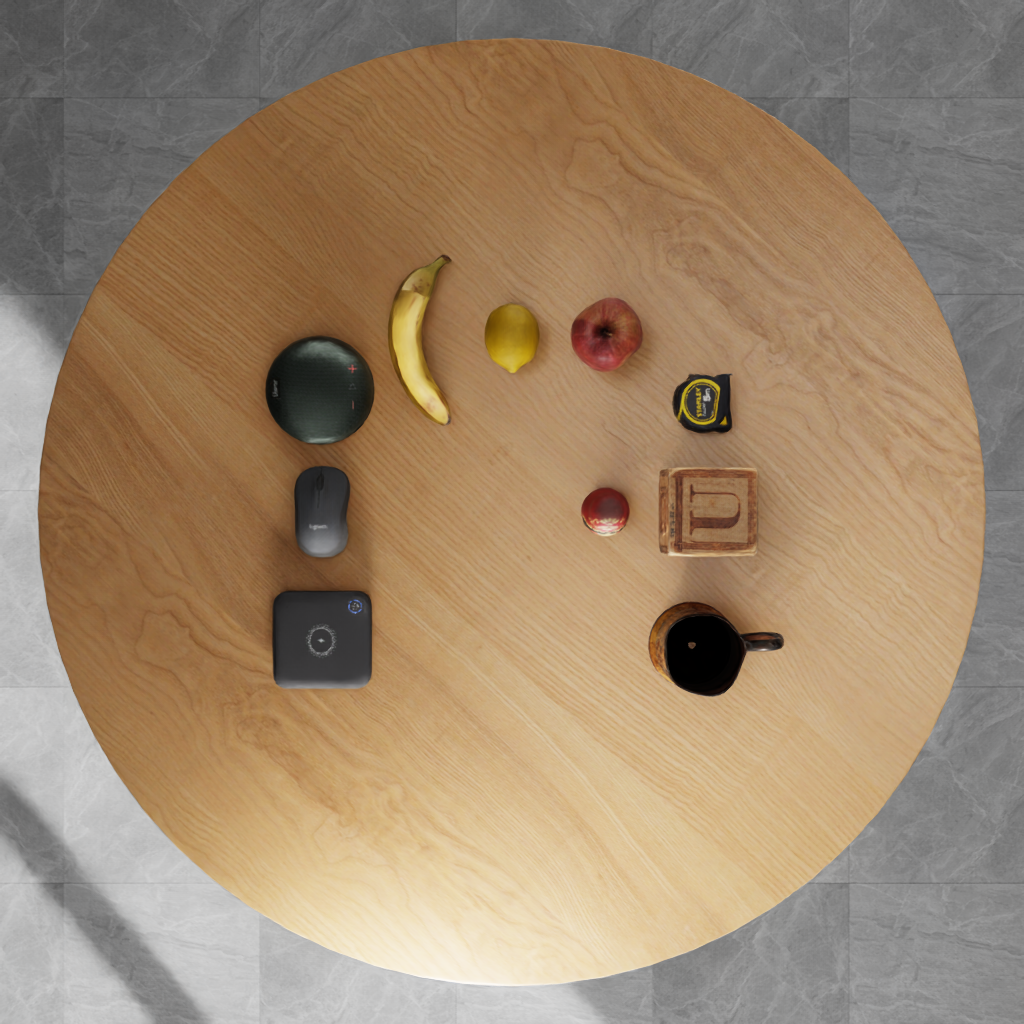}{0.5639}{8/32}
    &
    \QualCell{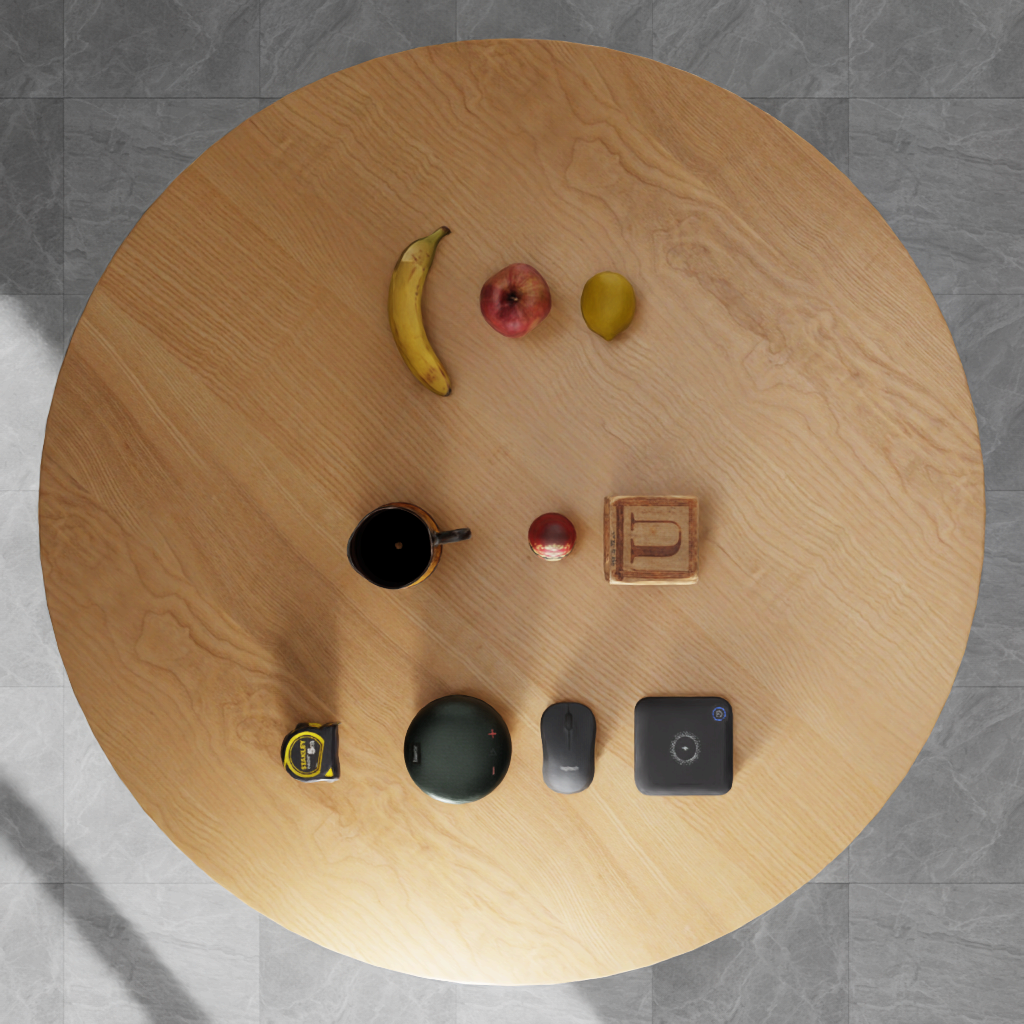}{0.7513}{26/32}
    &
    \QualCell{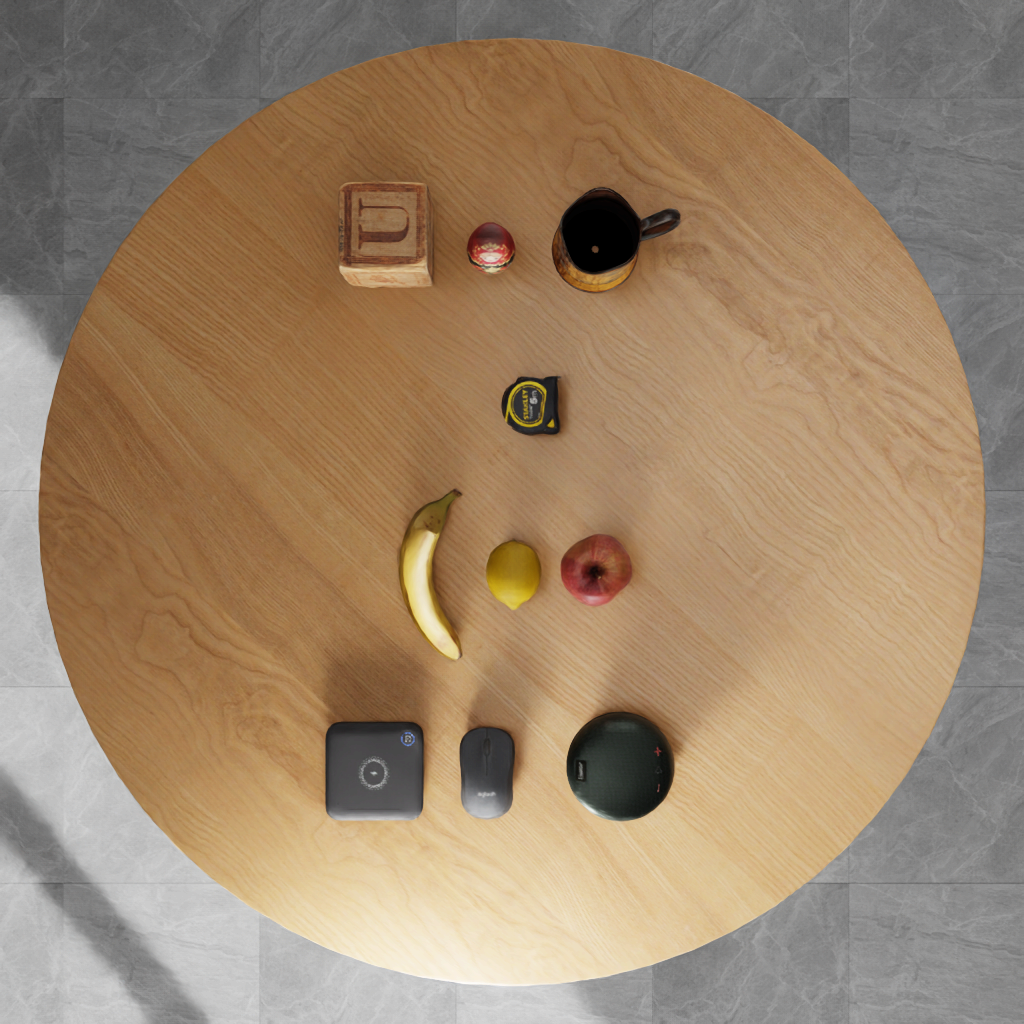}
        {\mathbf{0.7660}}{\textbf{31/32}}

    \\[3pt]

    % =====================================================
    % WTL - N = 11
    % =====================================================

    \QualLabel{WTL}{11}
    &
    \QualCell{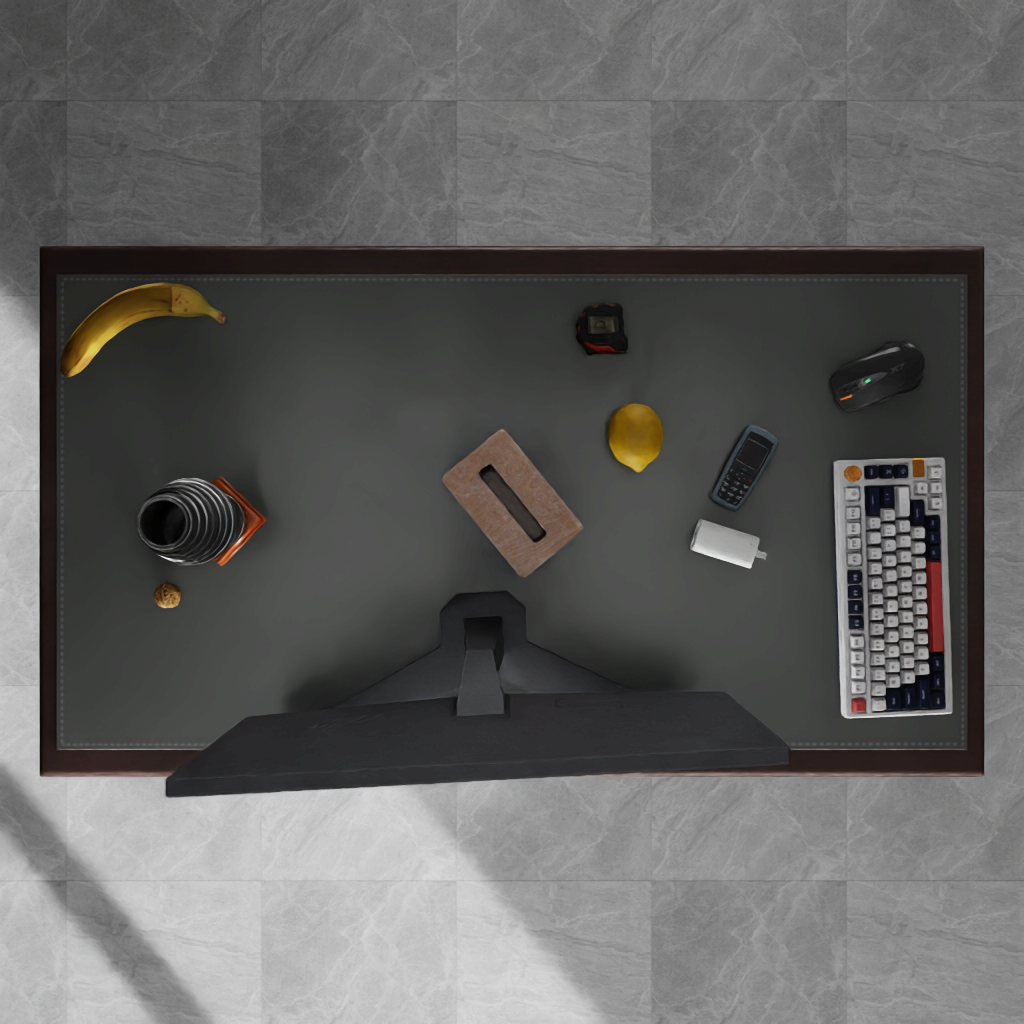}{0.3563}{28/32}
    &
    \QualCell{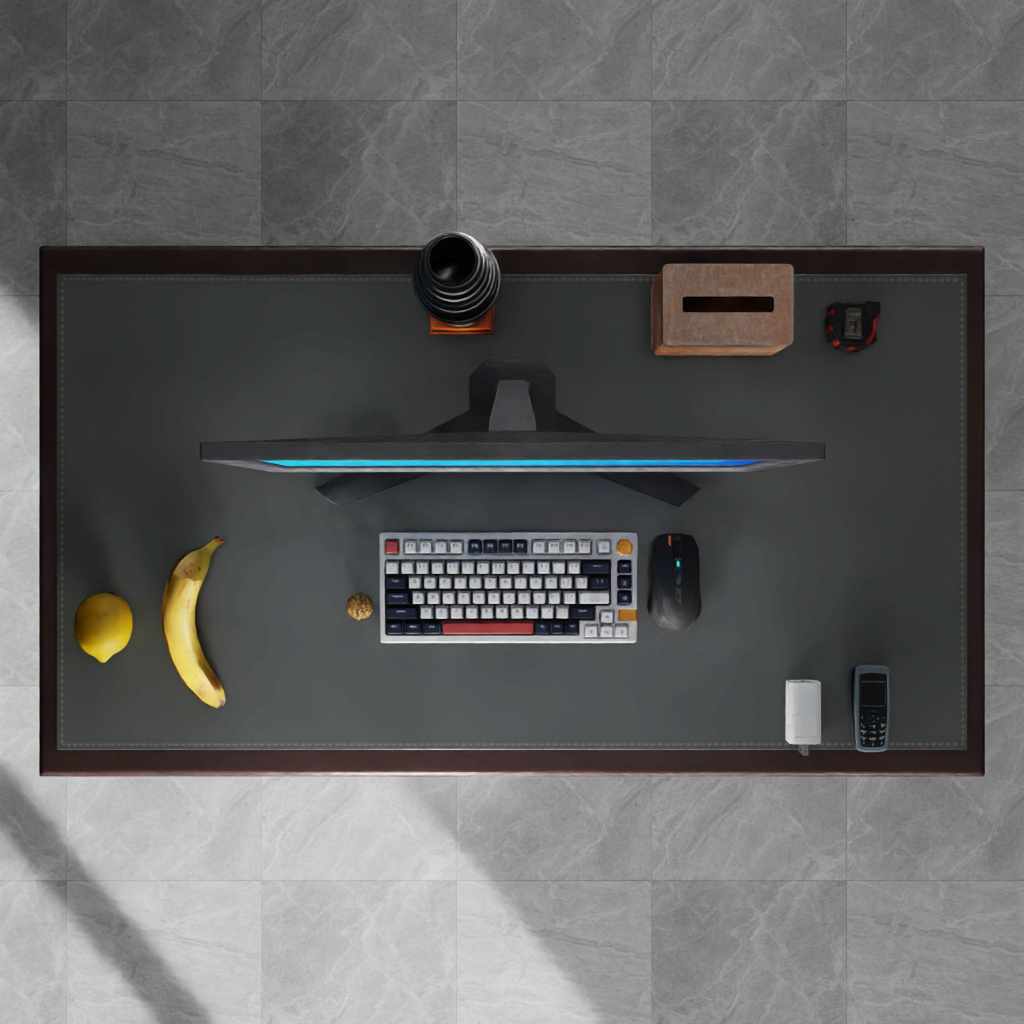}{0.6481}{2/32}
    &
    \QualCell{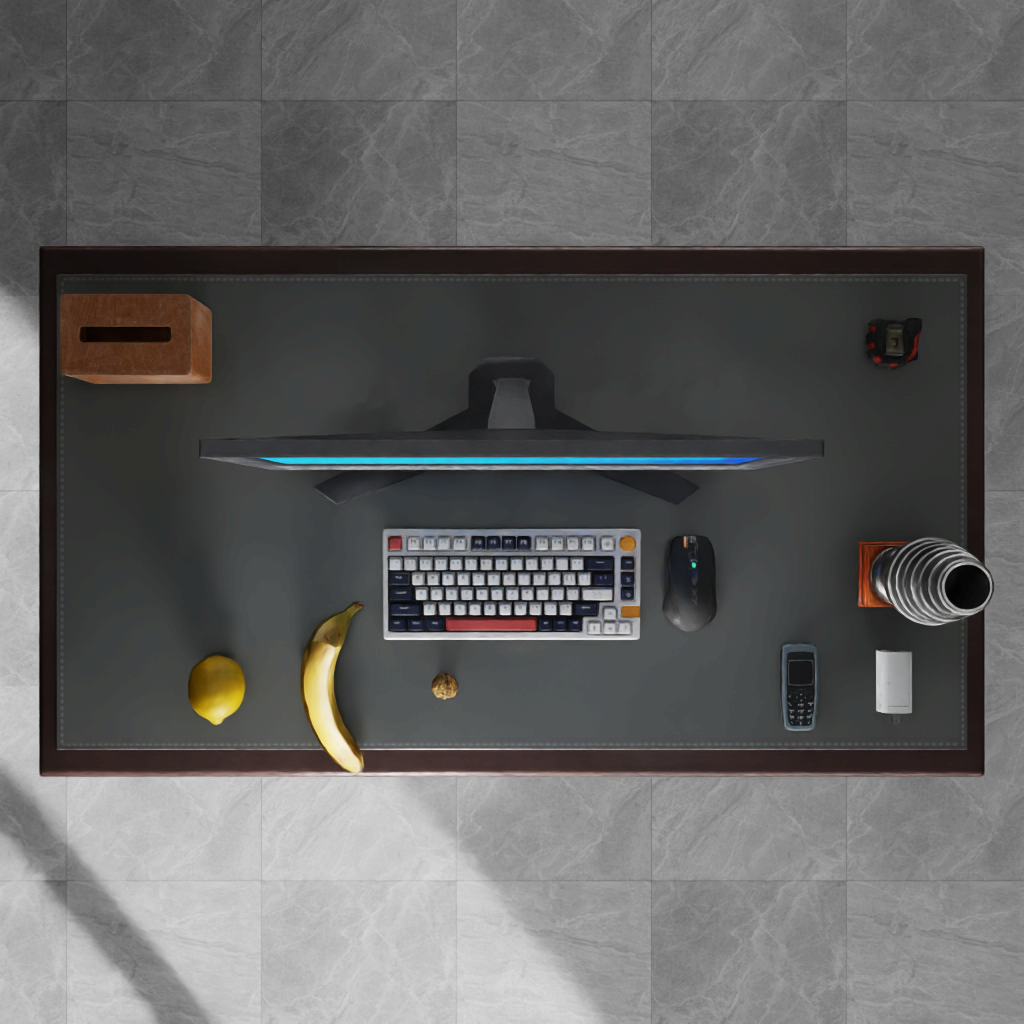}{0.6478}{12/32}
    &
    \QualCell{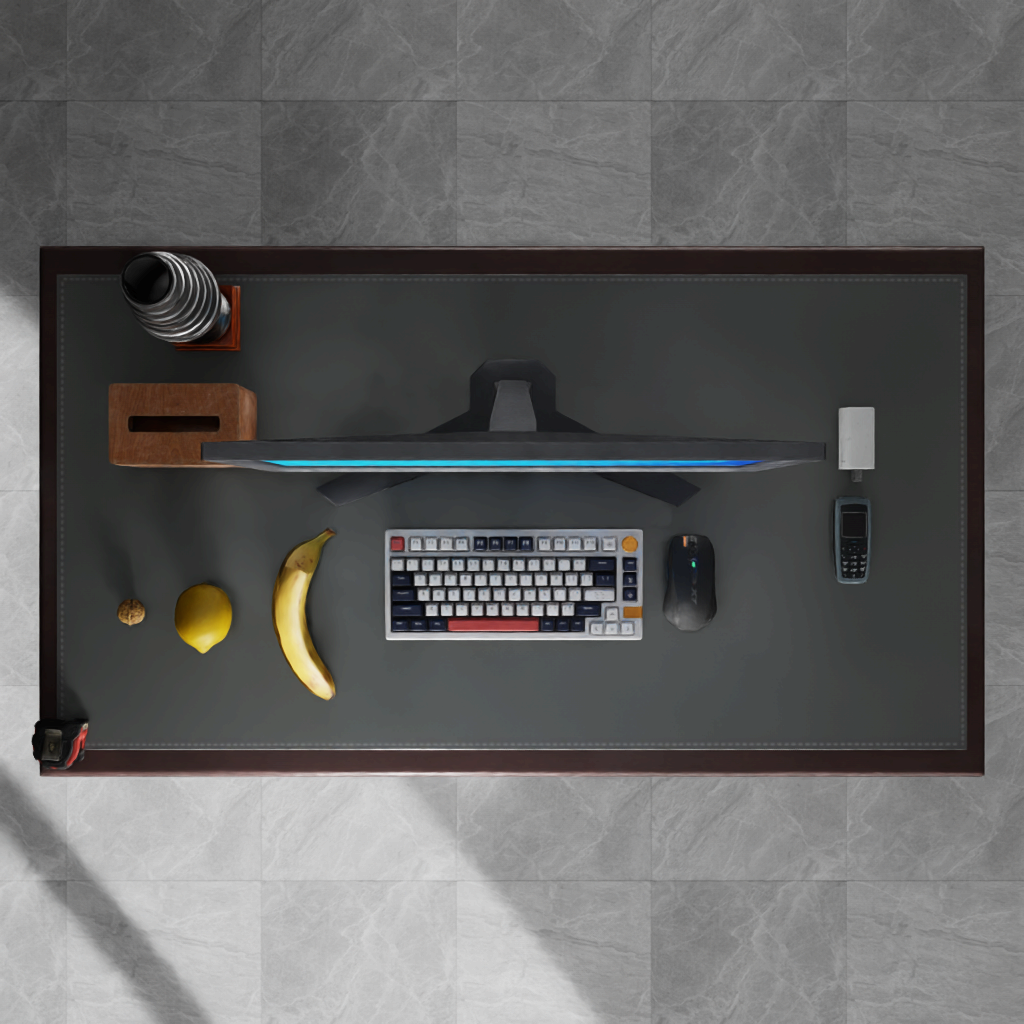}{0.6744}{9/32}
    &
    \QualCell{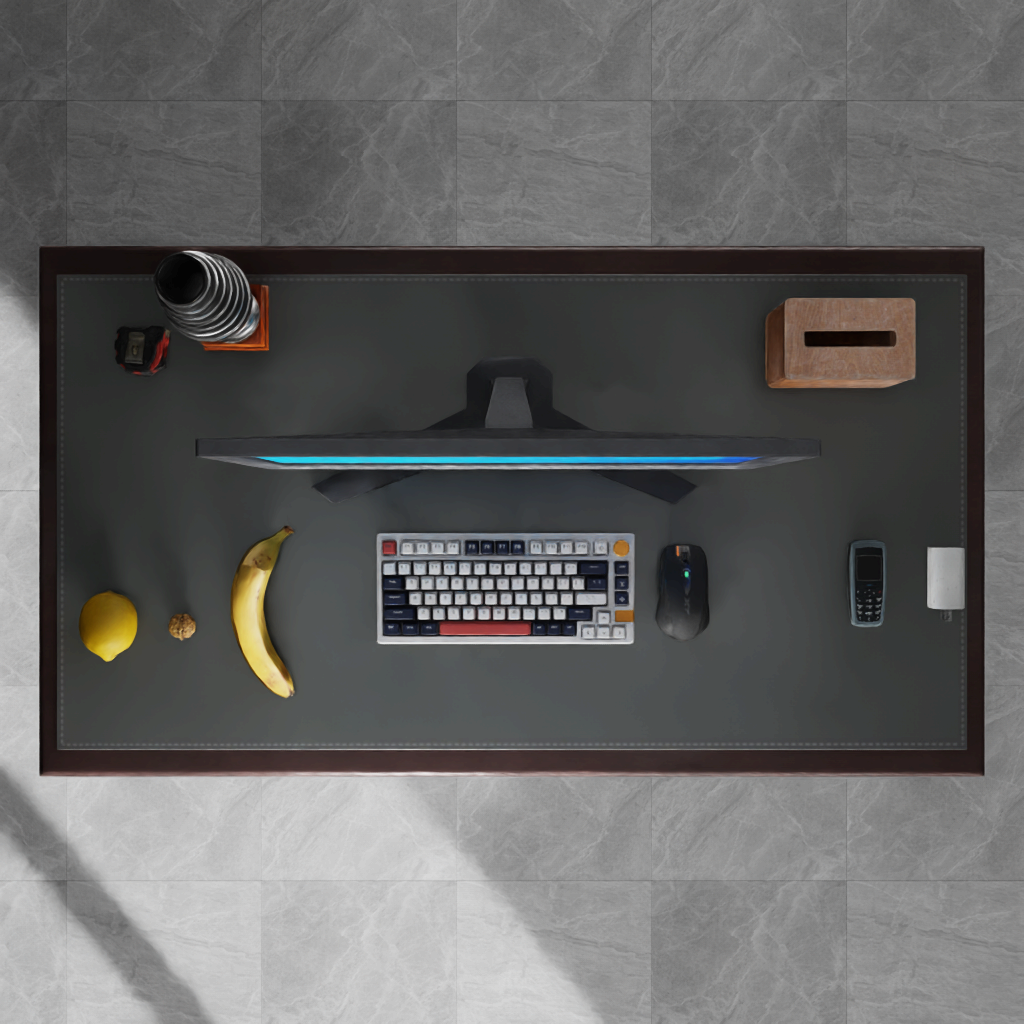}
        {\mathbf{0.6936}}{\textbf{32/32}}

    \\[3pt]

    % =====================================================
    % WTL - N = 13
    % =====================================================

    \QualLabel{WTL}{13}
    &
    \QualCell{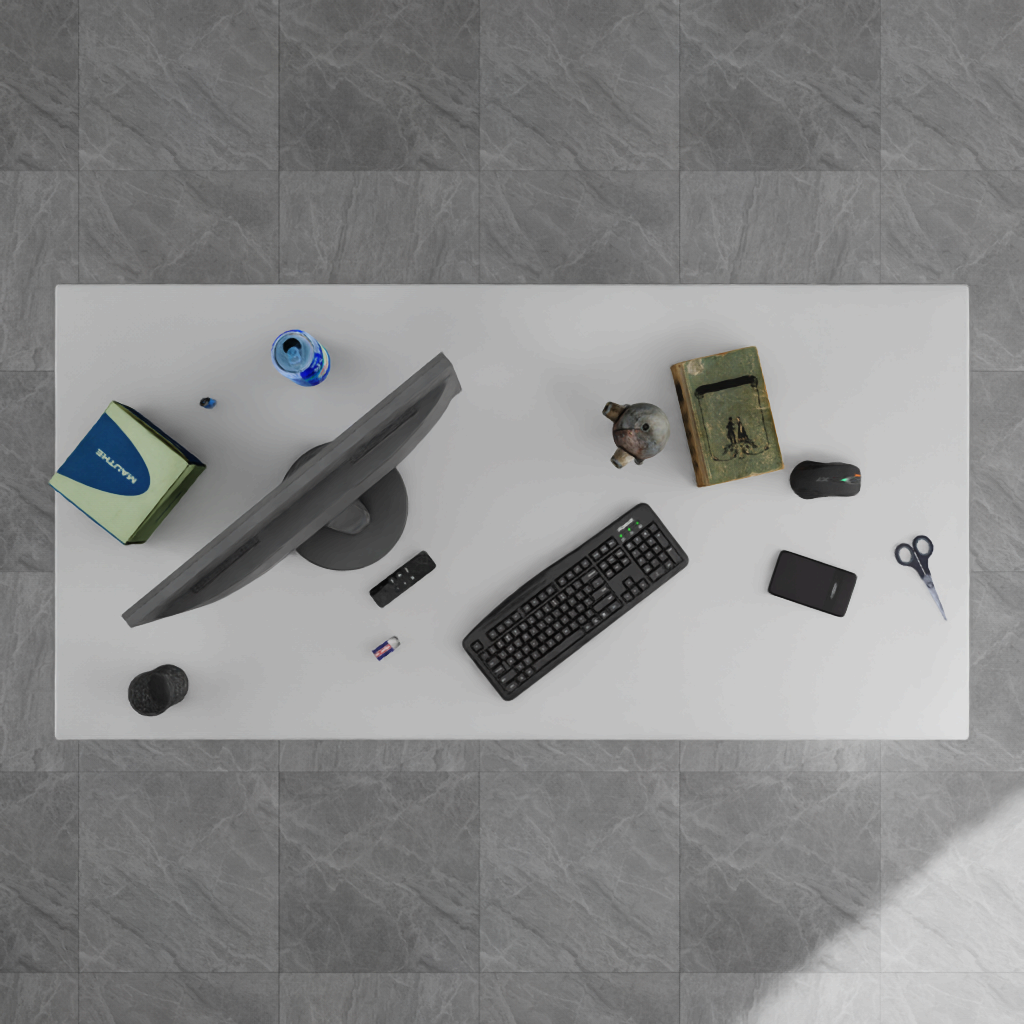}
        {0.3450}{\textbf{29/32}}
    &
    \QualCell{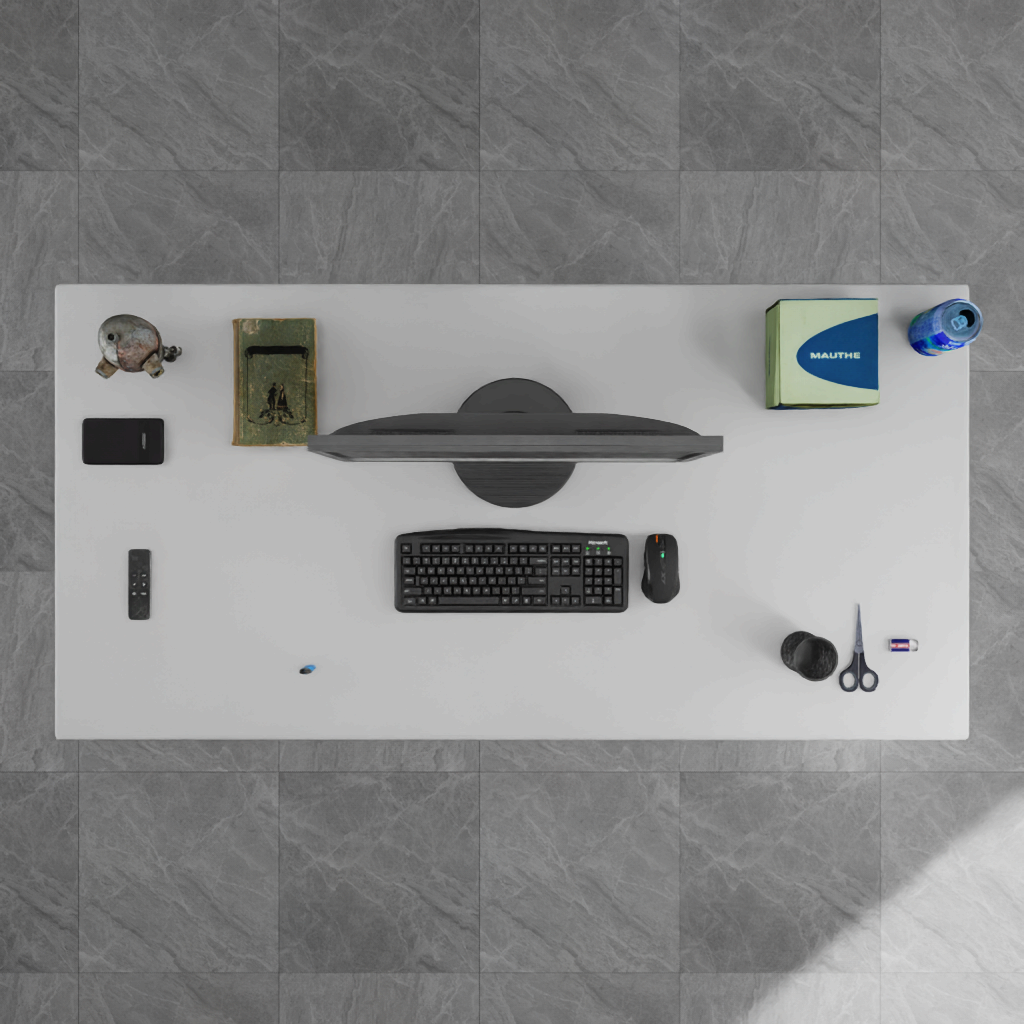}{0.6777}{1/32}
    &
    \QualCell{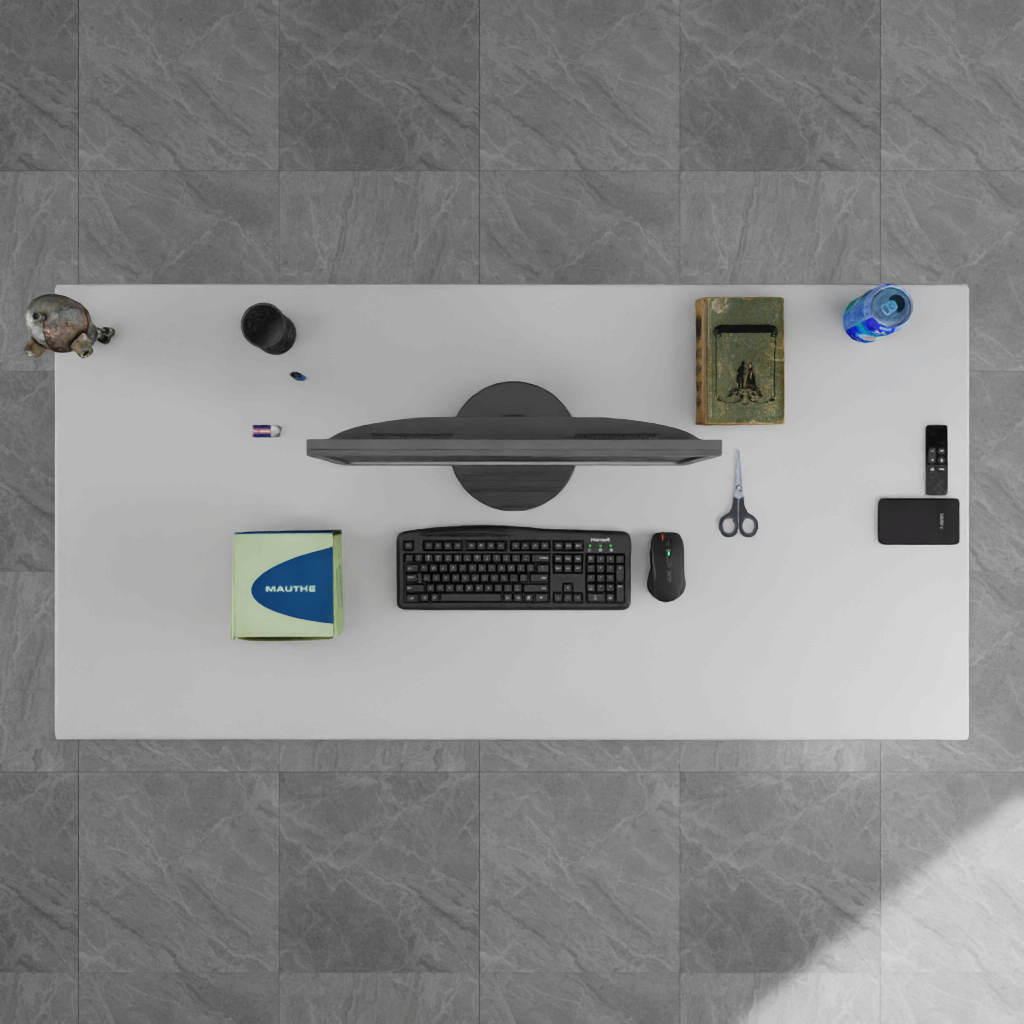}{0.5241}{10/32}
    &
    \QualCell{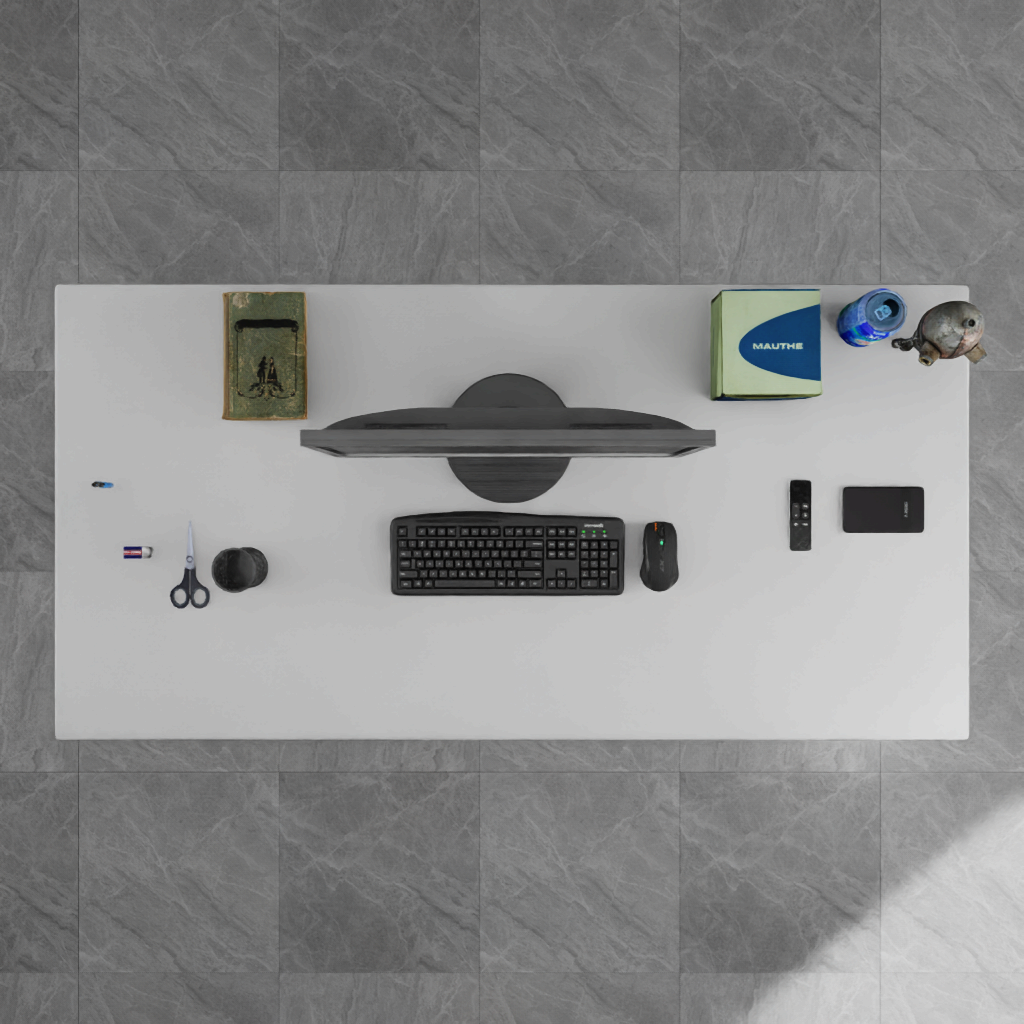}{0.6593}{3/32}
    &
    \QualCell{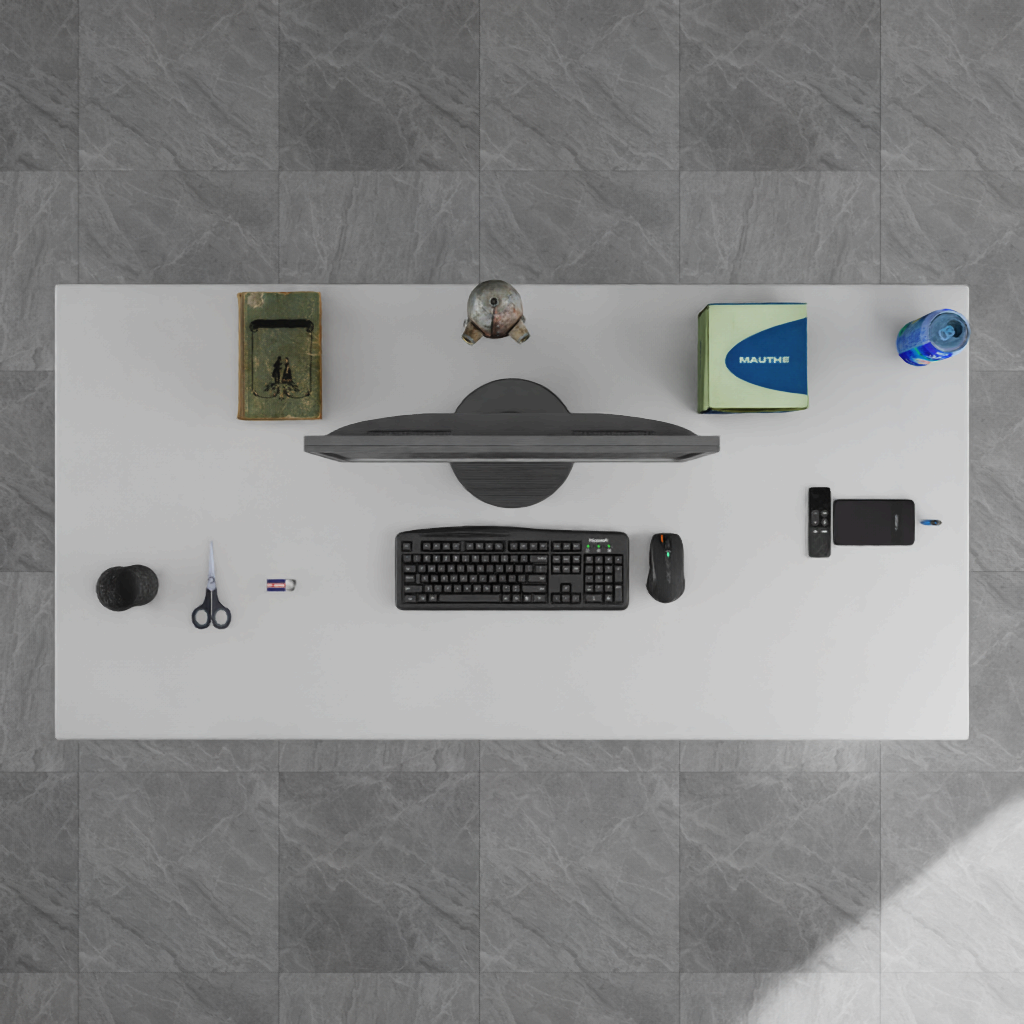}
        {\mathbf{0.7027}}{16/32}

    \\

    \end{tabular}

    \endgroup

    % =====================================================
    % Caption
    % =====================================================

    \caption{
    \textbf{Qualitative comparison on GTL and WTL.}
    The first two rows show GTL examples,
    while the last two rows show WTL examples.
    Each row corresponds to an identical object
    set and tabletop condition across five methods.
    $N$ denotes the number of objects,
    $\Psi$ denotes the tidiness score of the
    displayed layout, and Valid indicates
    the number of geometrically valid candidates
    out of 32.
    Bold values indicate the highest tidiness
    score or the largest number of valid
    candidates in each row.
    }

    \label{fig:qualitative_comparison}

\end{figure*}

\begin{table*}[t]
\centering

\caption{
Comparison of layout generation methods on GTL and WTL.
All methods generate 32 candidates per condition. GVR, CR, BVR, and FCR@32 are reported in \%.
PD denotes mean penetration depth (cm),
and $\Psi$ denotes mean tidiness score.
\textbf{Bold} indicates the best performance
for each metric.
}

\label{tab:comparison}

\small
\setlength{\tabcolsep}{4.5pt}

\begin{tabular}{llrrrrrr}
\toprule

Dataset & Method
& $\Psi\uparrow$
& GVR$\uparrow$
& CR$\downarrow$
& PD$\downarrow$
& BVR$\downarrow$
& FCR@32$\uparrow$ \\

\midrule

\multirow{5}{*}{GTL}
& SceneScore
& 0.4027
& 95.24
& 2.30
& 0.0485
& 2.58
& 100.00 \\

& StructDiffusion
& 0.6763
& 10.46
& 88.71
& 3.3445
& 11.92
& 73.86 \\

& LayoutGPT
& 0.6253
& 24.43
& 72.92
& 3.0616
& 10.97
& 73.44 \\

& LEGO-Net
& 0.6830
& 45.44
& 48.69
& 2.5646
& 21.15
& 98.34 \\

& \textbf{TidyFlow (Ours)}
& \textbf{0.7192}
& \textbf{88.91}
& \textbf{10.09}
& \textbf{2.8127}
& \textbf{3.72}
& \textbf{100.00} \\

\midrule

\multirow{5}{*}{WTL}
& SceneScore
& 0.3896
& 81.51
& 11.82
& 0.2692
& 12.25
& 100.00 \\

& StructDiffusion
& 0.6792
& 5.50
& 93.54
& 3.7342
& 24.53
& 35.31 \\

& LayoutGPT
& 0.6627
& 18.20
& 75.56
& 3.2747
& 37.93
& 44.27 \\

& LEGO-Net
& 0.6726
& 21.73
& 70.16
& 2.4276
& 41.65
& 77.73 \\

& \textbf{TidyFlow (Ours)}
& \textbf{0.6943}
& \textbf{57.91}
& \textbf{41.53}
& \textbf{2.7616}
& \textbf{6.23}
& \textbf{97.36} \\

\bottomrule
\end{tabular}

\end{table*}

\subsection{Agreement with Human Preferences}
\label{sec:human_preference}
Computed visual-aesthetic measures may differ from human judgments~\cite{zen2016aesthetics}; multidimensional design preferences exhibit rater variability~\cite{zhu2026taste}, while rankings and attribute-level ratings provide complementary evidence~\cite{zhao2026preferences}. We therefore conducted an anonymous survey, collecting 54 responses and retaining 47 valid responses after excluding seven. Participants rated two layouts from each of ten conditions on six attributes (four-point scale), and ranked six layouts from each of ten disjoint conditions. Both top-down and oblique views were provided.

We compare algorithm-induced rankings against human preferences using tie-adjusted Kendall's $\tau_b$, Spearman's $\rho$ with average ranks, pairwise agreement, and Top-3 overlap. Across the 47 valid responses, the ranking statistics are \textbf{0.4667}, \textbf{0.5543}, \textbf{73.33\%}, and \textbf{70.00\%}, respectively, indicating positive agreement for the evaluated responses. For attribute-level agreement across these 47 valid responses, only pairs with non-tied human and algorithm score differences are included (\cref{tab:human_dimensions}). Alignment (\textbf{87.50\%}), group organization (\textbf{85.71\%}), and orientation (\textbf{83.27\%}) achieve relatively higher agreement, whereas centering reaches \textbf{50.00\%}, indicating unequal alignment across dimensions.

This study collected subjective evaluations of tabletop arrangements through an anonymous questionnaire. All participants voluntarily participated after being informed of the study's purpose and provided informed consent before completing the questionnaire. 

\begin{table}[t]
\centering
\caption{Direction agreement with human attribute ratings; tied pairs are excluded.}
\label{tab:human_dimensions}
\small\setlength{\tabcolsep}{5pt}
\begin{tabular}{lr}
\toprule
Dimension & Agreement$\uparrow$ \\
\midrule
Compactness & 66.67\% \\
Alignment & \textbf{87.50\%} \\
Orientation & 83.27\% \\
Centering & 50.00\% \\
Clearance & 62.50\% \\
Group Organization & 85.71\% \\
\bottomrule
\end{tabular}
\end{table}

\subsection{Ablation and Training Stability}
\label{sec:ablation_robustness}

We study the effects of iterative refinement and Refiner
components under controlled evaluation protocols.

\subsubsection{Effect of Shared-Weight Refinement Iterations}
\label{sec:iteration_ablation}

To isolate inference-time iteration effects, we fix the trained
networks, indexed Raw Layout candidates, and safety-filtering rule,
varying only the number of shared-weight Refiner calls.
As shown in \cref{tab:iteration}, R1 accounts for \textbf{80.6\%}
and \textbf{64.1\%} of the total GVR improvement on GTL and WTL,
respectively. R3 adds \textbf{2.15} and \textbf{4.53} percentage
points over R2, while further reducing violations and penetration.
The mean functional-group centroid displacement after R3 is
\textbf{0.93/0.63\,cm} on GTL/WTL, consistent with local correction.
Overall, these results demonstrate that shared-weight iterative refinement progressively improves geometric validity on both datasets, with the largest gains achieved in the first iteration and further improvements obtained through subsequent refinements.

\begin{table}[t]
\centering
\caption{
Fixed-candidate iteration ablation on GTL and WTL.
Rates are in \%; $D_{\Sigma}$ denotes the mean total pairwise
penetration depth per candidate, and $\Delta_g$
denotes the mean functional-group centroid
displacement, both measured in cm.
}
\label{tab:iteration}

\begingroup
\footnotesize
\setlength{\tabcolsep}{1.5pt}
\renewcommand{\arraystretch}{0.95}

\resizebox{\columnwidth}{!}{%
\begin{tabular}{@{}*{8}{c}@{}}
\toprule
Dataset & Stage
& GVR$\uparrow$
& CR$\downarrow$
& BVR$\downarrow$
& $D_{\Sigma}\downarrow$
& FCR@32$\uparrow$
& $\Delta_g$ \\
\midrule

\multirow{4}{*}{GTL}
& Raw Layout
& 23.21 & 75.19 & 10.22 & 4.38 & 95.02 & 0.00 \\

& R1
& 76.18 & 22.51 & 4.72 & 0.95 & \textbf{100.00} & 0.57 \\

& R2
& 86.76 & 12.16 & 3.92 & 0.56 & \textbf{100.00} & 0.76 \\

& R3
& \textbf{88.91}
& \textbf{10.09}
& \textbf{3.72}
& \textbf{0.48}
& \textbf{100.00}
& 0.93 \\

\midrule

\multirow{4}{*}{WTL}
& Raw Layout
& 10.58 & 87.79 & 24.67 & 8.87 & 52.44 & 0.00 \\

& R1
& 40.91 & 58.42 & 7.21 & 4.22 & 91.30 & 0.49 \\

& R2
& 53.38 & 46.06 & 6.35 & 3.26 & 96.05 & 0.59 \\

& R3
& \textbf{57.91}
& \textbf{41.53}
& \textbf{6.23}
& \textbf{3.02}
& \textbf{97.36}
& 0.63 \\

\bottomrule
\end{tabular}%
}
\endgroup
\end{table}

\subsubsection{Refiner Component Analysis}
\label{sec:component_ablation}

We conduct five Refiner ablations targeting corruption design,
edge conditioning, and loss objectives: random corruption,
severity-matched corruption without topology scheduling,
removal of edge conditioning, and separate removal of repair or
preservation losses. Retrained variants share the same clean
references, target-pose supervision, training budget,
checkpoint-selection protocol, and fixed Flow candidates
with identical safety filtering.
As shown in \cref{tab:refiner_components}, the full model
achieves the highest GVR on both datasets, with random
corruption and removal of edge conditioning causing substantial
performance degradation. These results support topology-aware
data construction, explicit pairwise relations, and complementary
repair and preservation objectives.

\begin{table}[t]
\centering
\caption{
Refiner component analysis on GTL and WTL. PD is in cm; other metrics are percentages.
}
\label{tab:refiner_components}

\begingroup
\footnotesize
\setlength{\tabcolsep}{2pt}
\renewcommand{\arraystretch}{1.05}

\begin{tabular}{@{}llrrrrr@{}}
\toprule
Dataset
& Variant
& GVR$\uparrow$
& CR$\downarrow$
& PD$\downarrow$
& BVR$\downarrow$
& FCR@32$\uparrow$ \\
\midrule

\multirow{6}{*}{GTL}
& Full
& \textbf{88.95}
& \textbf{10.50}
& \textbf{2.609}
& \textbf{2.63}
& \textbf{100.00} \\

& w/o repair
& 88.29
& 11.20
& 2.681
& 2.81
& \textbf{100.00} \\

& w/o preserve
& 85.96
& 13.62
& 2.892
& 2.81
& \textbf{100.00} \\

& w/o topology
& 84.69
& 14.54
& 2.677
& 3.25
& \textbf{100.00} \\

& w/o edge
& 75.83
& 23.82
& 2.838
& 3.42
& \textbf{100.00} \\

& Random
& 75.27
& 22.50
& 2.831
& 5.24
& 99.59 \\

\midrule

\multirow{6}{*}{WTL}
& Full$^{\dagger}$
& \textbf{58.40}
& \textbf{40.99}
& \textbf{2.696}
& \textbf{5.82}
& \textbf{98.02} \\

& w/o repair
& 53.92
& 45.33
& 2.762
& 6.86
& 96.18 \\

& w/o preserve
& 54.70
& 44.95
& 2.827
& 6.23
& 96.18 \\

& w/o topology
& 51.99
& 47.58
& 2.702
& 5.99
& 95.78 \\

& w/o edge
& 55.91
& 43.53
& 3.372
& 8.91
& 97.36 \\

& Random
& 44.84
& 53.62
& 2.909
& 10.85
& 93.68 \\

\bottomrule
\end{tabular}

\endgroup
\end{table}

%% file: sec/conclusion.tex
\section{Conclusion}
\label{sec:conclusion}

In this work, we presented TidyFlow, a geometry-aware generative framework for tidy tabletop arrangement generation.
By learning a multimodal distribution of functionally organized layouts and leveraging collision patterns to guide geometric refinement, TidyFlow addresses both functional arrangement generation and geometric feasibility. Experiments on the GTL and WTL datasets demonstrate that our method substantially improves geometric validity, achieving validity rates of 88.91\% and 57.91\%, respectively.
Ablation studies validate the effectiveness of iterative refinement, while Refiner component analyses demonstrate the contributions of its key design choices.
Human preference evaluations further indicate that the proposed six-dimensional tidiness scoring function captures meaningful aspects of subjective tabletop organization.

Despite these improvements, TidyFlow still faces challenges in densely populated scenes with complex geometric constraints, where geometric validity decreases as the number of objects increases.
Future work will investigate more effective joint modeling of layout generation and geometric constraints, and evaluate the practical applicability of the framework through real-world robotic manipulation.

%% file: sec/X_suppl.tex
\section{Additional Method Details}
\label{sec:suppl_method}

We provide the formulation underlying the Refiner and the
six-dimensional tidiness evaluation used in TidyFlow.
The Refiner combines supervised geometric correction with
preservation of already valid arrangements, whereas the
predefined tidiness score evaluates the organization of
geometrically feasible layouts.

\subsection{Repair-and-Preservation Objective}
\label{sec:suppl_refiner_loss}

Rather than regenerating the entire arrangement, the Refiner
predicts local pose corrections to a Raw Layout
$X^{\mathrm{raw}}$, conditioned on the object and tabletop
information $C$. Given its corresponding clean Reference
Layout $X^\star$, the normalized prediction and target
increments are
\begin{equation}
\begin{aligned}
\hat X &= \mathcal N\!\left(X^{\mathrm{raw}}+
                    f_\omega(X^{\mathrm{raw}},C)\right),\\
\Delta X &= \hat X-X^{\mathrm{raw}},\qquad
\Delta X^\star=X^\star-X^{\mathrm{raw}}.
\end{aligned}
\label{eq:suppl_pred_increments}
\end{equation}
Here $\mathcal N$ renormalizes the sine--cosine yaw components.
We distinguish repair targets $M$, comprising objects involved
in input collisions or annotated boundary violations, from
the remaining valid objects $U$. This separation allows the
model to correct geometric conflicts without unnecessarily
moving objects that are already well placed.

For repair targets, $\mathcal L_{\mathrm{target}}$ supervises
the normalized four-dimensional pose increments. For the
unaffected objects, $\mathcal L_{\mathrm{stay}}$ penalizes
translation, while $\mathcal L_{\mathrm{yaw}}$ preserves
orientation with a symmetry-aware cosine penalty:
\begin{align}
\mathcal L_{\mathrm{target}} &=
\frac{\sum_{i\in M}\|\Delta X_i-\Delta X_i^\star\|_2^2}
     {4\max(|M|,1)},
\label{eq:suppl_l_target}\\
\mathcal L_{\mathrm{stay}} &=
\frac{\sum_{i\in U}\|\hat p_i-p_i^{\mathrm{raw}}\|_2^2}
     {2\max(|U|,1)}.
\label{eq:suppl_l_stay}
\end{align}
The yaw term averages
$1-\cos\!\bigl(k_i(\hat\theta_i-\theta_i^{\mathrm{raw}})\bigr)$
over applicable non-target objects, with $k_i=2$ for
$180^\circ$ rotational symmetry and $k_i=1$ otherwise;
continuously rotation-symmetric objects are excluded.
The positional quantities are normalized by the tabletop
scale $L=\tfrac12\max(W_T,D_T)$.

Collision supervision distinguishes pairs that already
overlap in $X^{\mathrm{raw}}$ from those that are initially
collision-free. Let $g_{ij}(X)$ be the signed SAT separation
gap (negative for overlap), and let
$\mathcal C^{\mathrm{raw}}$ and $\mathcal S^{\mathrm{raw}}$
denote the corresponding colliding and non-colliding pair
sets. We use
\begin{equation}
\ell_{ij}=\pospart{m_c-g_{ij}(\hat X)}^2,
\qquad m_c=0.0025/L,
\label{eq:suppl_pair_hinge}
\end{equation}
and minimize
\begin{align}
\mathcal L_{\mathrm{CR}}&=
\frac{\sum_{(i,j)\in\mathcal C^{\mathrm{raw}}}\ell_{ij}}
{\max(|\mathcal C^{\mathrm{raw}}|,1)},
\label{eq:suppl_l_cr}\\
\mathcal L_{\mathrm{CP}}&=
\frac{\sum_{(i,j)\in\mathcal S^{\mathrm{raw}}}w_{ij}\ell_{ij}}
{\max\!\left(\sum_{(i,j)\in\mathcal S^{\mathrm{raw}}}w_{ij},1\right)}.
\label{eq:suppl_l_cp}
\end{align}
The weight $w_{ij}$ is $1$ when at least one object belongs
to $M$, and $0.25$ otherwise. Thus, collision repair
addresses existing overlaps, while collision preservation
reduces the risk of disrupting initially safe pairs.
The clearance margin is $2.5$\,mm.

Boundary losses follow the same repair-and-preservation
principle. For the sampled footprint points $Q_i(X)$,
we obtain a conservative boundary clearance from their
signed distances $s_T(q)$ to the Tabletop Support Polygon:
\begin{equation}
\begin{aligned}
b_i(X)&=-\tau_b\log\!\sum_{q\in Q_i(X)}
   \exp\!\left(-\frac{s_T(q)}{\tau_b}\right),\\
\tau_b&=0.001/L.
\end{aligned}
\label{eq:suppl_softmin_boundary}
\end{equation}
Here positive clearance indicates points inside the tabletop.
Let $M_B$ denote annotated boundary-repair targets, and
$S_B=\{i:b_i(X^{\mathrm{raw}})\ge0,\ i\notin M_B\}$
the initially safe objects. We set
$t_i=\min(0.01/L,b_i(X^\star))$ and
$r_i=\min(0.01/L,b_i(X^{\mathrm{raw}}))$.
The boundary objective separates correction of existing
violations ($\mathcal L_{\mathrm{BR}}$), preservation of
available clearance ($\mathcal L_{\mathrm{BD}}$), and
prevention of new violations ($\mathcal L_{\mathrm{BN}}$):
\begin{align}
\mathcal L_{\mathrm{BR}}&=
\frac{\sum_{i\in M_B}\pospart{t_i-b_i(\hat X)}^2}
{\max(|M_B|,1)},
\label{eq:suppl_l_br}\\
\mathcal L_{\mathrm{BD}}&=
\frac{\sum_{i\in S_B}\pospart{r_i-b_i(\hat X)}^2}
{\max(|S_B|,1)},
\label{eq:suppl_l_bd}\\
\mathcal L_{\mathrm{BN}}&=
\frac{\sum_{i\in S_B}\pospart{-b_i(\hat X)}^2}
{\max(|S_B|,1)}.
\label{eq:suppl_l_bn}
\end{align}
These terms use a maximum reference clearance of $1$\,cm.
Padding is excluded from all averages, and an empty
applicable set contributes zero. Combining the eight terms
gives the training objective used in the main paper:
\begin{equation}
\begin{aligned}
\mathcal L_R={}&\mathcal L_{\mathrm{target}}
+0.075\mathcal L_{\mathrm{stay}}
+0.075\mathcal L_{\mathrm{yaw}}\\
&+0.08\mathcal L_{\mathrm{CR}}
+0.20\mathcal L_{\mathrm{CP}}\\
&+0.08\mathcal L_{\mathrm{BR}}
+0.08\mathcal L_{\mathrm{BD}}
+0.20\mathcal L_{\mathrm{BN}}.
\end{aligned}
\label{eq:suppl_l_total}
\end{equation}
The Refiner is trained for a single correction step;
neither safety filtering nor the tidiness score $\Psi$
is included in this objective.

\subsection{Safety-Constrained Inference}
\label{sec:suppl_safe_inference}

At inference, we apply the same Refiner weights for three
successive correction steps. Given a current layout $X$,
the network proposes $Y=\mathcal N(X+f_\omega(X,C))$.
We test the scaled updates
$X_\alpha=\mathcal N(X+\alpha(Y-X))$ for
$\alpha\in\{1,0.75,0.5,0.25,0\}$.
A step is admissible only if it creates no new colliding
pairs or boundary-violating objects:
\begin{equation}
\mathcal C(X_\alpha)\subseteq\mathcal C(X),\qquad
\mathcal B(X_\alpha)\subseteq\mathcal B(X).
\label{eq:suppl_safe_acceptance}
\end{equation}
Among admissible steps, we favor fewer violations, then
lower total penetration, and finally a larger step.
The geometric checker is rerun after each iteration.

\subsection{Six-Dimensional Tidiness Evaluation}
\label{sec:suppl_score}

Geometric validity alone does not indicate whether objects
are organized in a natural and functional manner. We therefore
score only geometrically valid layouts along six dimensions:
\emph{Compactness}, \emph{Orientation}, \emph{Alignment},
\emph{Centering}, \emph{Clearance}, and
\emph{Group Organization}, denoted by $(C,O,R,B,H,G)$.
Let $q_i$ denote each object's planar position in meters,
$(w_i,h_i)$ its footprint dimensions, and $g_i$ its
functional group. Below,
$h(z)=3z^2-2z^3$ for $z\in[0,1]$.

\paragraph{Compactness and Orientation.}
Compactness rewards nearby objects within the same functional
group, with a distance scale adjusted for footprint size.
For each group $g$ with $|g|\ge2$, let $\bar d_g$ be its
mean pairwise center distance and
$a_g=|g|^{-1}\sum_{i\in g}\sqrt{w_i h_i}$.
The compactness score is
\begin{equation}
C=\frac1{|\mathcal G_2|}\sum_{g\in\mathcal G_2}
\exp\!\left[-\frac{\bar d_g}{\max(0.08,2a_g)}\right],
\label{eq:suppl_compact}
\end{equation}
where $\mathcal G_2$ contains groups with at least two objects.
Orientation instead compares each applicable object's yaw
with that of its corresponding Reference Layout:
\begin{equation}
O=\frac1{|\mathcal I_\theta|}
\sum_{i\in\mathcal I_\theta}
\frac{1+\cos\!\left(k_i(\theta_i-\theta_i^\star)\right)}{2}.
\label{eq:suppl_orientation}
\end{equation}
The symmetry factors $k_i$ follow the same convention as
the yaw-preservation loss; continuously symmetric objects
are omitted. Unlike a reference-free orientation preference,
this component uses the clean Reference Layout's yaw.

\paragraph{Alignment and Centering.}
Alignment measures how closely group members form a row or
column. Let $r_x(g)$ and $r_y(g)$ be the coordinate spans
of the object centers within group $g$. The Reference Layout's
arrangement mode determines
$v_g=r_y(g)$ for a row, $v_g=r_x(g)$ for a column,
and $v_g=\min(r_x(g),r_y(g))$ otherwise. Then
\begin{equation}
R=\frac1{|\mathcal G_2|}\sum_{g\in\mathcal G_2}
\exp(-v_g/0.12).
\label{eq:suppl_alignment}
\end{equation}
Centering discourages a collective shift toward the edge of
the tabletop. Using footprint areas $A_i$, tabletop polygon
centroid $c_T$, and tabletop bounding-box diagonal $D$,
\begin{equation}
\bar q=\frac{\sum_iA_iq_i}{\sum_iA_i},\qquad
B=\max\!\left(0,1-\frac{\|\bar q-c_T\|_2}{D/2}\right).
\label{eq:suppl_centering}
\end{equation}

\paragraph{Clearance and Group Organization.}
Clearance measures available spacing around individual
objects. Let $d_{ij}\ge0$ be the minimum unsigned distance
between two footprints (zero for overlap), and
$d_i=\min_{j\ne i}d_{ij}$. The score is
\begin{equation}
H=\frac1N\sum_{i=1}^N
h\!\left(\operatorname{clip}\!\left(
\frac{d_i-0.01}{0.06-0.01},0,1\right)\right).
\label{eq:suppl_clearance}
\end{equation}
Group Organization complements Compactness by assessing
whether same-group neighbors are closer than objects in
other groups. For applicable objects $i\in\mathcal I$,
let $d_i^{\mathrm{same}}$ and $d_i^{\mathrm{cross}}$ denote
the nearest same-group and cross-group footprint distances:
\begin{equation}
G=\frac1{|\mathcal I|}\sum_{i\in\mathcal I}
h\!\left(\operatorname{clip}\!\left(
\frac{d_i^{\mathrm{cross}}-d_i^{\mathrm{same}}}{0.08},0,1
\right)\right).
\label{eq:suppl_grouping}
\end{equation}

\paragraph{Overall score.}
The fixed weights, in the order above, are
$w=(0.30,0.15,0.20,0.20,0.15,0.10)$.
Components with no applicable objects or groups are omitted,
and the remaining scores are combined as in the main paper:
\begin{equation}
\Psi(X;C)=
\frac{\sum_{d\in\mathcal A(C)}w_d s_d(X;C)}
{\sum_{d\in\mathcal A(C)}w_d}.
\label{eq:suppl_score_applicable}
\end{equation}
This predefined score ranks valid Reference Layout candidates
during dataset construction and evaluates generated layouts.
The scoring weights were fixed before the human preference
survey and are not fitted to its responses.

\section{Datasets and Evaluation Protocol}
\label{sec:suppl_protocol}

\subsection{Dataset Descriptions}
\label{sec:suppl_datasets}
We evaluate on General Tabletop Layouts (GTL) and Workstation
Tabletop Layouts (WTL). GTL contains everyday objects, including
books, stationery, auxiliary tools, decorations, and fruits.
WTL additionally includes workstation objects such as monitors,
keyboards, and mice, with corresponding functional placement constraints.
\cref{tab:suppl_splits} summarizes the condition-level splits.
\begin{table}[t]
\centering
\caption{GTL and WTL condition-level dataset splits.}
\label{tab:suppl_splits}
\footnotesize\setlength{\tabcolsep}{4pt}
\begin{tabular}{lrrrr}
\toprule
Dataset & Train & Val & Test & Total\\
\midrule
GTL & 2,015 & 244 & 241 & 2,500\\
WTL & 4,817 & 585 & 759 & 6,161\\
\bottomrule
\end{tabular}
\end{table}
Conditions are separated by complete
object-set signatures: no identical object set appears in different
splits, although individual assets and tabletop shapes may recur.
Evaluation therefore targets novel object combinations rather than
entirely unseen object assets. Each condition has an associated
geometrically valid reference layout.

\subsection{Implementation Details}
\label{sec:suppl_implementation}
The Flow generator uses a six-layer Transformer with hidden width 256,
eight attention heads, and feed-forward width 1,024. Each object is
represented by a 147-dimensional static descriptor, combining
semantic (64), functional-group (32), and symmetry (16) embeddings,
three geometric dimensions, and a 32-dimensional footprint encoding.
The support polygon is represented by 128 sampled boundary points,
each containing two normalized positions, two outward-normal
components, and a signed vertex-turning feature $\kappa$. The boundary encoder produces eight 256-dimensional tabletop tokens.

Flow and Refiner are trained separately with AdamW.
For Flow training, the batch size is 64, the learning rate is \(2\times10^{-4}\), the weight decay is \(10^{-5}\), and the gradient-clipping threshold is 1.0.
The reported Flow checkpoints correspond to approximately 20,000 parameter updates on GTL and 22,500 on WTL. The Refiner uses four
edge-aware attention layers and is trained for up to 96 epochs
with the eight-term objective in \cref{eq:suppl_l_total}.

At inference, each condition is sampled with $K=32$ independent
Gaussian initial states. A 32-interval ODE integrator uses Heun
predictor--corrector updates for the first 31 intervals and a final
Euler step, requiring 63 velocity evaluations. The resulting
Raw Layouts undergo three shared-weight, safety-constrained Refiner
iterations before the final hard validity check and ranking.

\subsection{Evaluation Metrics}
\label{sec:suppl_metrics}
Let $M$ be the number of test conditions, $K$ the number of
candidates per condition, and $X_m^{(k)}$ candidate $k$ under
condition $m$. Let $\mathcal C(X)$ be the set of colliding
object pairs and $\mathcal B(X)$ the set of boundary-violating
objects. Define
\begin{equation}
\operatorname{Valid}(X)=\mathbb I\left[
\mathcal C(X)=\varnothing\land\mathcal B(X)=\varnothing\right].
\label{eq:suppl_valid}
\end{equation}
The geometric validity, collision, and boundary violation rates are
\begin{align}
\mathrm{GVR}&=\frac{1}{MK}\sum_{m=1}^{M}\sum_{k=1}^{K}
\operatorname{Valid}(X_m^{(k)}),
\label{eq:suppl_gvr}\\
\mathrm{CR}&=\frac{1}{MK}\sum_{m,k}
\mathbb I[\mathcal C(X_m^{(k)})\ne\varnothing],
\label{eq:suppl_cr}\\
\mathrm{BVR}&=\frac{1}{MK}\sum_{m,k}
\mathbb I[\mathcal B(X_m^{(k)})\ne\varnothing].
\label{eq:suppl_bvr}
\end{align}
CR and BVR may overlap within a candidate and therefore do not
necessarily add up to $1-\mathrm{GVR}$.

Condition-level feasible coverage is defined as
\begin{equation}
\mathrm{FCR@}K=\frac{1}{M}\sum_{m=1}^{M}
\mathbb I\!\left[\exists k\leq K:\,
\operatorname{Valid}(X_m^{(k)})=1\right].
\label{eq:suppl_fcr}
\end{equation}
Unless stated otherwise, $K=32$. The penetration depth for a
colliding pair is $p_{ij}(X)=\pospart{-g_{ij}(X)}$ in physical
centimeters after de-normalization. We report mean penetration
depth over colliding pairs,
\begin{equation}
\mathrm{PD}=
\frac{\sum_{m,k}\sum_{(i,j)\in\mathcal C(X_m^{(k)})}p_{ij}(X_m^{(k)})}
{\max\bigl(1,\sum_{m,k}|\mathcal C(X_m^{(k)})|\bigr)},
\label{eq:suppl_pd}
\end{equation}
and, separately, mean \emph{total} penetration per candidate,
\begin{equation}
D_{\Sigma}=\frac{1}{MK}\sum_{m,k}
\sum_{(i,j)\in\mathcal C(X_m^{(k)})}p_{ij}(X_m^{(k)}).
\label{eq:suppl_dsigma}
\end{equation}
PD and $D_{\Sigma}$ use different denominators and must not be
interchanged. The tidiness score $\Psi$ is computed using
\cref{sec:suppl_score} for geometrically valid candidates.

\newcommand{\FirstCandidateImage}[1]{%
    \raisebox{-0.5\height}{%
        \includegraphics[
            width=0.175\textwidth,
            height=0.175\textwidth,
            keepaspectratio
        ]{#1}%
    }%
}

% =========================================================
% Dataset label: vertically centered
% =========================================================

\newcommand{\ConditionRowLabel}[1]{%
    \parbox[c][0.175\textwidth][c]{0.062\textwidth}{%
        \centering
        \normalsize\bfseries #1%
    }%
}

% =========================================================
% Qualitative Comparison Figure
% Four rows x Five methods
% =========================================================

\begin{figure*}[t]
    \centering

    \begingroup

    % Reduce horizontal spacing
    \setlength{\tabcolsep}{1pt}

    % Compact vertical spacing
    \renewcommand{\arraystretch}{0.95}

    \begin{tabular}{@{}c*{5}{c}@{}}

    {\normalsize\textbf{Dataset}}
    &
    {\normalsize\textbf{SceneScore}}
    &
    {\normalsize\textbf{StructDiffusion}}
    &
    {\normalsize\textbf{LayoutGPT}}
    &
    {\normalsize\textbf{LEGO-Net}}
    &
    {\normalsize\textbf{TidyFlow (Ours)}}

    \\[3pt]

    \ConditionRowLabel{GTL}
    &
    \FirstCandidateImage{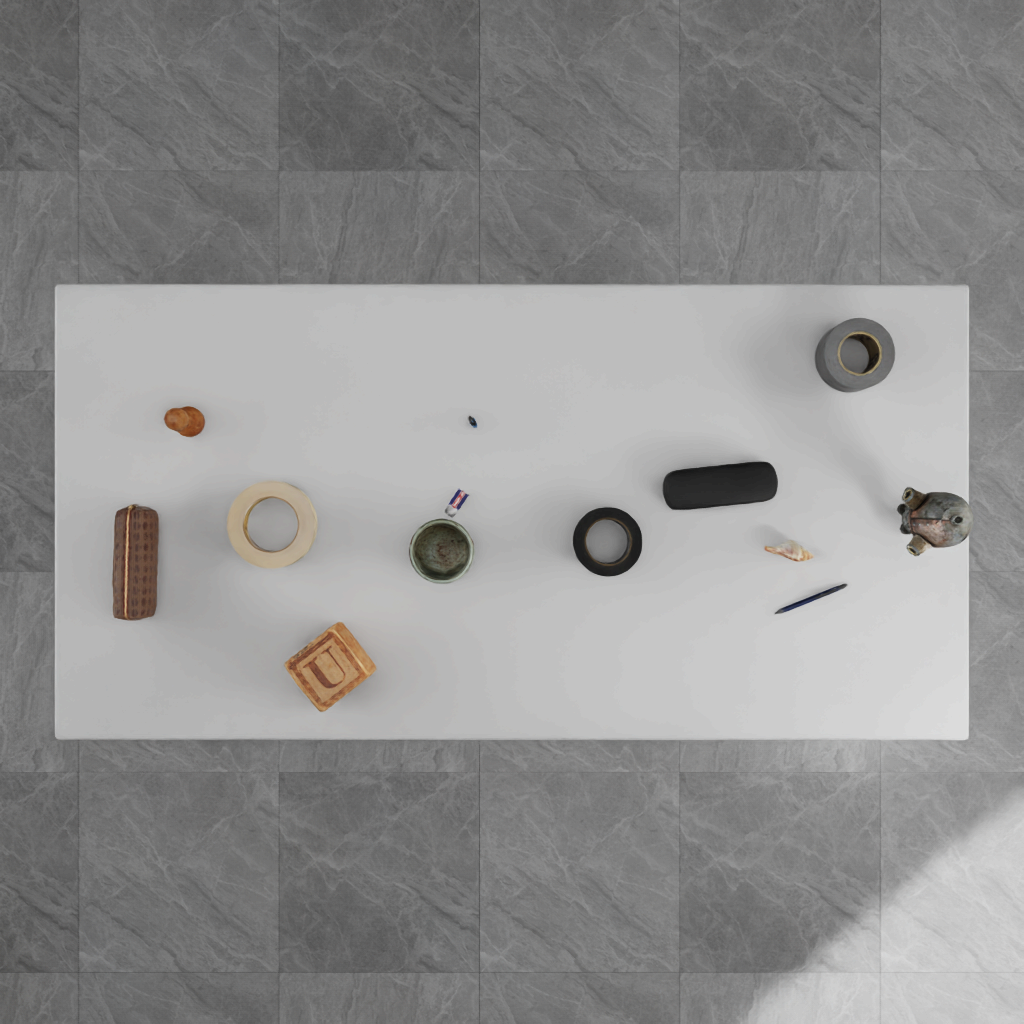}
    &
    \FirstCandidateImage{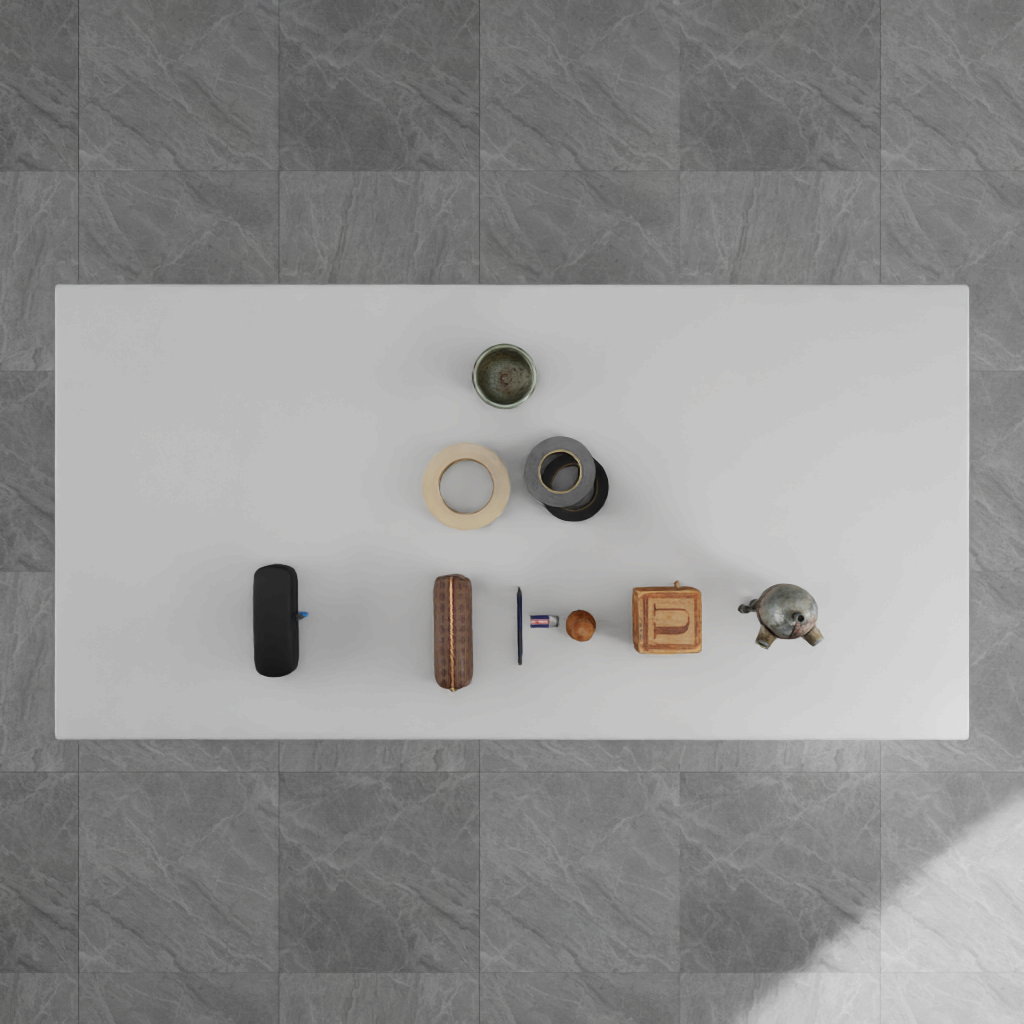}
    &
    \FirstCandidateImage{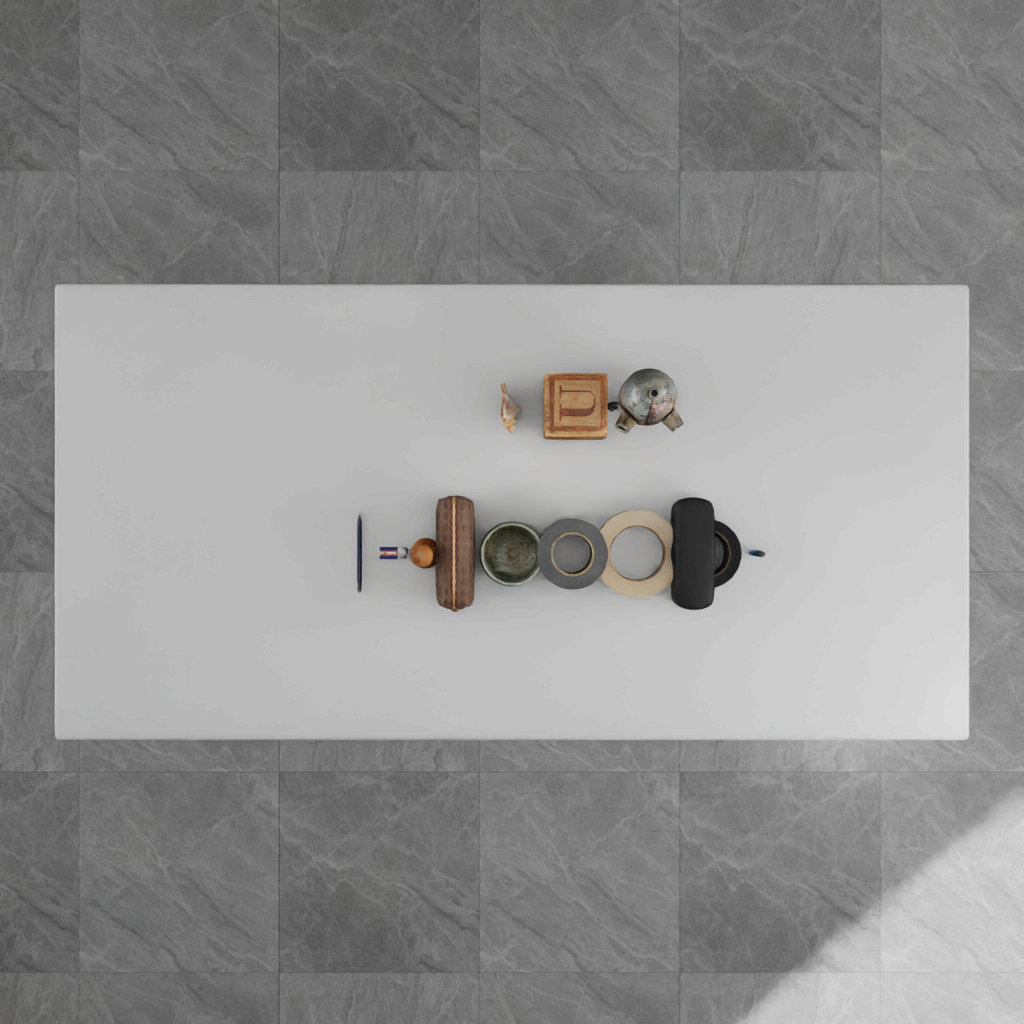}
    &
    \FirstCandidateImage{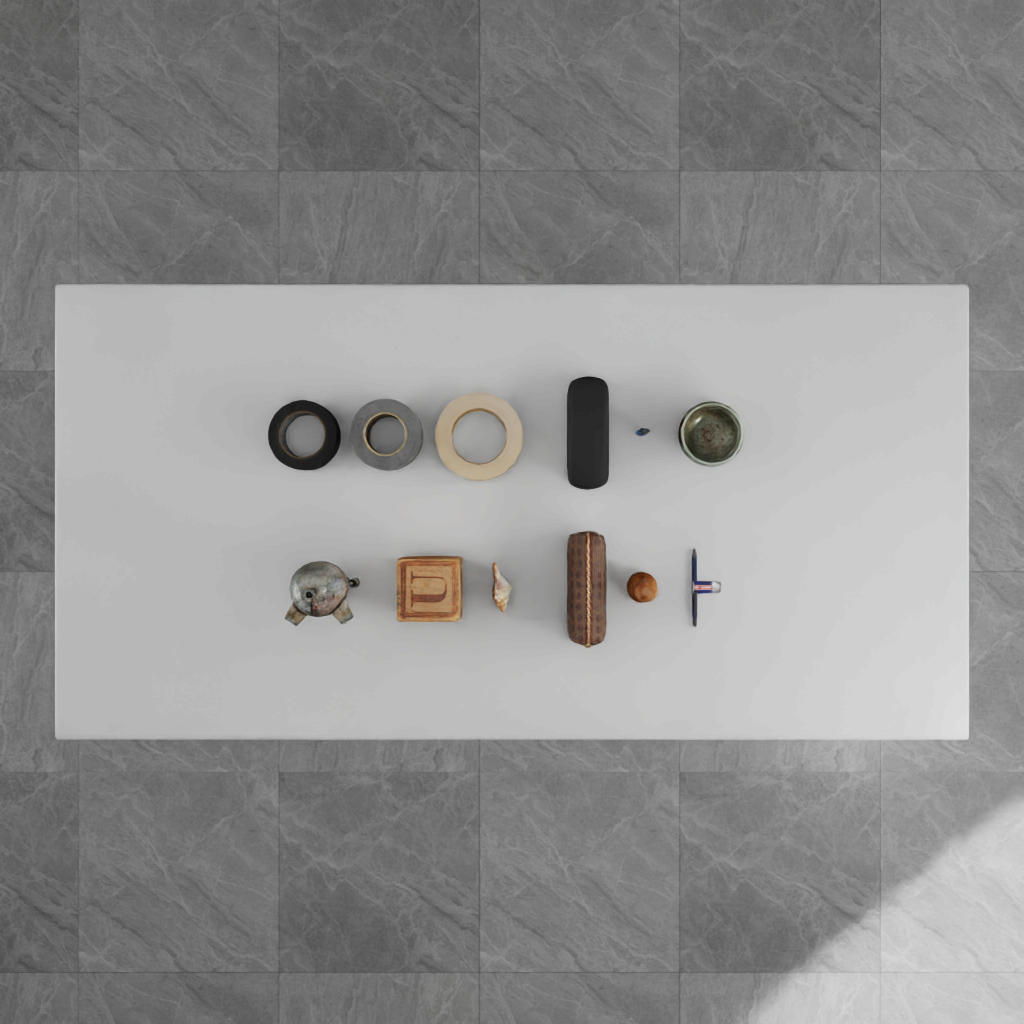}
    &
    \FirstCandidateImage{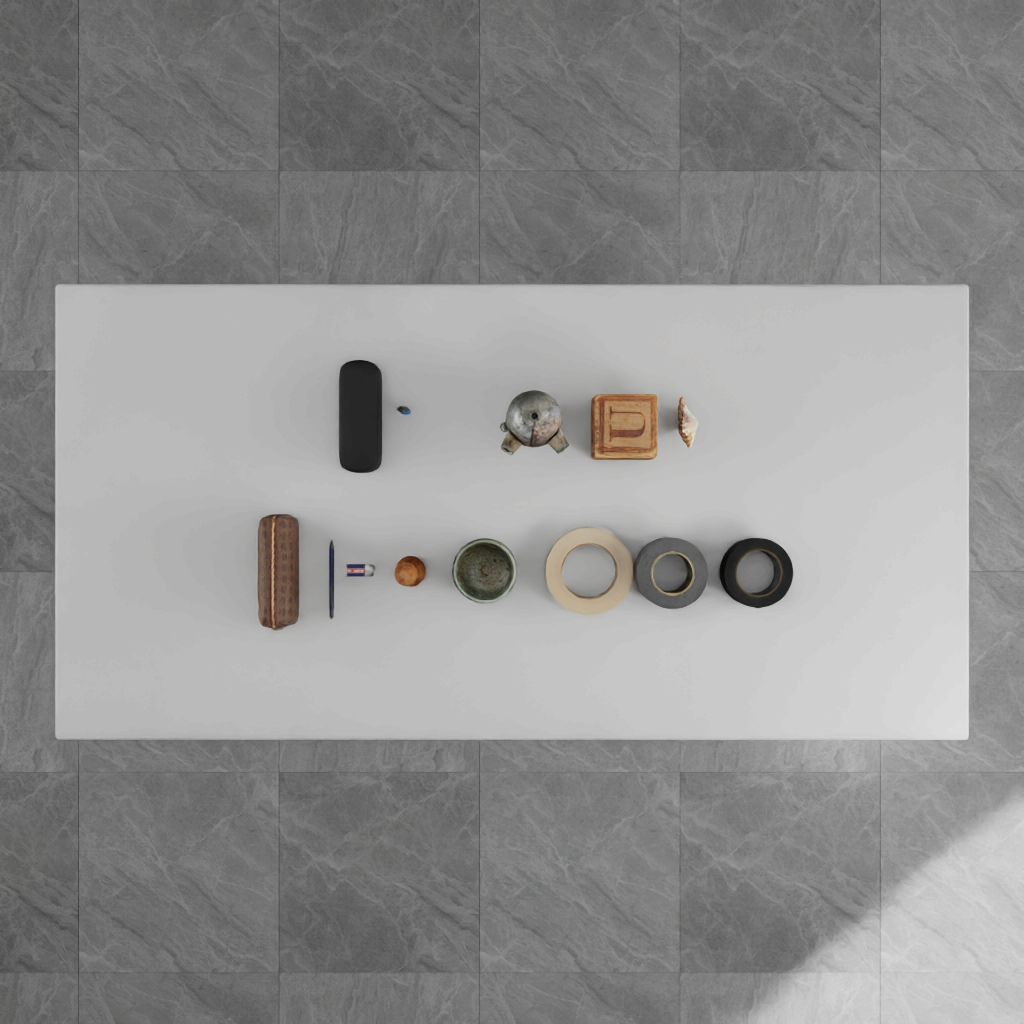}

    \\[1pt]

    \ConditionRowLabel{GTL}
    &
    \FirstCandidateImage{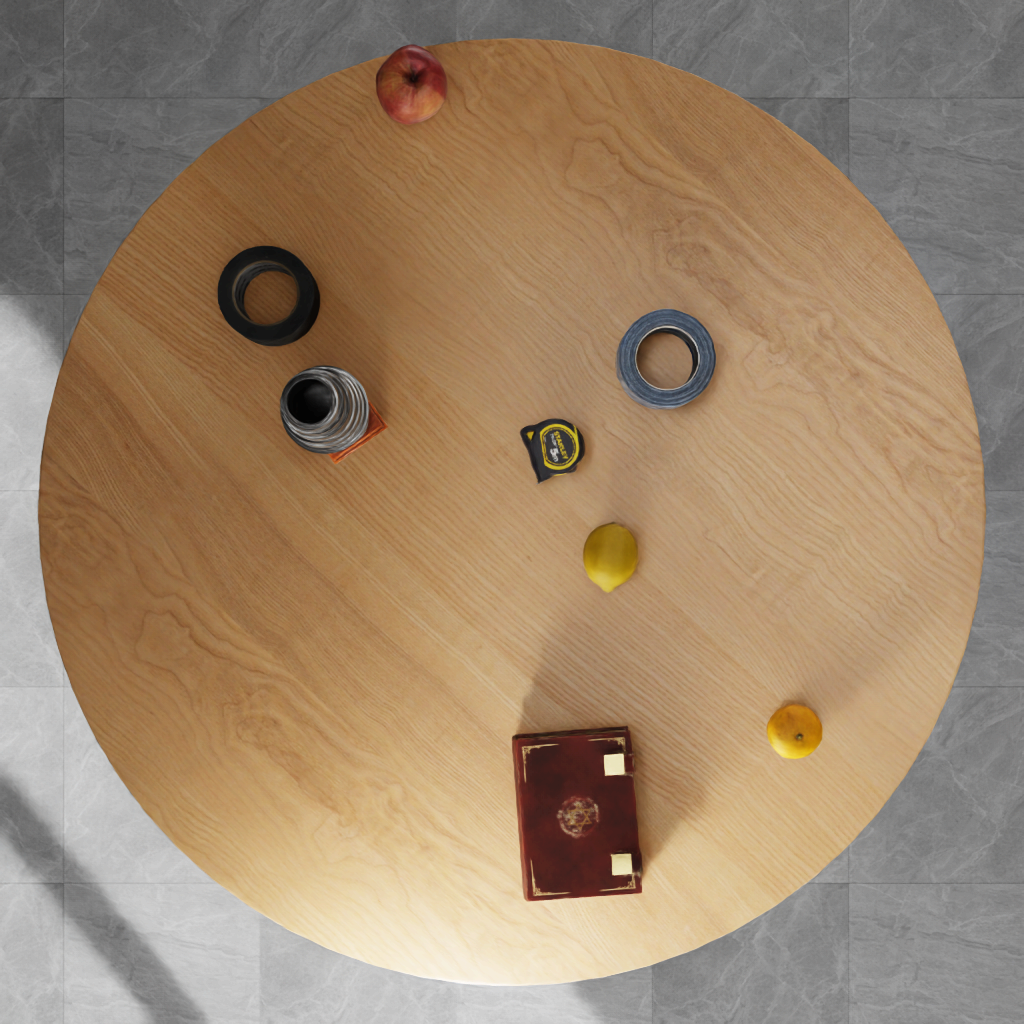}
    &
    \FirstCandidateImage{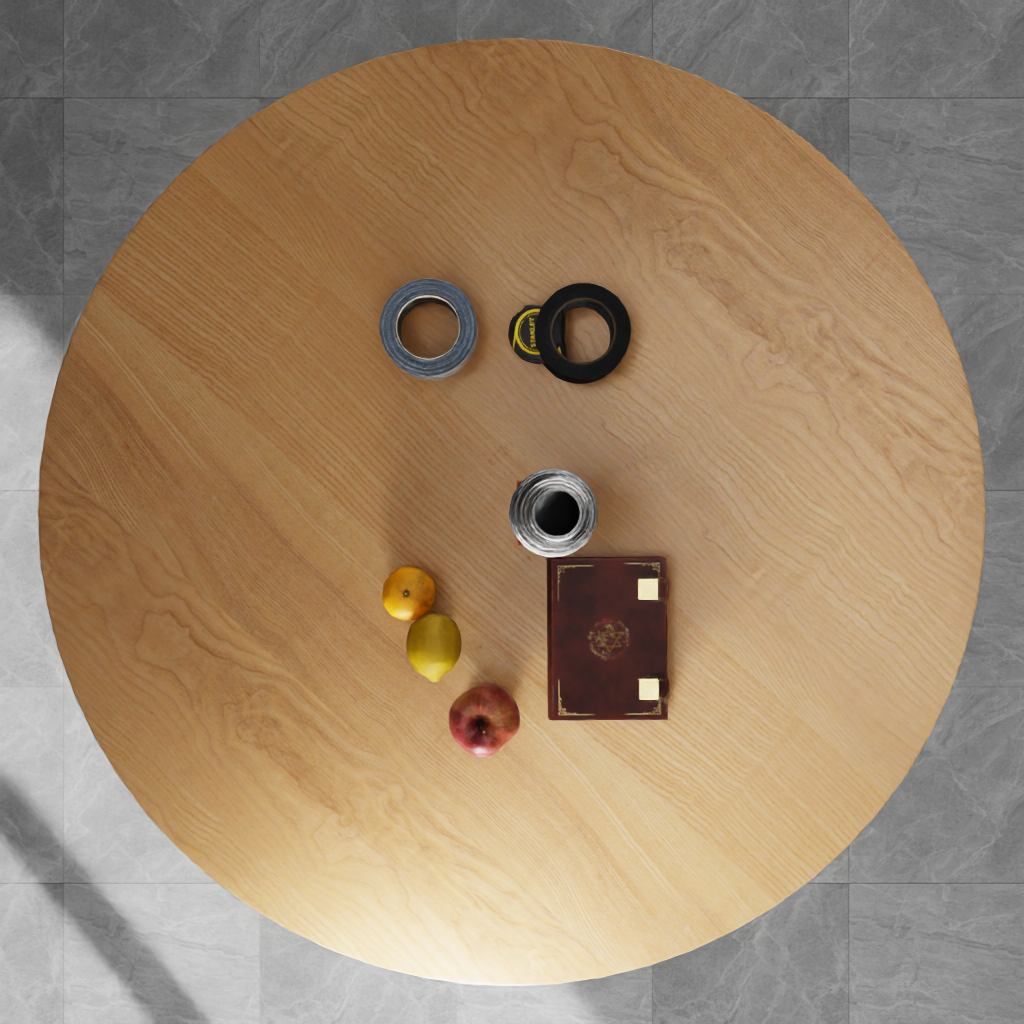}
    &
    \FirstCandidateImage{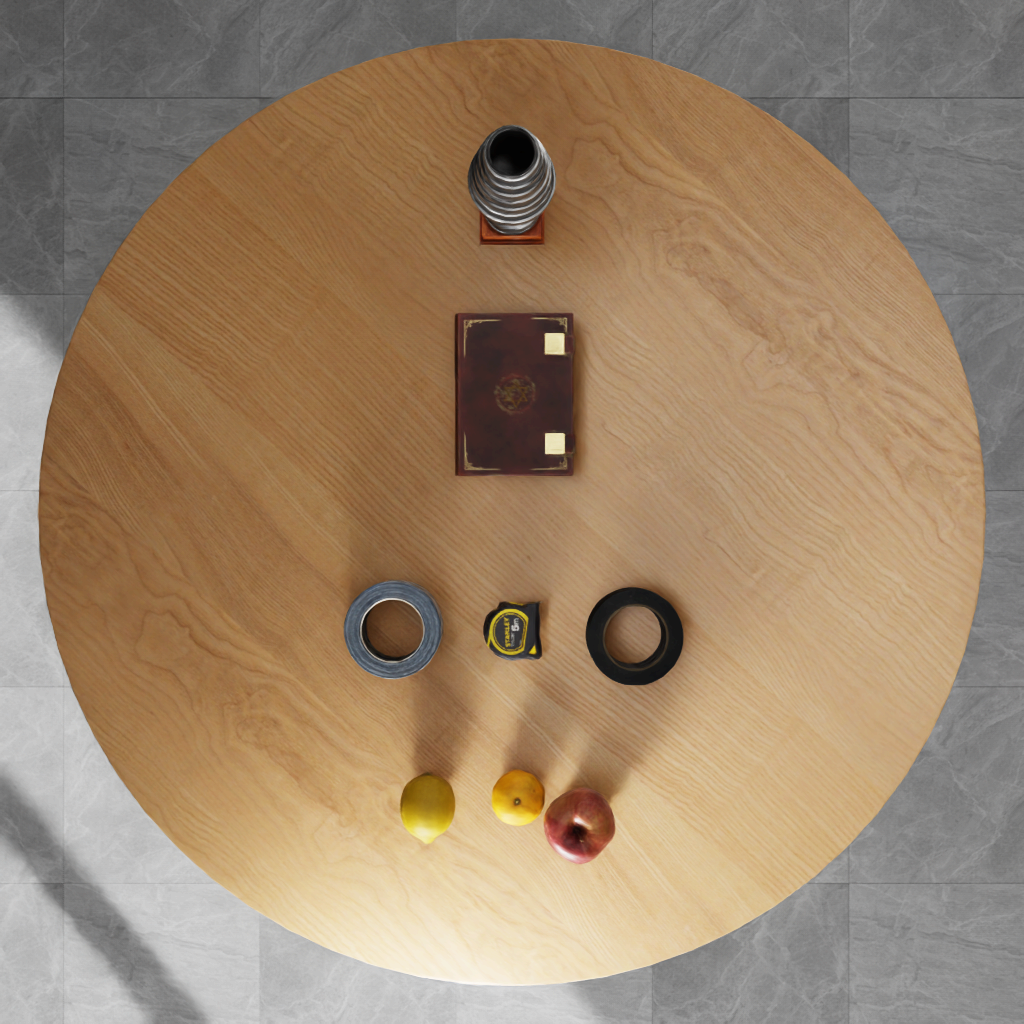}
    &
    \FirstCandidateImage{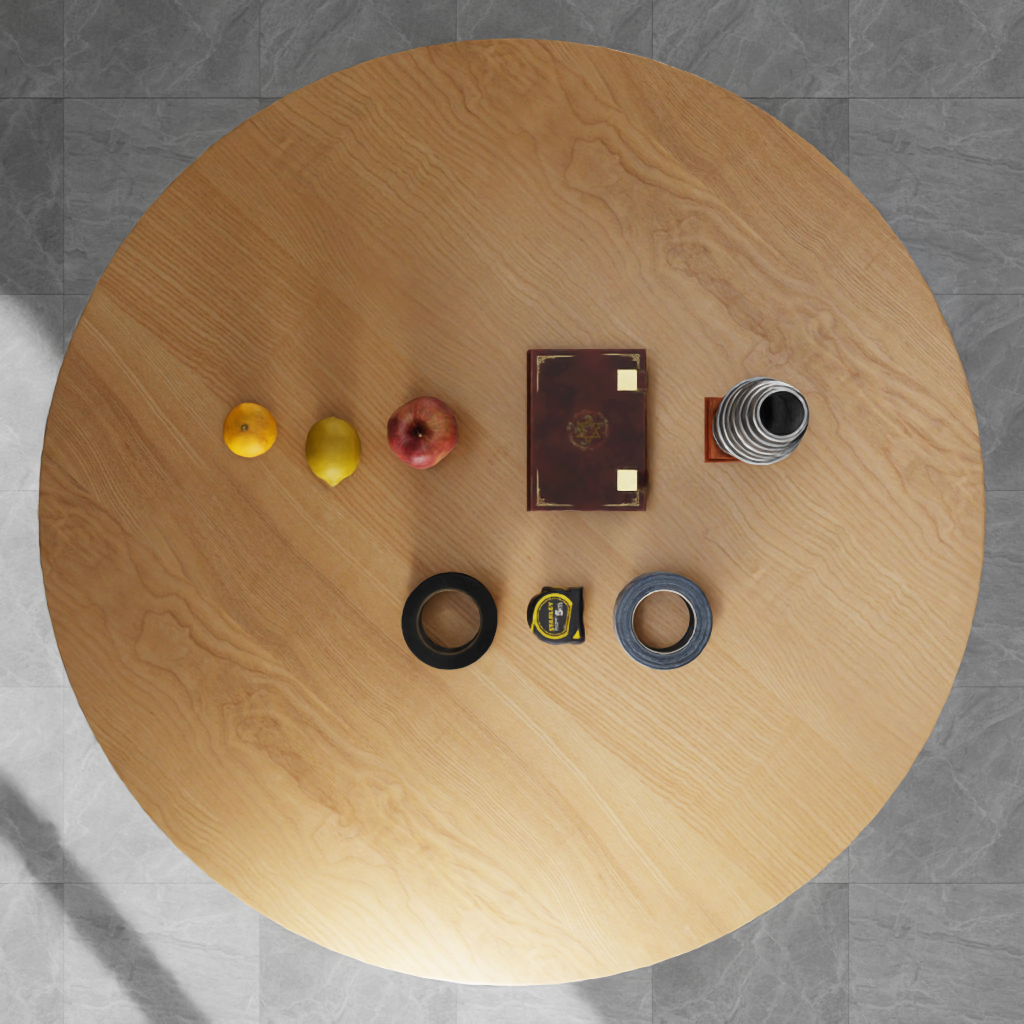}
    &
    \FirstCandidateImage{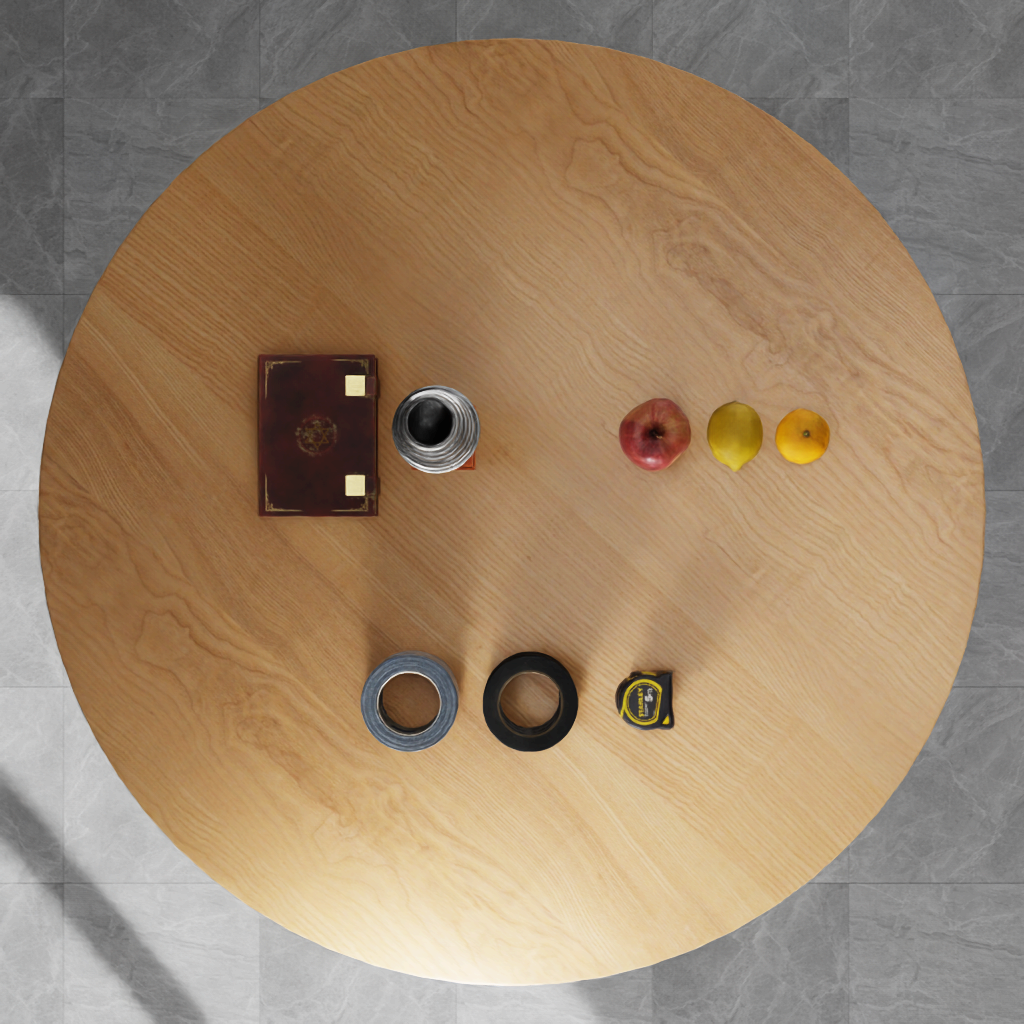}

    \\[1pt]

    \ConditionRowLabel{WTL}
    &
    \FirstCandidateImage{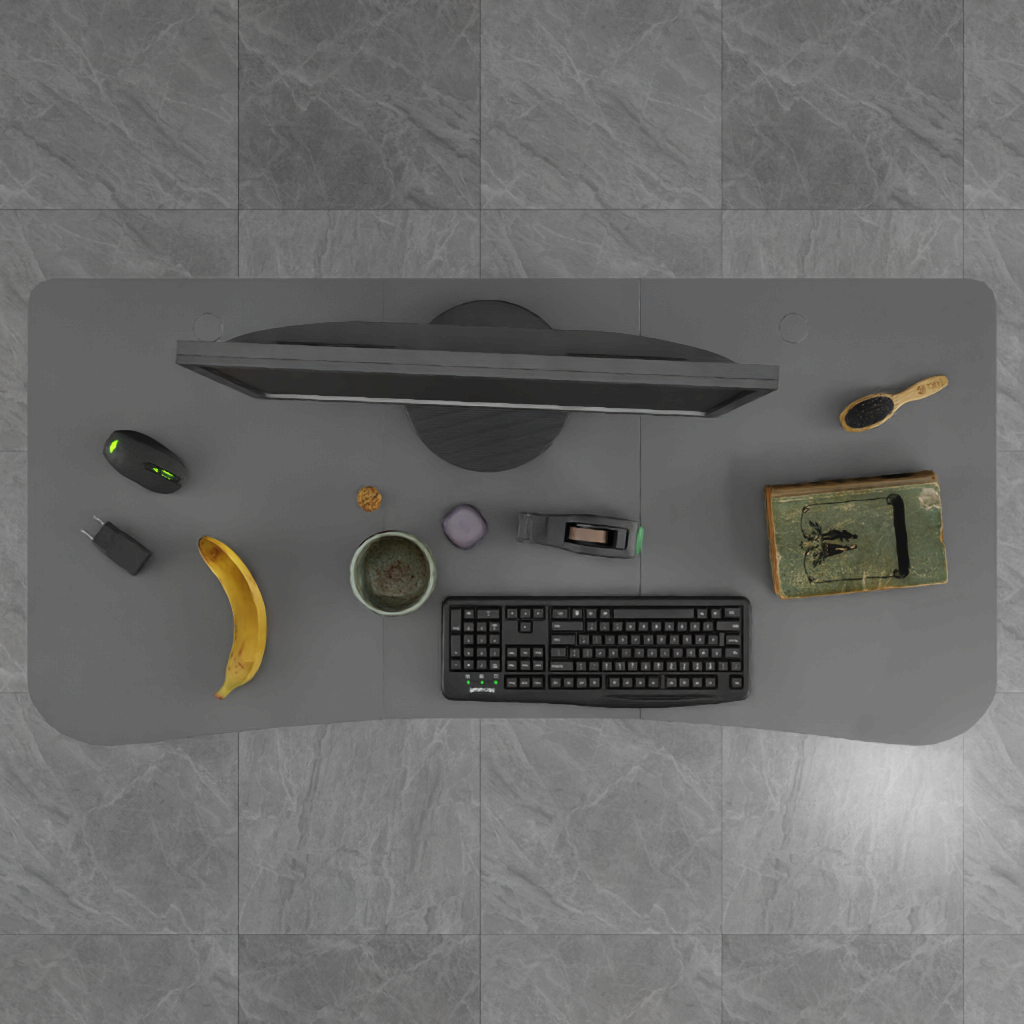}
    &
    \FirstCandidateImage{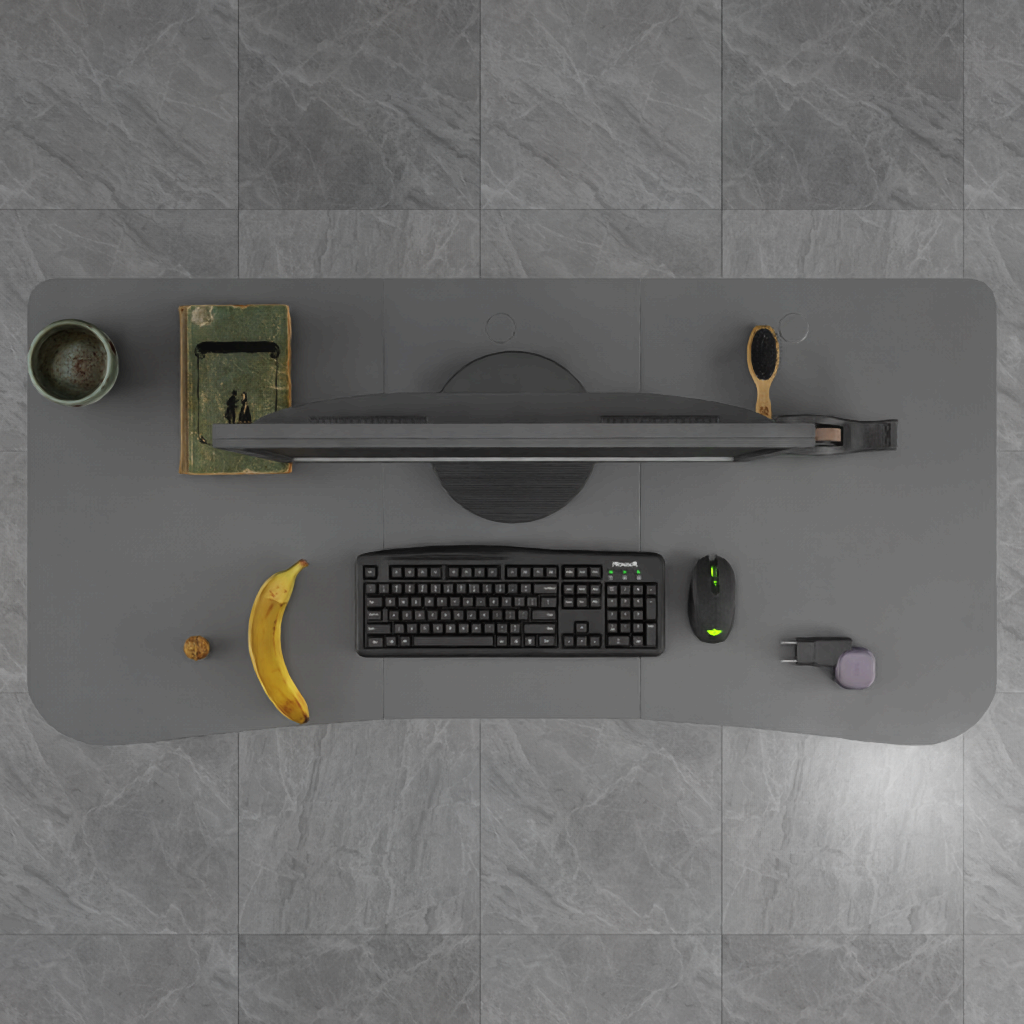}
    &
    \FirstCandidateImage{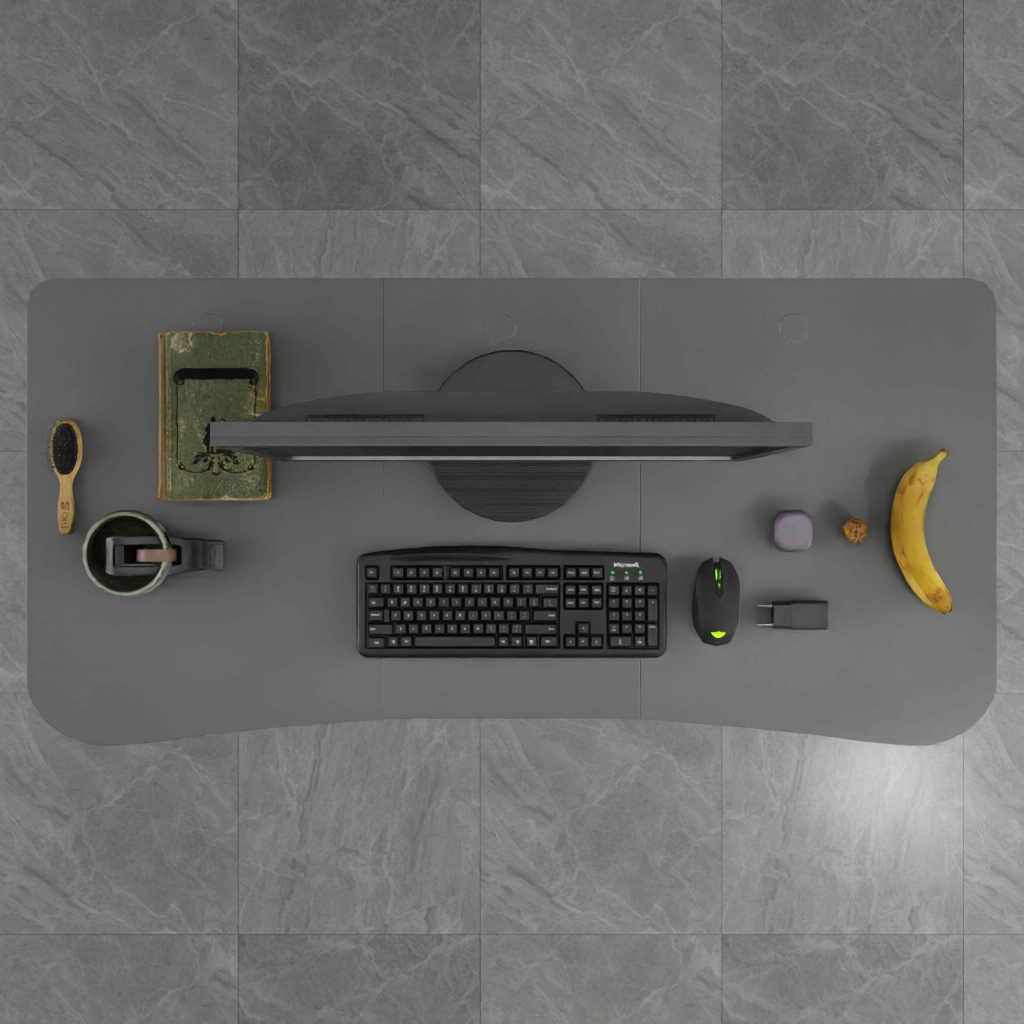}
    &
    \FirstCandidateImage{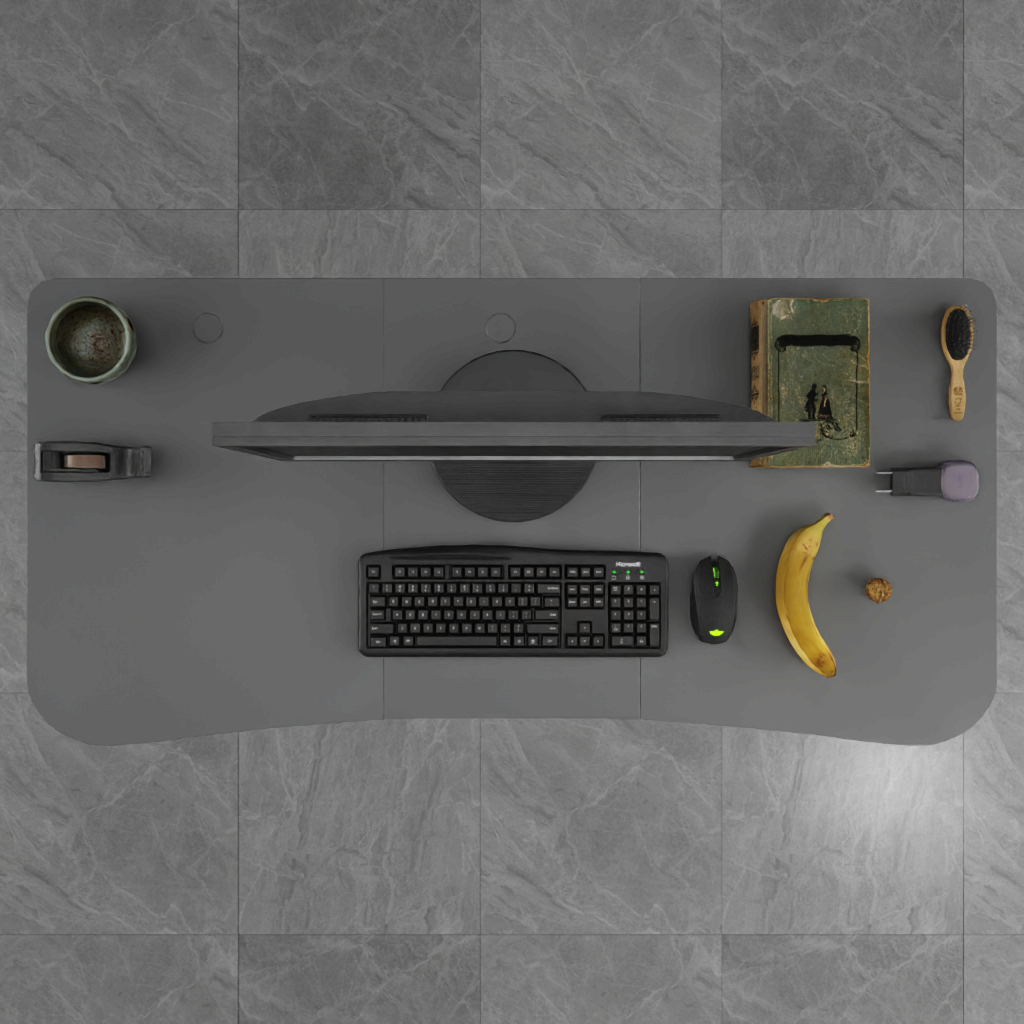}
    &
    \FirstCandidateImage{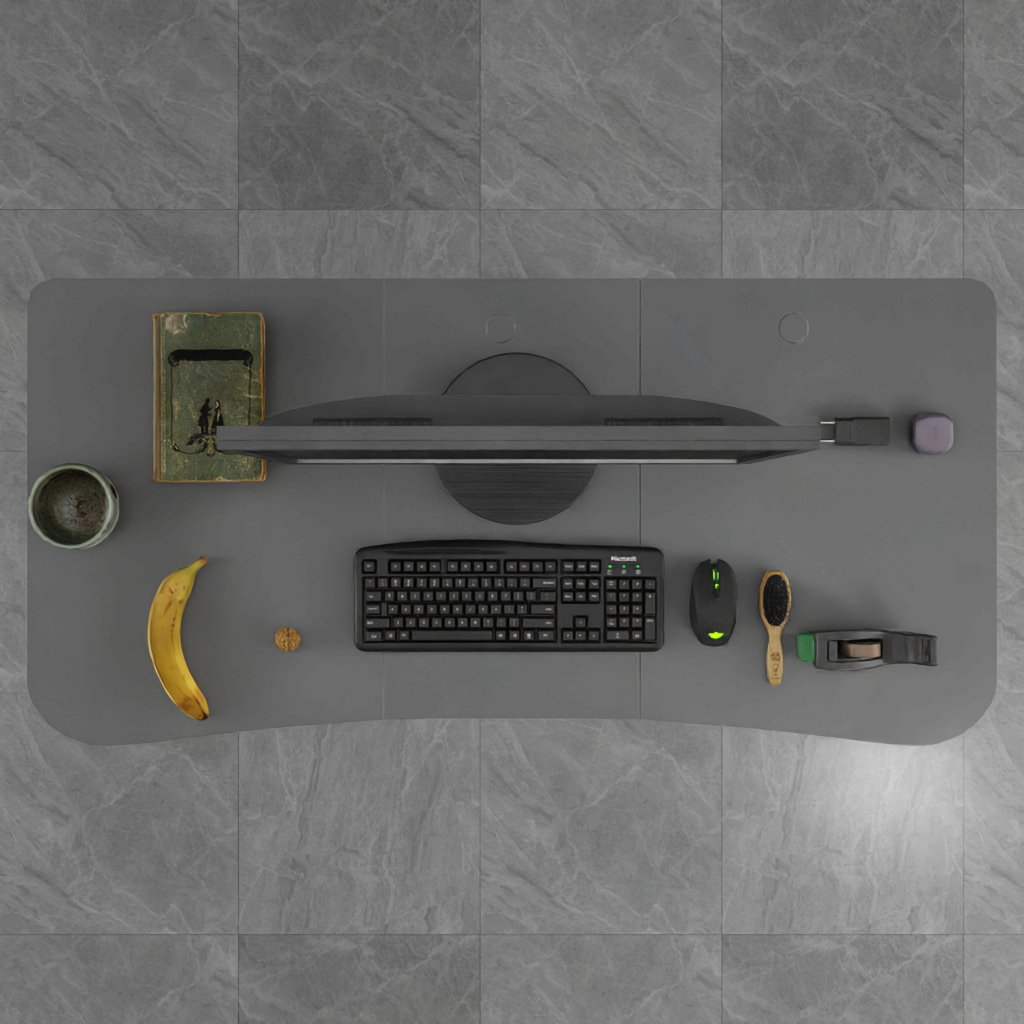}

    \\[1pt]

    \ConditionRowLabel{WTL}
    &
    \FirstCandidateImage{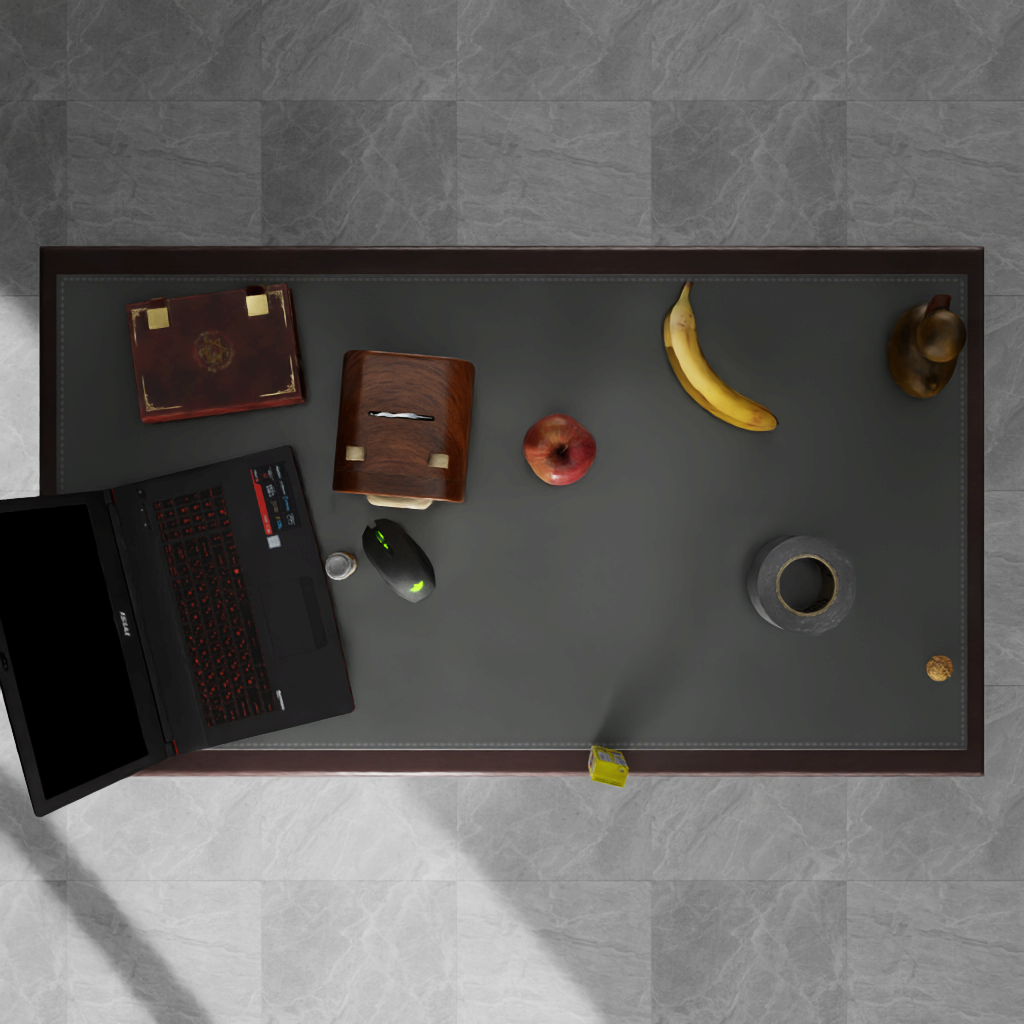}
    &
    \FirstCandidateImage{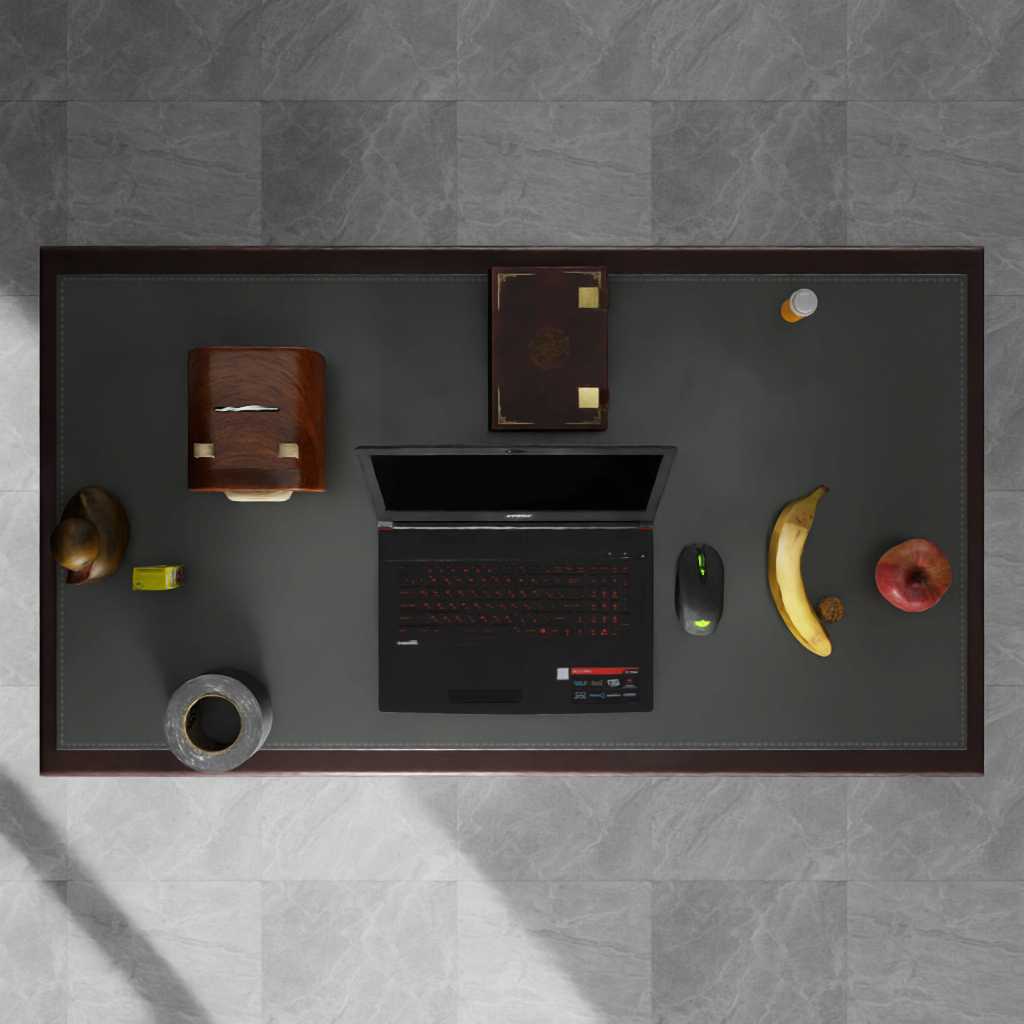}
    &
    \FirstCandidateImage{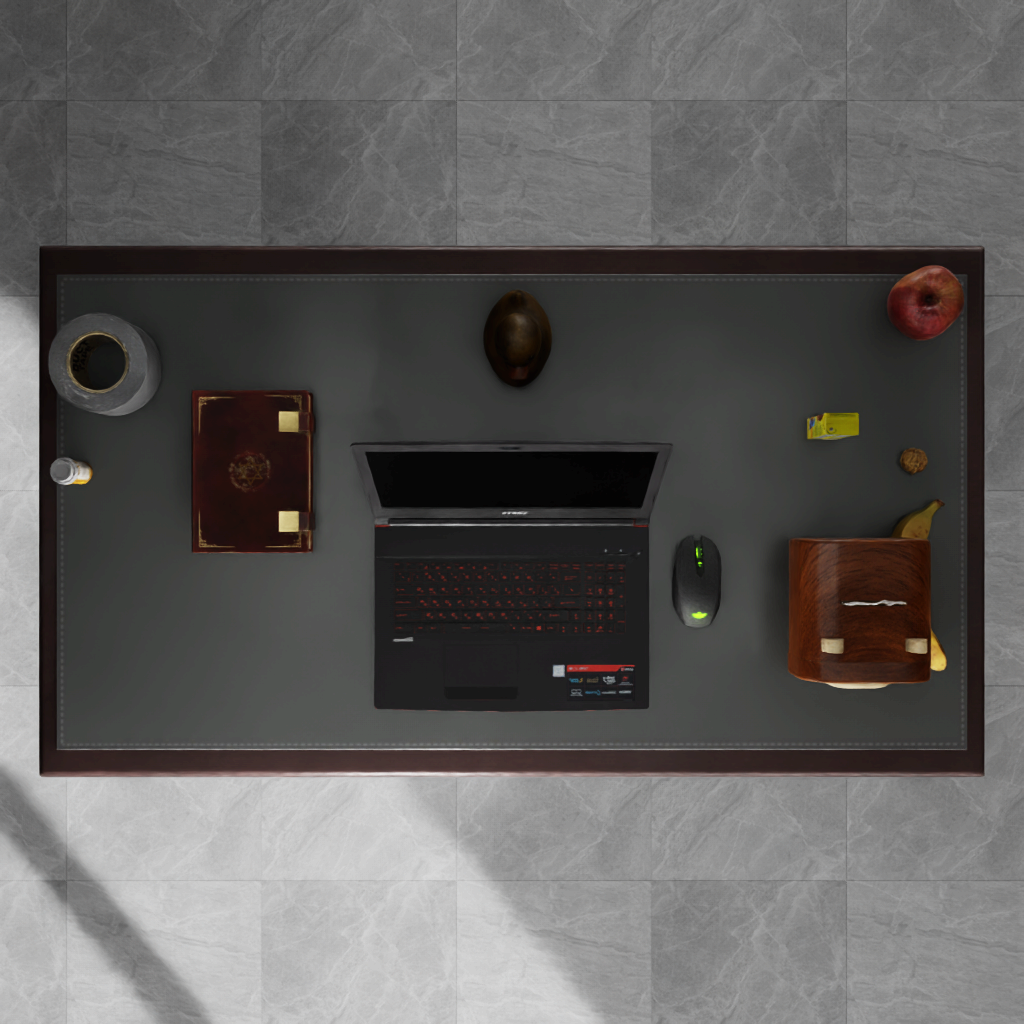}
    &
    \FirstCandidateImage{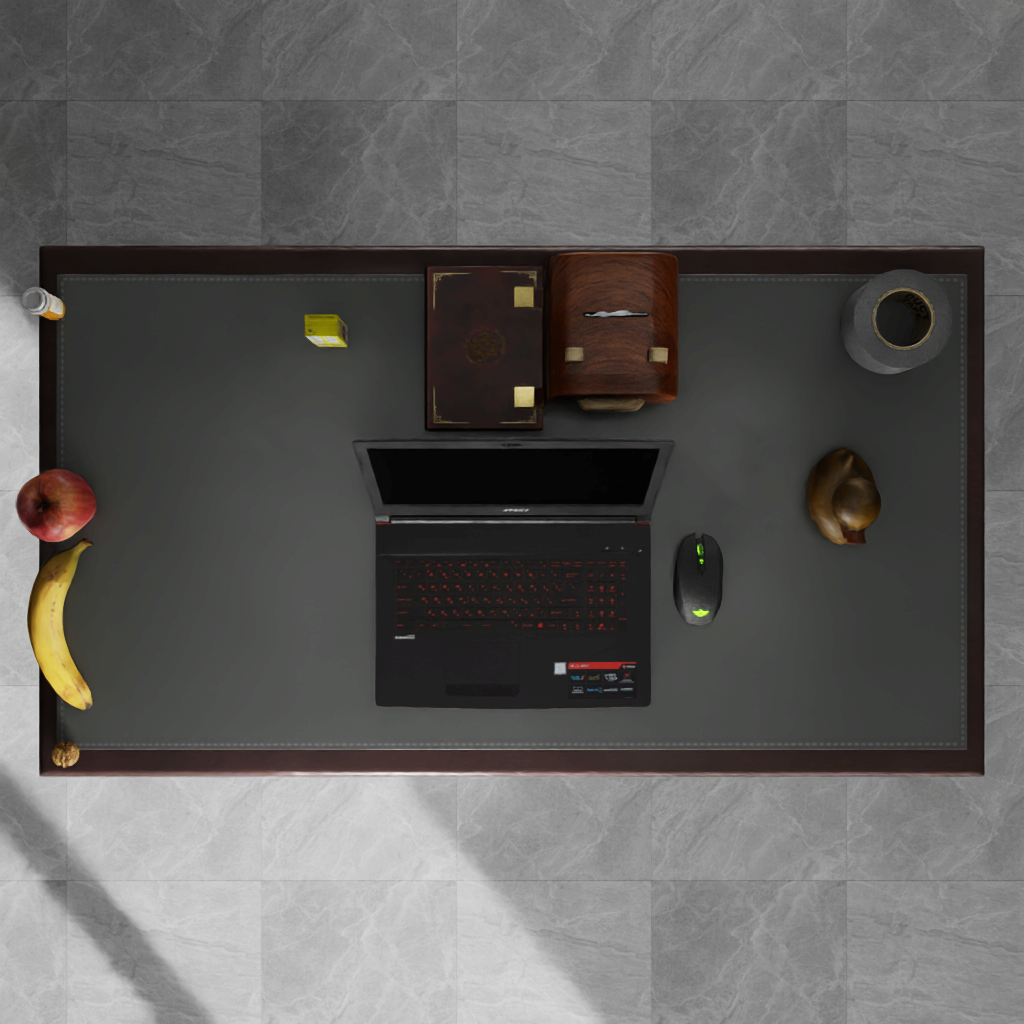}
    &
    \FirstCandidateImage{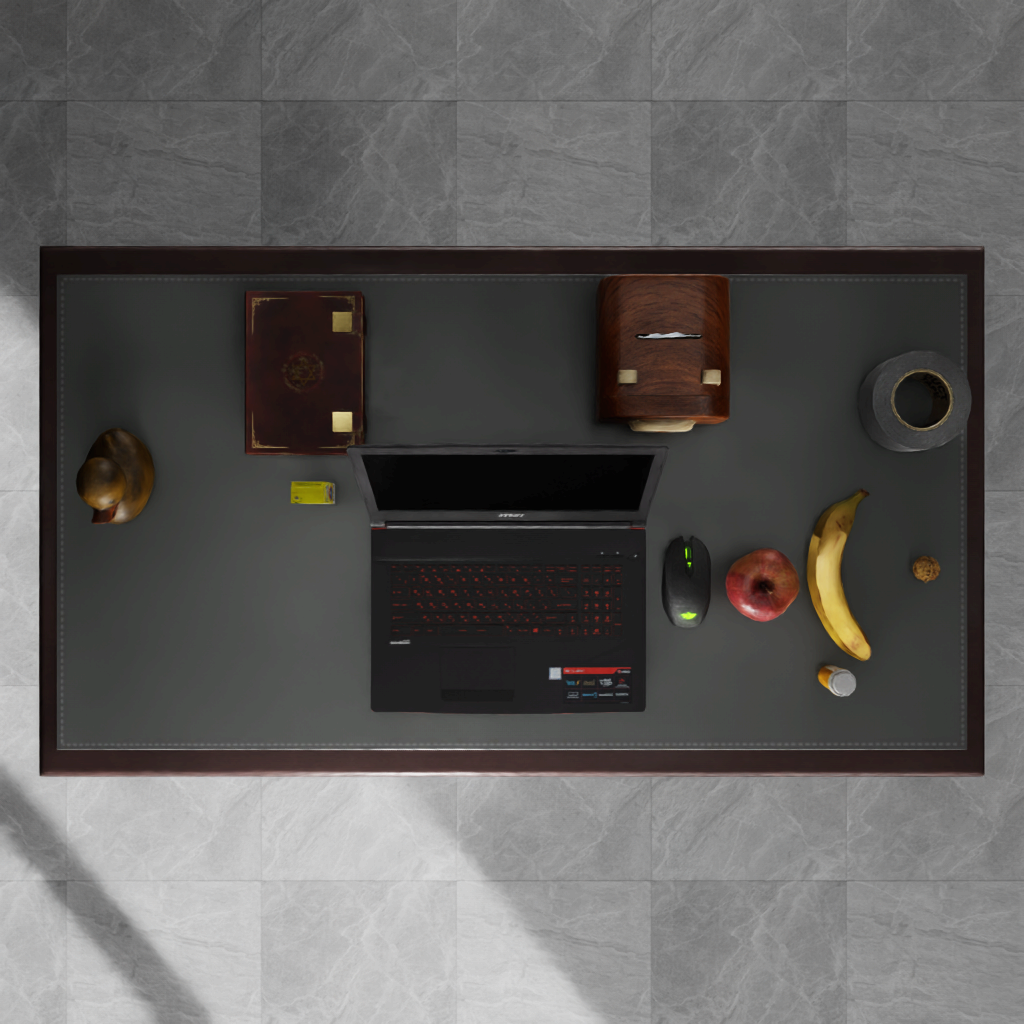}

    \\

    \end{tabular}

    \endgroup

    \caption{
    \textbf{Single-candidate qualitative comparison
    on GTL and WTL.}
    Each row shows the first generated candidate
    (index $0$) of five methods under identical
    object sets and tabletop conditions.
    The first two rows correspond to GTL,
    while the last two correspond to WTL.
    }

    \label{fig:suppl_candidate_zero}

\end{figure*}

\section{Additional Experimental Results}
\label{sec:suppl_experiments}

% =========================================================
% C.1 Single-Candidate Qualitative Comparison
% =========================================================

\subsection{Single-Candidate Qualitative Comparison}
\label{sec:suppl_candidate_zero}

To further examine single-sample generation quality,
we compare the first generated candidate (index $0$)
of five methods under identical object sets and
tabletop conditions.

As shown in Fig.~\ref{fig:suppl_candidate_zero},
some competing methods struggle to generate
geometrically valid layouts in a single attempt,
often exhibiting object collisions or boundary
violations. In contrast, TidyFlow produces
functionally organized and geometrically feasible
arrangements in the selected examples.
These results highlight the importance of
balancing functional organization and geometric
constraints without relying on multi-candidate
selection.

\subsection{Multimodal Tabletop Layout Generation}
\label{sec:suppl_multimodal}

\begin{figure*}[t]
    \centering

    {\small\textbf{GTL}}\par\vspace{3pt}

    \begin{minipage}[t]{0.19\textwidth}
        \centering
        \includegraphics[width=\linewidth]{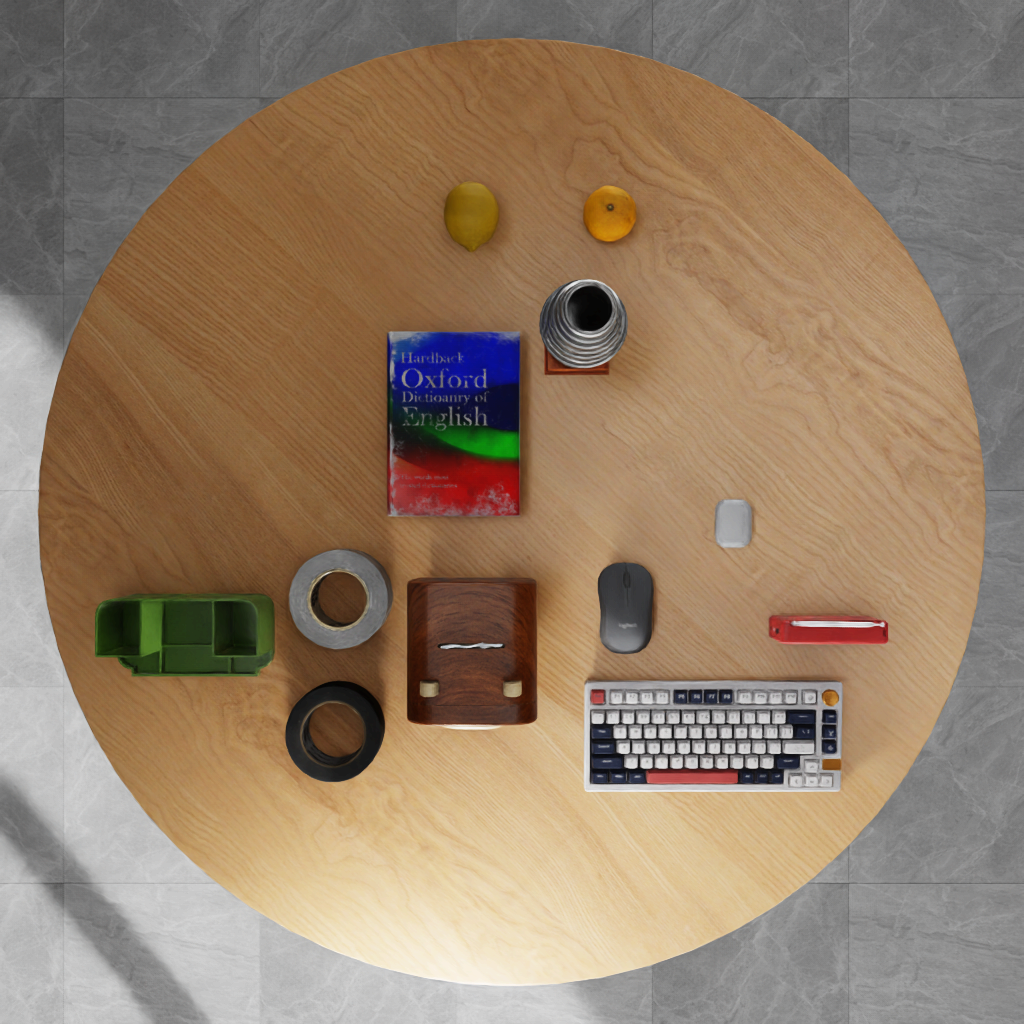}
        \par\vspace{2pt}
        {\small $\Psi=0.5966$}
    \end{minipage}
    \hfill
    \begin{minipage}[t]{0.19\textwidth}
        \centering
        \includegraphics[width=\linewidth]{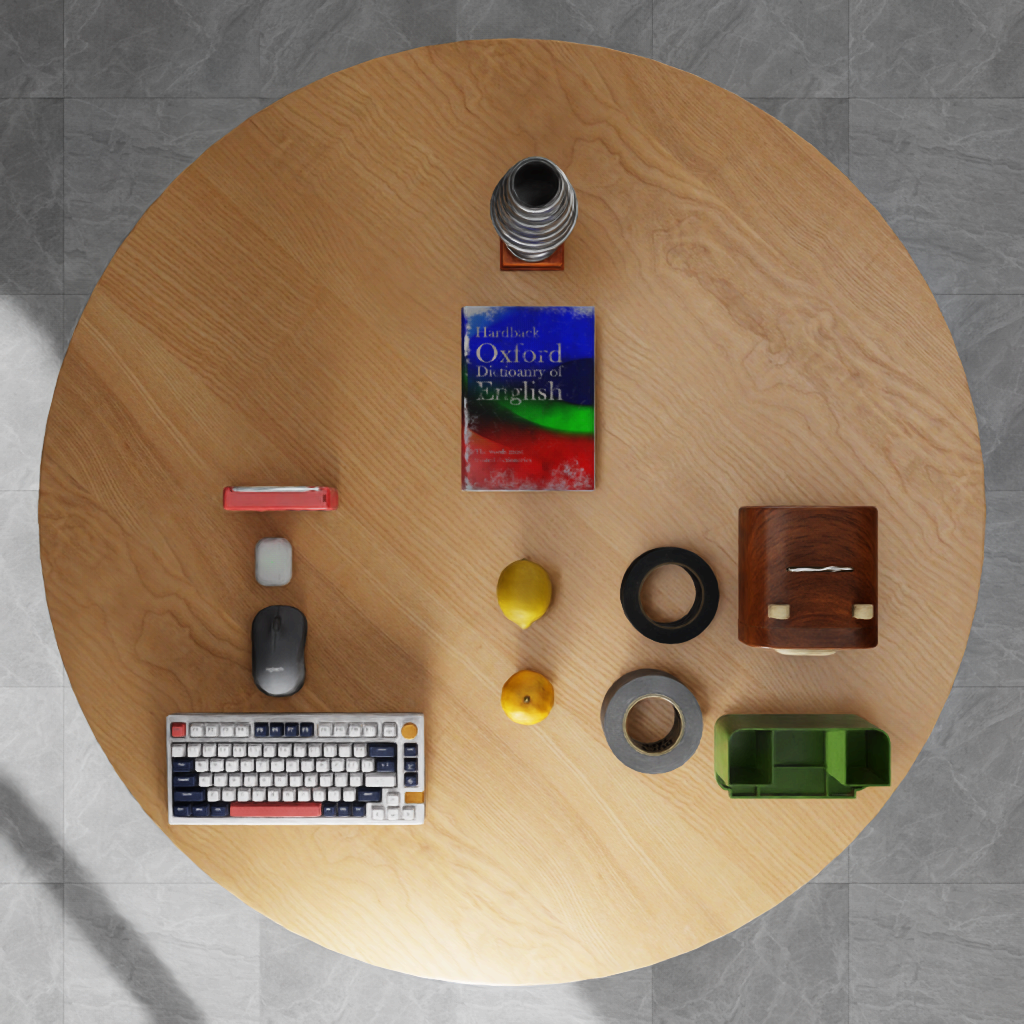}
        \par\vspace{2pt}
        {\small $\Psi=0.6101$}
    \end{minipage}
    \hfill
    \begin{minipage}[t]{0.19\textwidth}
        \centering
        \includegraphics[width=\linewidth]{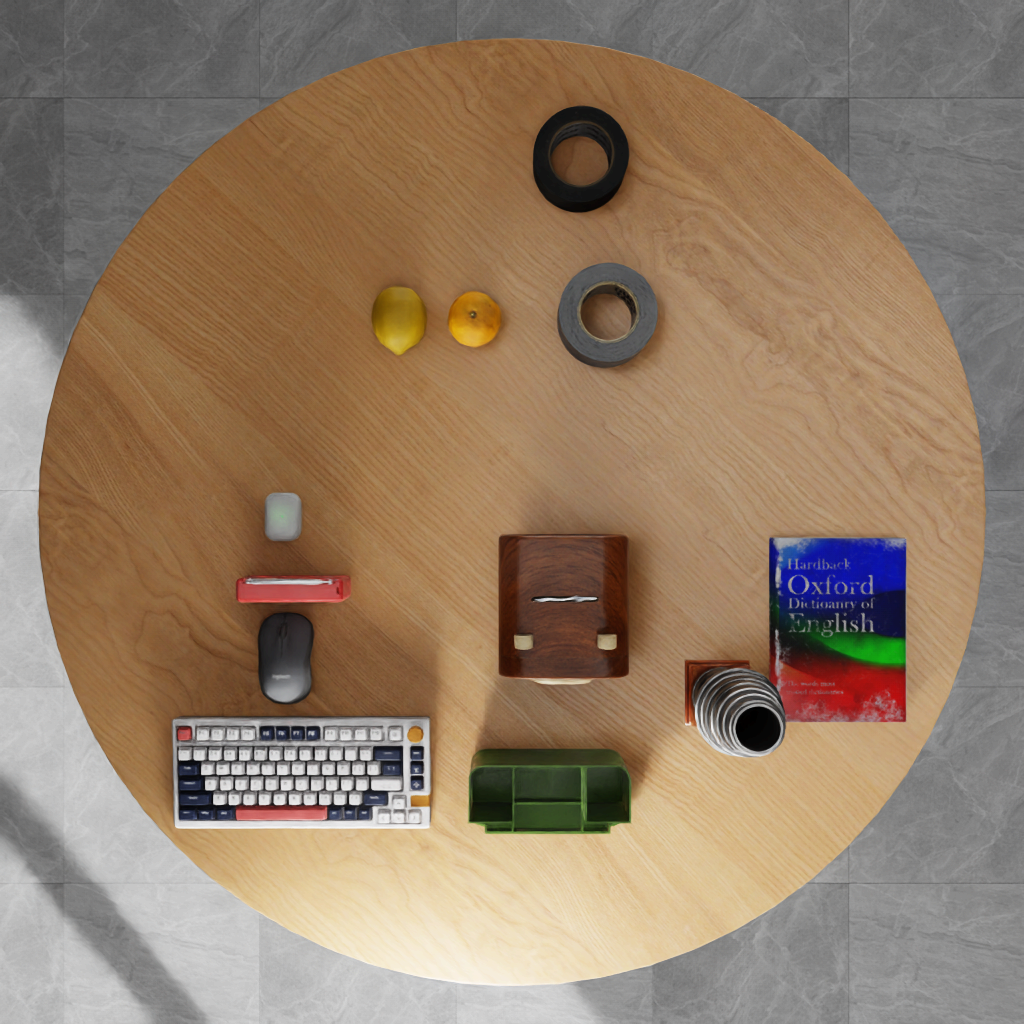}
        \par\vspace{2pt}
        {\small $\Psi=0.6246$}
    \end{minipage}
    \hfill
    \begin{minipage}[t]{0.19\textwidth}
        \centering
        \includegraphics[width=\linewidth]{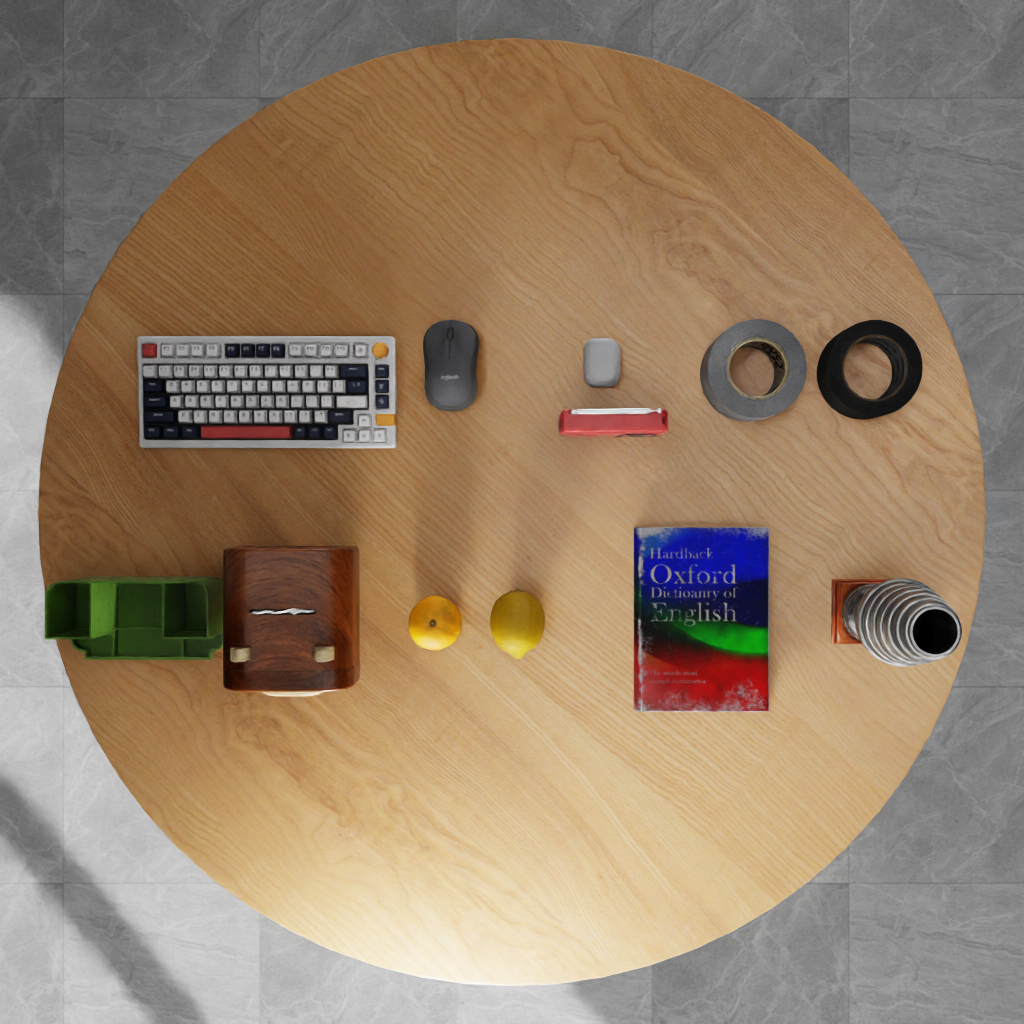}
        \par\vspace{2pt}
        {\small $\Psi=0.6482$}
    \end{minipage}
    \hfill
    \begin{minipage}[t]{0.19\textwidth}
        \centering
        \includegraphics[width=\linewidth]{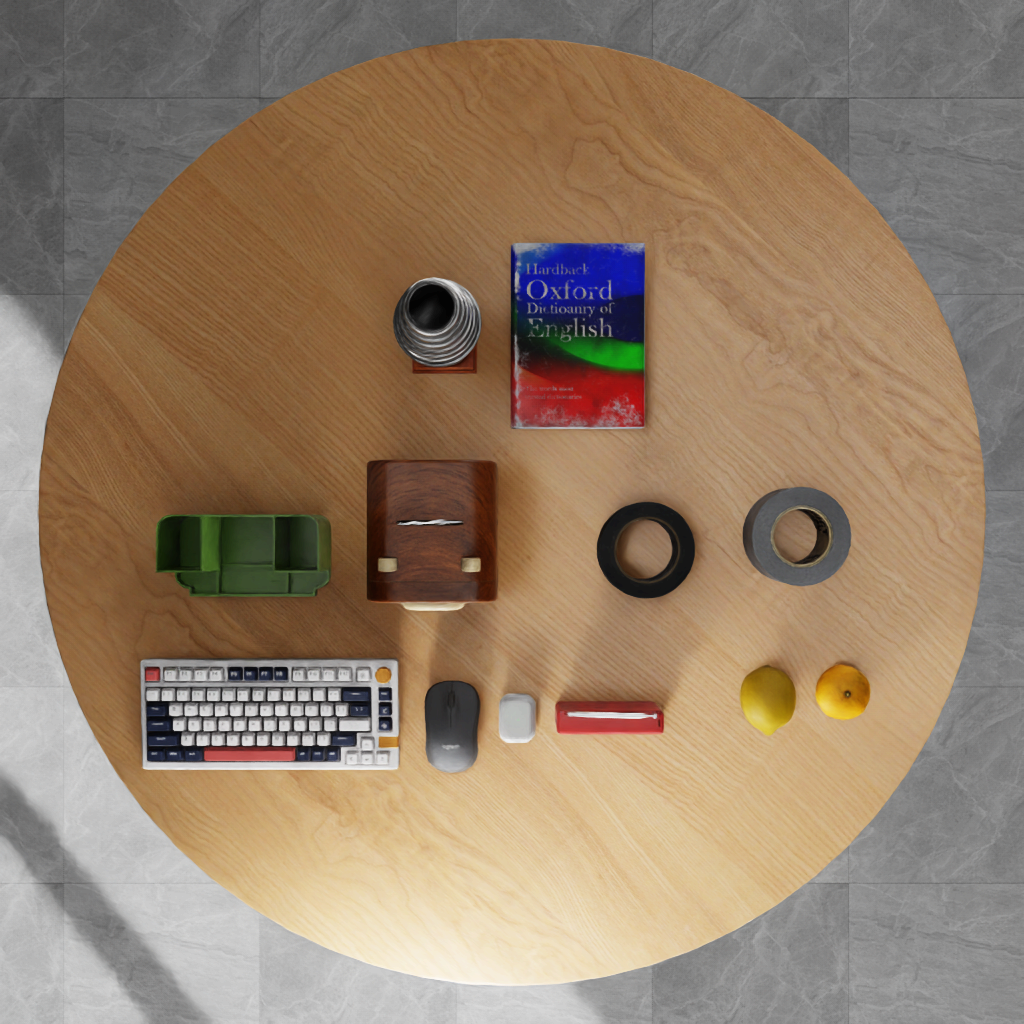}
        \par\vspace{2pt}
        {\small $\Psi=0.6974$}
    \end{minipage}

    \par\vspace{10pt}

    {\small\textbf{WTL}}\par\vspace{3pt}

    \begin{minipage}[t]{0.19\textwidth}
        \centering
        \includegraphics[width=\linewidth]{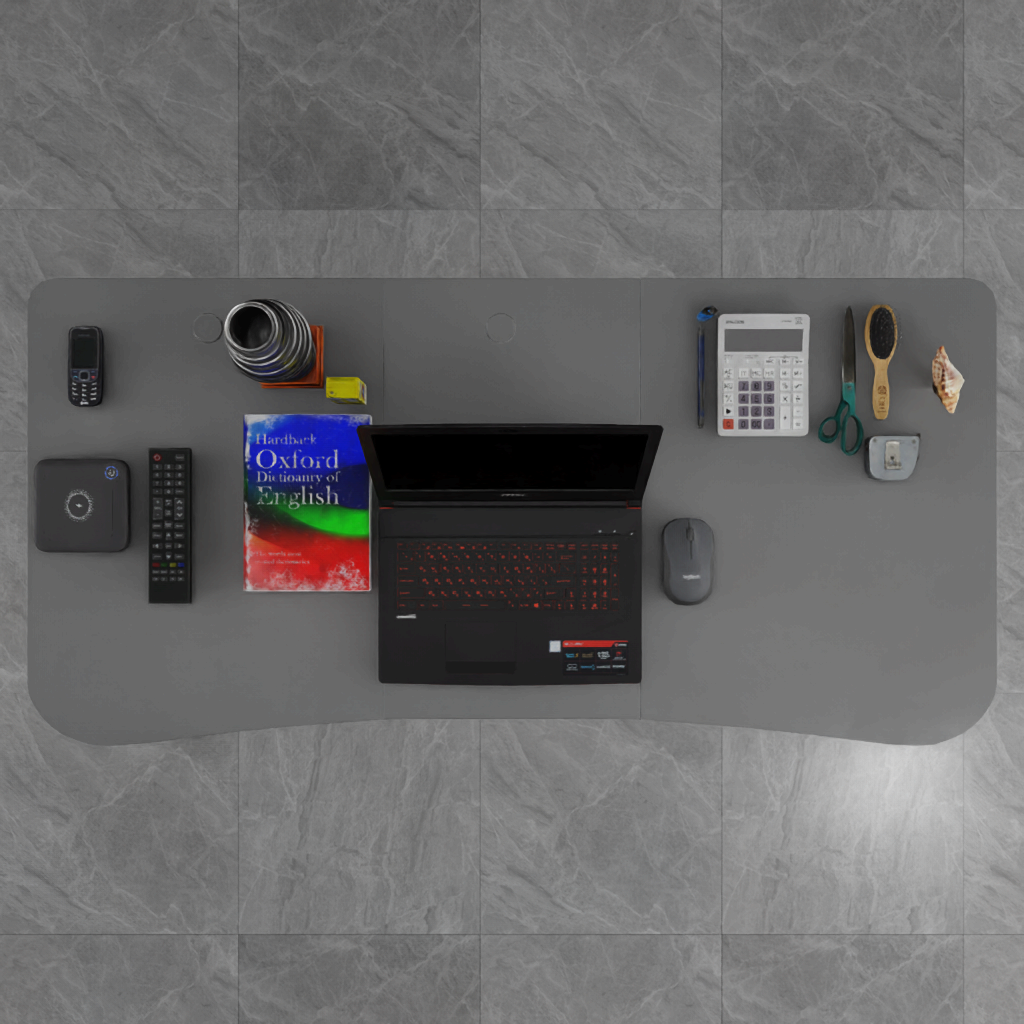}
        \par\vspace{2pt}
        {\small $\Psi=0.5740$}
    \end{minipage}
    \hfill
    \begin{minipage}[t]{0.19\textwidth}
        \centering
        \includegraphics[width=\linewidth]{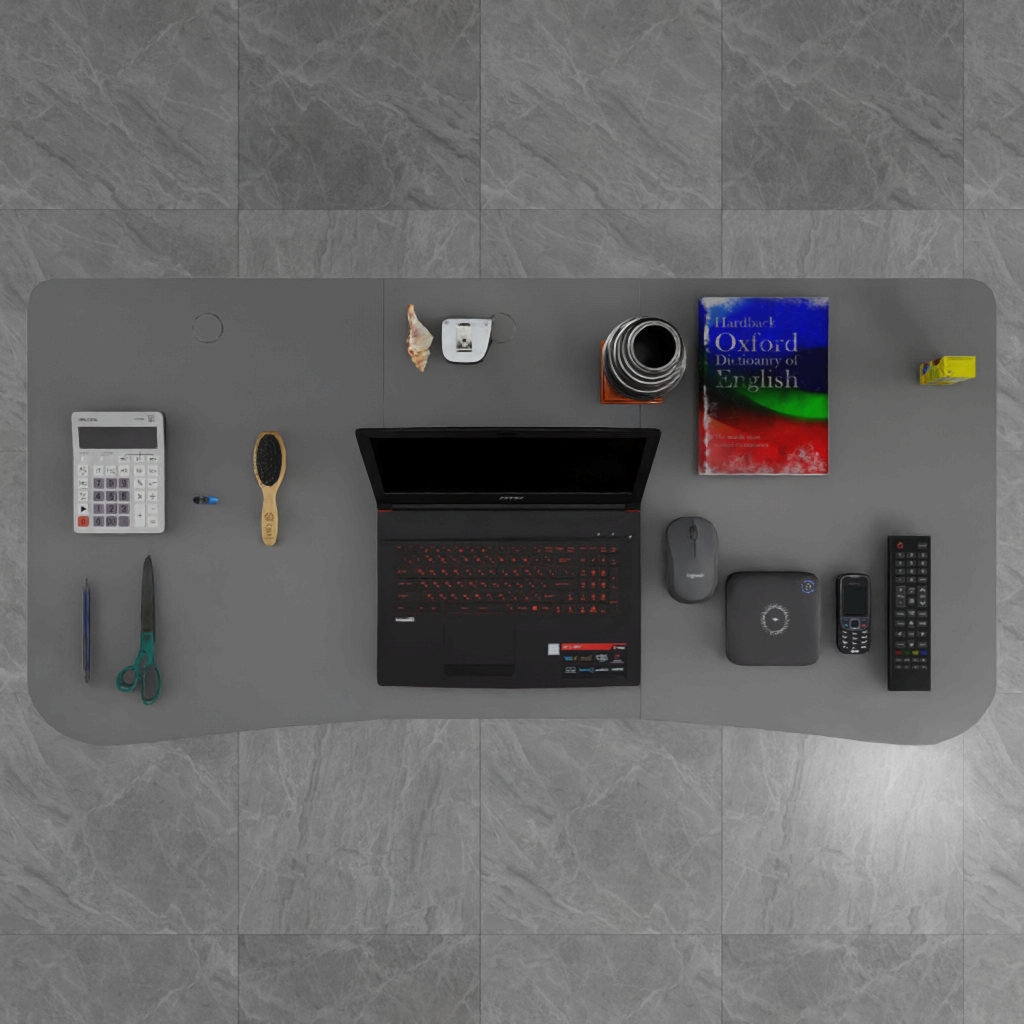}
        \par\vspace{2pt}
        {\small $\Psi=0.6070$}
    \end{minipage}
    \hfill
    \begin{minipage}[t]{0.19\textwidth}
        \centering
        \includegraphics[width=\linewidth]{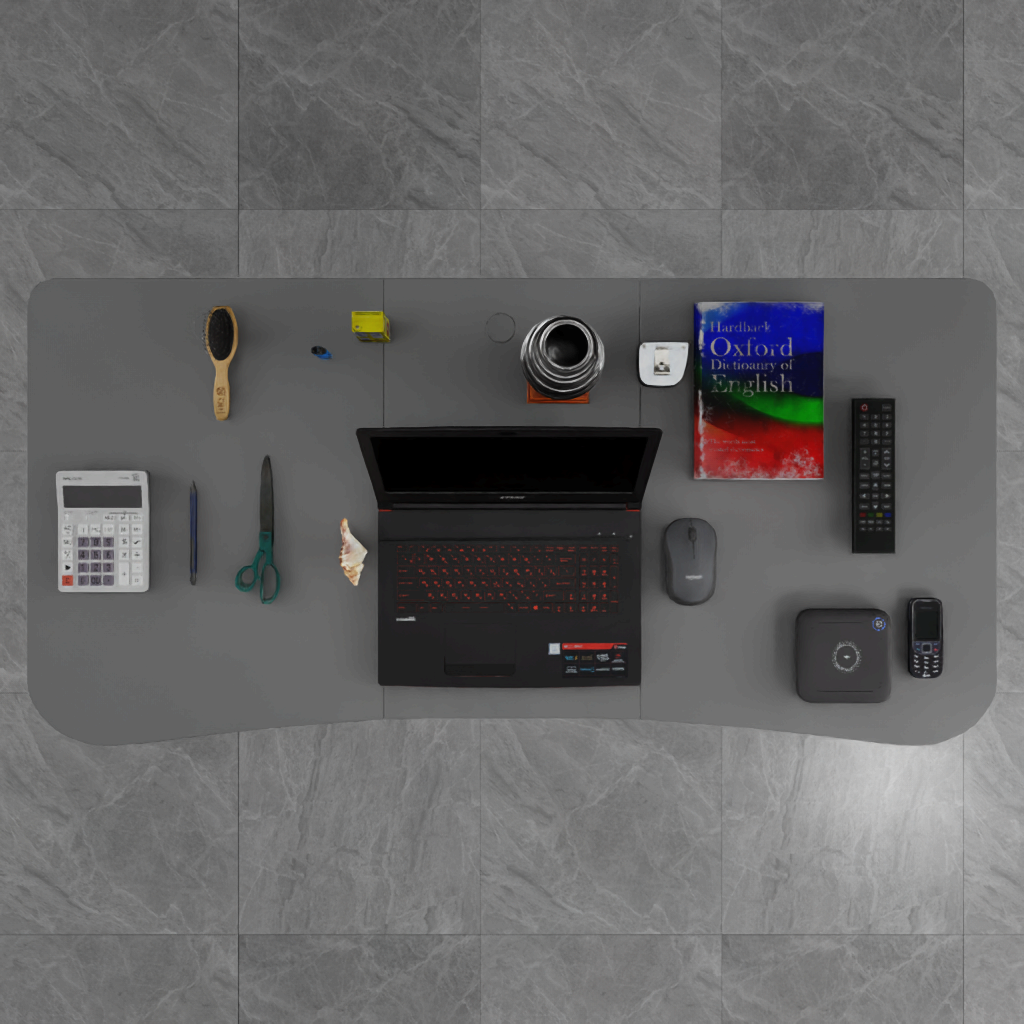}
        \par\vspace{2pt}
        {\small $\Psi=0.6152$}
    \end{minipage}
    \hfill
    \begin{minipage}[t]{0.19\textwidth}
        \centering
        \includegraphics[width=\linewidth]{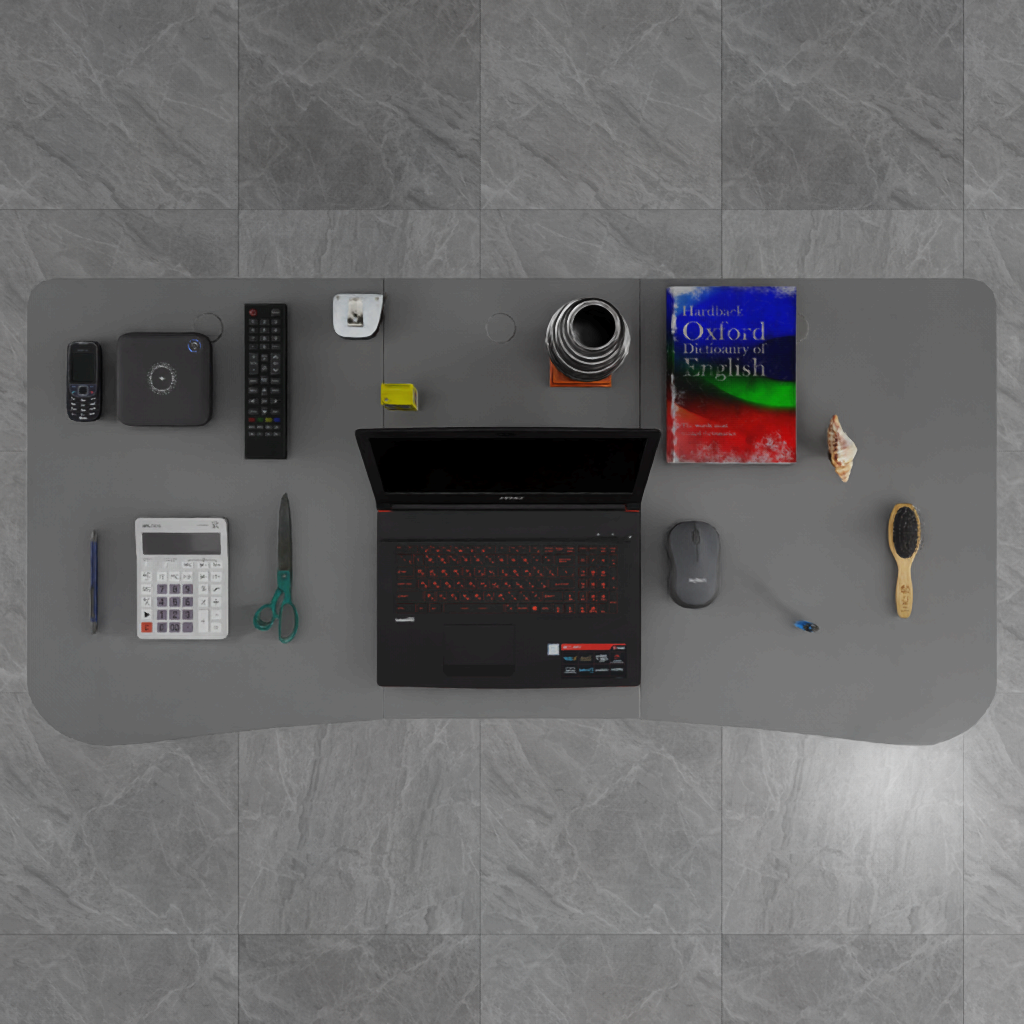}
        \par\vspace{2pt}
        {\small $\Psi=0.6383$}
    \end{minipage}
    \hfill
    \begin{minipage}[t]{0.19\textwidth}
        \centering
        \includegraphics[width=\linewidth]{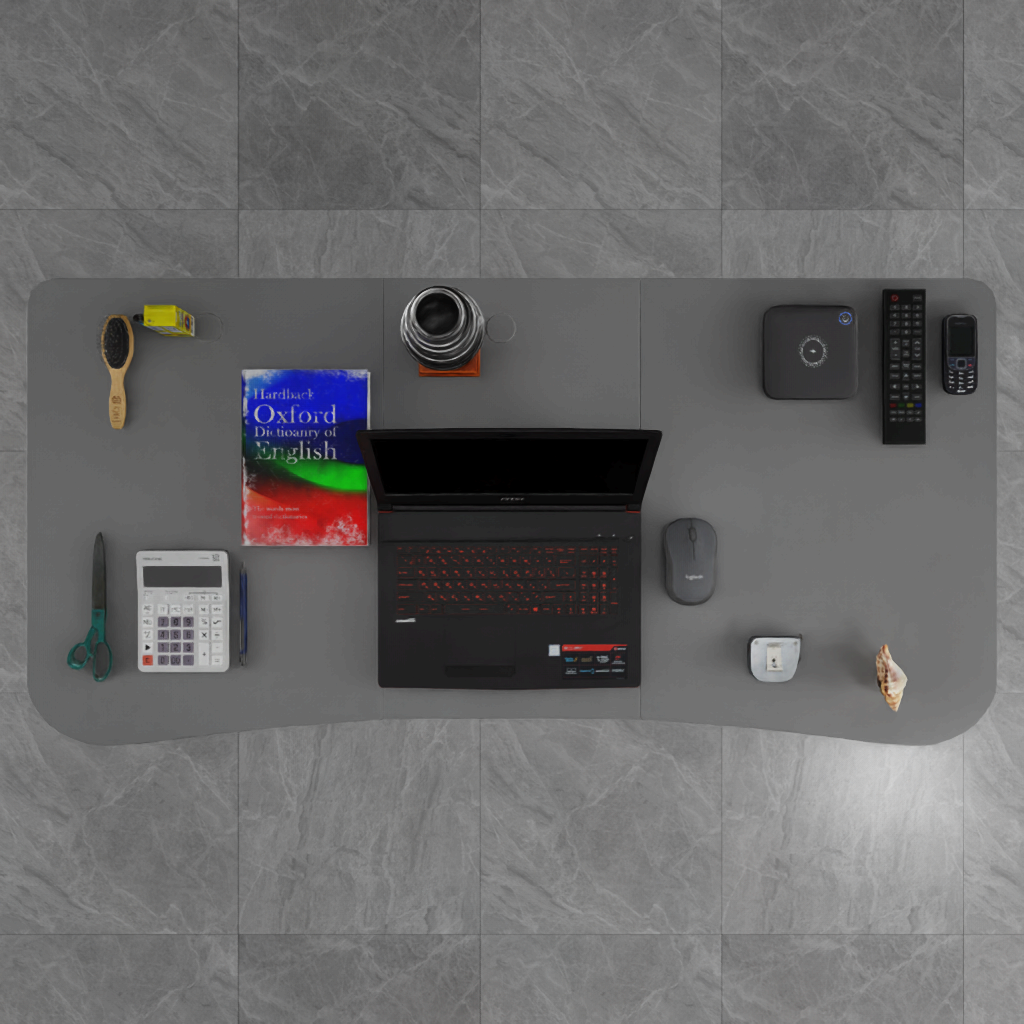}
        \par\vspace{2pt}
        {\small $\Psi=0.6514$}
    \end{minipage}

    \caption{
    \textbf{Multimodal layout generation on GTL and WTL.}
    Each row shows five layouts generated from different
    Gaussian initializations under the same condition.
    Layouts are ordered by tidiness score $\Psi$
    in ascending order.
    }

    \label{fig:multimodal_comparison}

\end{figure*}

To examine layout diversity, we present five
TidyFlow-generated arrangements for a fixed
object set and tabletop condition in each dataset.
As shown in Fig.~\ref{fig:multimodal_comparison},
different Gaussian initializations yield distinct
spatial configurations while preserving the same
objects and tabletop geometry.
For visualization, layouts are ordered by their
tidiness scores $\Psi$.

% =========================================================
% C.3 Stress Tests across Tabletop Areas and Object Counts
% =========================================================

\subsection{Stress Tests across Tabletop Areas and Object Counts}
\label{sec:suppl_stress}

We further investigate the performance of TidyFlow
under different tabletop areas and object counts.
All experiments use the complete GTL and WTL
test sets, with 32 candidates generated per condition.

% =========================================================
% C.3.1 Performance across Tabletop Areas
% =========================================================

\paragraph{Performance across tabletop areas.}

We consider six tabletop geometries with support
polygon areas ranging from $0.679$ to
$1.268\,\mathrm{m}^2$.
Based on the actual polygon area $A$, rather than
the bounding-box area, we divide the tabletops
into three groups: small ($A<0.8\,\mathrm{m}^2$),
medium ($0.8\leq A<1.2\,\mathrm{m}^2$), and
large ($A\geq1.2\,\mathrm{m}^2$).

As shown in \cref{tab:suppl_table_area},
geometric validity generally increases with
tabletop area.
On GTL, GVR increases from $82.73\%$ on small
tabletops to $92.97\%$ on large ones.
The difference is more pronounced on WTL,
where GVR increases from $44.93\%$ to $67.26\%$,
accompanied by lower collision and boundary
violation rates.
These results suggest that additional tabletop
space facilitates geometrically feasible arrangements,
particularly in workstation scenarios.

\begin{table}[t]
\centering

\caption{
Performance across tabletop areas on GTL and WTL.
$M$ denotes the number of test conditions.
All rates are in \%.
}

\label{tab:suppl_table_area}

\footnotesize
\setlength{\tabcolsep}{4pt}

\begin{tabular}{@{}llrrrrr@{}}
\toprule

Dataset & Area & $M$ &
GVR$\uparrow$ &
CR$\downarrow$ &
BVR$\downarrow$ &
FCR@32$\uparrow$ \\

\midrule

GTL & Small  & 80 & 82.73 & 15.31 & 7.89 & 100.00 \\
    & Medium & 81 & 91.01 & 8.22  & 2.28 & 100.00 \\
    & Large  & 80 & 92.97 & 6.76  & 1.02 & 100.00 \\

\midrule

WTL & Small  & 272 & 44.93 & 54.24 & 10.70 & 93.01 \\
    & Medium & 238 & 62.96 & 36.75 & 4.10  & 99.58 \\
    & Large  & 249 & 67.26 & 32.20 & 3.40  & 100.00 \\

\bottomrule
\end{tabular}

\end{table}

% =========================================================
% C.3.2 Performance across Object Counts
% =========================================================

\paragraph{Performance across object counts.}

We further group test conditions into three ranges
according to the number of objects.
As shown in \cref{tab:suppl_object_groups},
geometric validity decreases as object count
increases on both datasets.

On GTL, GVR decreases from $96.29\%$ for
6--8 objects to $74.10\%$ for 12--15 objects,
while FCR@32 remains at $100.00\%$.
A larger degradation is observed on WTL,
where GVR decreases from $88.79\%$ to $44.51\%$
over the same object-count ranges.
The grouped statistics are computed by weighting
each object-count stratum by its number of
test conditions.

\begin{table}[t]
\centering

\caption{
Performance across object-count groups
on GTL and WTL.
$M$ denotes the number of test conditions.
All rates are in \%.
}

\label{tab:suppl_object_groups}

\footnotesize
\setlength{\tabcolsep}{4pt}

\begin{tabular}{@{}llrrrrr@{}}
\toprule

Dataset & $N$ & $M$ &
GVR$\uparrow$ &
CR$\downarrow$ &
BVR$\downarrow$ &
FCR@32$\uparrow$ \\

\midrule

GTL & 6--8   & 90  & 96.29 & 3.02  & 1.18 & 100.00 \\
    & 9--11  & 106 & 88.95 & 9.88  & 3.66 & 100.00 \\
    & 12--15 & 45  & 74.10 & 24.72 & 8.96 & 100.00 \\

\midrule

WTL & 6--8   & 170 & 88.79 & 10.19 & 1.74 & 100.00 \\
    & 9--11  & 180 & 71.67 & 27.71 & 3.70 & 100.00 \\
    & 12--15 & 304 & 44.51 & 55.10 & 7.56 & 97.37 \\

\bottomrule
\end{tabular}

\end{table}

\begin{table}[t]
\centering

\caption{
Performance on high-density WTL conditions
with 15--18 objects.
$M$ denotes the number of test conditions.
All rates are in \%.
}

\label{tab:suppl_high_density}

\footnotesize
\setlength{\tabcolsep}{4pt}

\begin{tabular}{@{}rrrrrr@{}}
\toprule

$N$ & $M$ &
GVR$\uparrow$ &
CR$\downarrow$ &
BVR$\downarrow$ &
FCR@32$\uparrow$ \\

\midrule

15 & 91 & 35.68 & 63.87 & 10.65 & 94.51 \\
16 & 34 & 26.19 & 73.44 & 11.21 & 94.12 \\
17 & 51 & 23.71 & 76.10 & 13.30 & 92.16 \\
18 & 20 & 16.41 & 83.59 & 20.47 & 70.00 \\

\bottomrule
\end{tabular}

\end{table}

% =========================================================
% Table 5: Stability across Independent Training Seeds
% =========================================================

\begin{table*}[!t]
\centering

\caption{
Stability across three independent training seeds
on GTL and WTL.
Results are reported in \%.
Mean $\pm$ SD denotes the mean and sample standard
deviation across three runs.
}

\label{tab:suppl_seeds}

\footnotesize
\setlength{\tabcolsep}{5pt}
\renewcommand{\arraystretch}{1.1}

\begin{tabular}{@{}lrrrrr@{}}
\toprule

Dataset &
Raw GVR$\uparrow$ &
R3 GVR$\uparrow$ &
R3 CR$\downarrow$ &
R3 BVR$\downarrow$ &
FCR@32$\uparrow$ \\

\midrule

GTL (Seed1)
& 18.02 & 88.10 & 11.31 & 2.96 & 100.00 \\

GTL (Seed2)
& 20.19 & 86.55 & 12.21 & 3.99 & 100.00 \\

GTL (Seed3)
& 22.82 & 86.67 & 11.77 & 4.69 & 100.00 \\

\textbf{GTL (Mean $\pm$ SD)}
& $\mathbf{20.34\pm2.40}$
& $\mathbf{87.11\pm0.86}$
& $\mathbf{11.77\pm0.45}$
& $\mathbf{3.88\pm0.87}$
& $\mathbf{100.00\pm0.00}$ \\

\midrule

WTL (Seed1)
& 10.66 & 57.81 & 41.64 & 6.26 & 96.84 \\

WTL (Seed2)
& 9.10 & 57.74 & 42.14 & 3.02 & 97.50 \\

WTL (Seed3)
& 8.61 & 56.89 & 43.00 & 2.66 & 96.44 \\

\textbf{WTL (Mean $\pm$ SD)}
& $\mathbf{9.46\pm1.07}$
& $\mathbf{57.48\pm0.51}$
& $\mathbf{42.26\pm0.69}$
& $\mathbf{3.98\pm1.98}$
& $\mathbf{96.93\pm0.53}$ \\

\bottomrule
\end{tabular}

\end{table*}

% =========================================================
% C.3.3 High-Density Workstation Conditions
% =========================================================

\paragraph{High-density workstation conditions.}

We further examine WTL conditions containing
15--18 objects, as reported in
\cref{tab:suppl_high_density}.
GVR decreases from $35.68\%$ at $N=15$
to $16.41\%$ at $N=18$, while collision and
boundary violation rates increase substantially.

Nevertheless, FCR@32 remains at $70.00\%$
for the most crowded conditions, indicating
that multiple candidate generation can still
recover at least one feasible layout in a
substantial proportion of challenging scenes.

Workstation layouts are particularly challenging
because large objects, such as laptops and monitors,
occupy considerable tabletop space, leaving
less room for surrounding accessories.
As object count increases, the remaining free
space becomes more constrained, making it harder
to satisfy collision and boundary requirements
simultaneously.

% =========================================================
% Discussion
% =========================================================

Overall, these results indicate that geometric
feasibility becomes more challenging as the
available tabletop space decreases or the number
of objects increases.
The stronger degradation on WTL is consistent
with the greater spatial occupancy and functional
constraints of workstation arrangements.

These comparisons are descriptive rather than
strictly controlled experiments, since tabletop
shapes, object dimensions, and object combinations
may also vary across test conditions.

\subsection{Stability across Independent Training Seeds}
\label{sec:suppl_seeds}

To assess training stability, we independently
train TidyFlow using three random seeds with
identical data splits, model configurations,
and training protocols.
All runs use a maximum capacity of 24 objects
and share the same initial-noise bank during
evaluation to isolate training randomness.

As shown in \cref{tab:suppl_seeds},
TidyFlow achieves mean GVRs of
$87.11\pm0.86\%$ on GTL and
$57.48\pm0.51\%$ on WTL after refinement.
The relatively small standard deviations
indicate consistent geometric validity
across independent training runs.

% No author-identifying information or links in this anonymous review draft.
% References can be added here if the final supplement cites extra literature.